\documentclass[preprint,3p,sort&compress,10pt]{elsarticle}
\pdfoutput=1
\usepackage[T1]{fontenc}
\usepackage[utf8]{inputenc}
\usepackage[hidelinks]{hyperref}

\journal{...}

\usepackage{amsmath}
\usepackage{amssymb}
\usepackage{amsthm}

\usepackage[usenames,dvipsnames,svgnames,table]{xcolor}
\usepackage{graphicx}
\usepackage{xspace}
\usepackage{xparse}
\usepackage{enumerate}
\usepackage{tabularx}
\usepackage{booktabs}
\usepackage{csquotes}

\usepackage{latexsym}
\usepackage{stmaryrd}
\usepackage{wasysym}
\usepackage{multicol}
\usepackage{multirow}
\usepackage{subcaption}
\usepackage{xspace}
\usepackage{graphicx}
\usepackage{booktabs}
\usepackage{arydshln}
\usepackage{enumerate}
\usepackage{csquotes}
\usepackage{xparse}
\usepackage{tabularx}
\usepackage[kerning,spacing,tracking,expansion]{microtype}

\usepackage{tikz}
\usetikzlibrary{fadings}
\usetikzlibrary{shapes,decorations,positioning}
\usetikzlibrary{fadings,shapes,decorations,positioning,calc}
\usetikzlibrary{decorations.markings,decorations.pathreplacing}
\usetikzlibrary{shapes.geometric,automata,positioning}
\usetikzlibrary{shadows}\usetikzlibrary{calc}
\usetikzlibrary{decorations,decorations.pathmorphing,decorations.pathreplacing}
\usetikzlibrary{calc}

\DeclareDocumentCommand{\todo}{o g}{\IfNoValueTF{#1}{\begingroup\color{magenta}TODO: #2\endgroup}{\begingroup\color{magenta}#1 #2\endgroup}}

\DeclareDocumentCommand{\remark}{o g}{\IfNoValueTF{#1}{\begingroup\color{magenta}Anmerkung: #2\endgroup}{\begingroup\color{magenta}#1 #2\endgroup}}
\makeatletter
\newcommand{\ifLatexThree}[2]{\@ifpackageloaded{xparse}{#1}{#2}}
\newcommand{\ifAMSmath}[2]{\@ifpackageloaded{amsmath}{#1}{#2}}
\newcommand{\ifMathSCR}[2]{\@ifpackageloaded{mathrsfs}{#1}{#2}}
\newcommand{\ifMathHyperREF}[2]{\@ifpackageloaded{hyperref}{#1}{#2}}
\makeatother

\ifLatexThree{%
	\NewDocumentCommand{\headword}{s o m}{\IfBooleanTF{#1}{#3}{\textbf{#3}}\IfNoValueTF{#2}{\index{#3}}{\index{#2}}}%
}{%
	\makeatletter
	\def\@headword#1{\textbf{#1}\index{#1}}%
	\def\@@headword#1{#1\index{#1}}%
	\def\headword#1{\@ifstar\@headword{#1}\@@headword{#1}}%
	\makeatother
}

\makeatletter
\@ifundefined{naturals}{}{}
\@ifundefined{ol}{}{}
\makeatother

\makeatletter
\newcommand{\textlabelmarker}[1]{%
	\protected@edef\@currentlabel{#1}%
	\phantomsection%
}
\newcommand{\textlabel}[2]{%
	\textlabelmarker{#1}%
	#1\label{#2}%
}
\makeatother

\ifx\nmableit\undefined
\makeatletter
\ifx\centernot\undefined
\newcommand*{\centernot}{%
	\mathpalette\@centernot
}
\fi
\def\@centernot#1#2{%
	\mathrel{%
		\rlap{%
			\settowidth\dimen@{$\m@th#1{#2}$}%
			\kern.5\dimen@
			\settowidth\dimen@{$\m@th#1=$}%
			\kern-.5\dimen@
			$\m@th#1\not$%
		}%
		{#2}%
	}%
}
\makeatother
\DeclareRobustCommand\nmableitSymb{\mathrel|\mkern-.5mu\joinrel\sim} %
\newcommand{\nmableit}{\ensuremath{\mbox{$\,\nmableitSymb\,$}}} %
\fi

\makeatletter
\ifMathHyperREF{%
	\newcommand{\seqref}[1]{\hyperref[{#1}]{\textup{\tagform@split{\getrefnumber{#1}}}}}%
}{%
	\newcommand{\seqref}[1]{\textup{\tagform@split{\getrefnumber{#1}}}}%
}
\newcommand\tagform@split[1]{%
	\begingroup
	\m@th\normalfont(\ignorespaces #1\unskip\@@italiccorr)%
	\endgroup
}
\makeatother

\makeatletter
\newcommand{\leqnomode}{\tagsleft@true\let\veqno\@@leqno}
\newcommand{\reqnomode}{\tagsleft@false\let\veqno\@@eqno}
\newcommand{\pushright}[1]{\ifmeasuring@#1\else\omit\hfill$\displaystyle#1$\fi\ignorespaces}
\newcommand{\pushleft}[1]{\ifmeasuring@#1\else\omit$\displaystyle#1$\hfill\fi\ignorespaces}
\newcommand{\specialcell}[1]{\ifmeasuring@#1\else\omit$\displaystyle#1$\ignorespaces\fi}

\makeatother
\newcommand{\ksIF}{\text{if }}
\newcommand{\ksTHEN}{\text{, then }}

\newcommand{\ksAND}{\text{ and }}
\newcommand{\ksOR}{\text{ or }}

\newcommand{\ksIMPLIES}{\text{ implies }}

\newcommand{\ksIFFlong}{\text{ if and only if }}

\newcommand{\ksForAll}{\text{ for all }}
\newcommand{\ksOtherwise}{\text{otherwise}}

\newcommand{\modelsOf}[1]{\ensuremath{\llbracket #1\rrbracket}}
\newcommand{\modelsOfES}[1]{\modelsOf{#1}}
\newcommand{\negOf}[1]{{\ensuremath{\neg{#1}}}}

\newcommand{\minOf}[2]{\ensuremath{\min(#1,#2)}}

\renewcommand{\modelsOf}[1]{\ensuremath{\ksMod(#1)}}

\DeclareMathOperator{\Cn}{Cn}

\DeclareMathOperator{\ksBel}{Bel}
\newcommand{\beliefsOf}[1]{\ensuremath{\ksBel(#1)}}

\newcommand{\setAllES}{\ensuremath{\mathcal{E}}}

\newcommand{\propLang}{\ensuremath{\mathcal{L}}}

\newcommand{\ramseyCond}[2]{\ensuremath{(\,#1\,|\,#2\,)}}
\newcommand{\contractionCond}[2]{\ensuremath{[\,#1\,|\,#2\,]}}

\ifMathSCR{%
}{%
}

\usepackage{environ,xparse,xkeyval,ifthen}
\newif\ifpostulatepresent\postulatepresentfalse

\ExplSyntaxOn
\NewEnviron{postulate}[2]%
{%
	\tl_gclear:N \g_tmpa_tl%
	\tl_gput_right:Nn \g_tmpa_tl { \ifpostulatepresent\\[\smallskipamount]\fi\global\postulatepresenttrue }%
	\tl_gput_right:Nn \g_tmpa_tl { \noindent\ignorespaces\ifthenelse{\equal{#1}{}}{}{(\textlabel{#1}{pstl:#1})} & }%
	\tl_gput_right:No \g_tmpa_tl { \BODY }%
	\tl_gput_right:Nn \g_tmpa_tl { & \hspace{1.5\medskipamount} #2  }%
	\group_insert_after:N \g_tmpa_tl%
}
\ExplSyntaxOff

\newtheorem{theorem}{Theorem}[section]
\newtheorem{proposition}[theorem]{Proposition}

\newtheorem{corollary}[theorem]{Corollary}

\newtheorem{definition}[theorem]{Definition}
\newtheorem{example}[theorem]{Example}

	\makeatletter
\renewcommand{\textlabel}[2]{%
	\protected@edef\@currentlabel{#1}%
	\phantomsection%
	#1\label{#2}%
}
\makeatother

\renewcommand{\headword}[1]{\emph{#1}}

\makeatletter
\renewcommand{\leqnomode}{\tagsleft@true}
\renewcommand{\reqnomode}{\tagsleft@false}
\makeatother

\newcommand{\revision}{\ensuremath{*}}
\newcommand{\contraction}{\ensuremath{\div}}
\newcommand{\change}{\ensuremath{\circ}}

\renewcommand{\modelsOf}[1]{\llbracket{#1}\rrbracket}

\begin{document}

\begin{frontmatter}

\title{A Conditional Perspective on the Logic of Iterated Belief Contraction}

\author[FUH]{Kai Sauerwald}
\ead{kai.sauerwald@fernuni-hagen.de}
\author[TUD]{Gabriele Kern-Isberner}
\ead{gabriele.kern-isberner@cs.tu-dortmund.de}
\author[FUH]{Christoph Beierle}
\ead{christoph.beierle@fernuni-hagen.de}
\address[FUH]{%
	FernUniversität in Hagen, Faculty of Mathematics and Computer Science, Knowledge-Based Systems, 58084 Hagen, Germany}
\address[TUD]{%
	TU Dortmund University, Department of Computer Science, Information Engineering, 44221 Dortmund, Germany}

\begin{abstract}
Classical AGM belief change theory provides only limited information on how to deal with iterative belief revision, leading to the question of how appropriate iteration principles look alike. 
The contributions to the interconnection between iterative belief revision and conditional beliefs lead to fundamental insights on that matter. 
Besides of clarifying to what amount iterated revision should respect conditional beliefs, these investigations identified several principles for iterated revision that are today widely accepted by the belief change community and beyond.
Specifically, due to the seminal work by Darwiche and Pearl, it is standard, that principles for iterated revision and change of conditional beliefs are considered as two sides of the same coin. 
However, for iterated contraction the situation is different, there are only a small amount of iteration principles present in the literature. 
Moreover, for none of these iterated principles for contraction, there is a known equivalent formulation in terms of change of conditional beliefs.
	
In this article, we consider iteration principles for contraction, with the goal of identifying properties for contractions that respect  conditional beliefs. Therefore, we investigate and evaluate four groups of iteration principles for contraction which consider the dynamics of conditional beliefs.
For all these principles, we provide semantic characterization theorems and provide formulations by postulates which highlight how the change of beliefs and of conditional beliefs is constrained, whenever that is possible.
The first group is similar to the syntactic Darwiche-Pearl postulates. 
As a second group, we consider semantic postulates for iteration of contraction by Chopra, Ghose, Meyer and Wong, and by Konieczny and Pino Pérez, respectively, and  we provide novel syntactic counterparts.	
Third, we propose a contraction analogue of the independence condition by Jin and Thielscher. For the fourth group we consider natural and moderate contraction by Nayak.
Methodically, we make use of conditionals for contractions, so-called contractionals and furthermore, we propose and employ the novel notion of \( \alpha \)-equivalence for formulating some of the new postulates.
	
\end{abstract}
 

\end{frontmatter}

\section{Introduction}
\label{sec:introduction}

For the three main classes of theory change, revision, expansion and contraction, different characterizations are known \cite{KS_FermeHansson2018}, which are heavily supported by the correspondence between revision and contraction via the Levi and Harper identities \cite{KS_Levi1977,KS_Harper1976}.
The situation is different for iterated belief change, focussing on belief change operators which, due to their nature, can be applied iteratively and thus, to more than one epistemic state. 
In this field, one of the most influential articles is the seminal paper by Darwiche and Pearl \cite{KS_DarwichePearl1997}, establishing the insight that belief sets are not a sufficient representation for iterated belief revision.
An agent has to encode more information about her belief change strategy into her \emph{epistemic state}.
The common approach to epistemic states is by a semantic encoding, also established by Darwiche and Pearl \cite{KS_DarwichePearl1997}, which is an extension of Katsuno and Mendelzon's characterization of the class of revisions by Alchourr{\'{o}}n, G{\"{a}}rdenfors and Makinson \cite{KS_AlchourronGaerdenforsMakinson1985} in terms of plausibility orderings \cite{KS_KatsunoMendelzon1992}, where it is assumed that the epistemic states contain an order of the possible worlds (or interpretations).
Another insight presented in the work by Darwiche and Pearl is that the revision strategy deeply corresponds with conditional beliefs.
Based on this insight, Darwiche and Pearl provided a three-fold view on iterated revision, focussing on the change of \emph{beliefs}, of \emph{conditional beliefs}, and of \emph{relations over possible worlds}.
As revision should respect the change of conditional beliefs \cite{KS_Boutilier1996}, it is desirable to have additional postulates that guarantee intended behaviour in forthcoming changes.
This leads to the so-called Darwiche-Pearl postulates for iterated revision, which are a de facto standard set of postulates for iterated revision \cite{KS_DarwichePearl1997}.

Similar work has been done partially in recent years for iterated contraction. 
Caridroit, Konieczny and Marquis \cite{KS_CaridroitKoniecznyMarquis2015} provided postulates for contraction in propositional logic and a characterization with plausibility orders in the style of Katsuno and Mendelzon \cite{KS_KatsunoMendelzon1992}.
By this characterization, the main characteristic of a contraction  with $ \alpha $ is  that the worlds of the previous state remain plausible and that the most plausible counter-models of $ \alpha $ become plausible.
Chopra, Ghose, Meyer and Wong \cite{KS_ChopraGhoseMeyerWong2008} transferred these to the Darwiche-Pearl framework of epistemic states, contributed semantic postulates for contraction on epistemic states in the fashion of Darwiche and Pearl, and equivalent syntactic postulates that depend on a revision function.
In the same framework, Konieczny and {Pino P{\'{e}}rez} provided additional syntactic iteration postulates for contraction which are independent of revision operators \cite{KS_KoniecznyPinoPerez2017}. %
However, none of these approaches on iterated contraction provides a focus on conditionals like the work by Darwiche and Pearl \cite{KS_DarwichePearl1997}. 
Note that characterization results from revision do not transfer easily between iterated revision and contraction,
because for iterated belief change the connection between revision and contraction via the Levi and Harper identities is not straightforward \cite{KS_KoniecznyPinoPerez2017}. 
Thus, the connection between these kinds of changes is an interesting subject of research on its own \cite{KS_KoniecznyPinoPerez2017,KS_ChandlerBooth2019,KS_NayakGoebelOrgunPham2006,KS_BoothChandler2019,KS_BoothChandler2016,KS_HildSpohn2008}.

In this article, we investigate various principles for iterated contraction in the fashion Darwiche and Pearl did for revision \cite{KS_DarwichePearl1997}.
    Therefore, we present a three-fold view on iterated principles for contraction, considering the change of \emph{beliefs}, the change of \emph{conditional beliefs}, and the change of \emph{relations over possible worlds}. 
    We focus in particular on conditional beliefs, and approach iteration principles for contraction from the perspective of conditional beliefs. 
    The iteration principles we consider for contraction are inspired by principles for iterated revision and comprise four different groups.

At first, we consider postulates for iterated contraction that are similar to the syntactic versions of the  Darwiche-Pearl postulates for iterated revision \cite{KS_DarwichePearl1997}.
For this, we use specific conditionals for contraction, so-called contractionals, which were first studied by Bochman \cite{KS_Bochman2001}.
The rationale is that the Darwiche-Pearl postulates are invariance conditions for conditionals which encodes revisions, so-called Ramsey-test conditionals.
We obtain that a straight-forward way of reformulation leads either to (partially) implausible postulates or to postulates not cover all cases one might expect.
Whenever possible, we provide characterization theorems for the new postulates.

Second, we consider analogue postulates to the semantic version of the Darwiche-Pearl postulates.
We show that these new set of postulates and the set of postulates given by Konieczny and {Pino P{\'{e}}rez} \cite{KS_KoniecznyPinoPerez2017} define the same class of contraction operators in light of the basic postulates.
However, we argue that our new postulates highlight new aspects of iterated contraction operators.
Especially, the new postulates highlight the specific role of conditionals in the same manner as the postulates for iterative revision by Darwich and Pearl do.
For some of the new postulates, we make use of a novel equivalence relation for epistemic states with respect to a proposition \( \alpha \), called \( \alpha \)-equivalence, yielding an alternative formulation of these postulates.
Furthermore, we argue that our new postulates are more succinct than the postulates by Konieczny and {Pino P{\'{e}}rez}, because they deal less with changes of disjunctive beliefs.
Succinctness of postulates is of particular importance when concepts on iterated belief change developed for changes in propositional logic are translated to other formalisms \cite{KS_ShapiroPagnuccoLesperanceLevesque2011,KS_DelgrandePeppasWoltran2013,KS_DelgrandePeppas2015,KS_DelgrandePeppasWoltran2018}, and also when iterated belief contraction is used for modelling phenomena, like forgetting (cf.\ \cite{KS_EiterKern-Isberner2019} for a recent survey,
or \cite{BeierleKernIsbernerSauerwaldBockRagni2019KIzeitschrift,KS_Kern-IsbernerBockSauerwaldBeierle2019}).

Third, we explore a notion of independence for contraction in the sense of independence for revision by Jin and Thielischer \cite{KS_JinThielscher2007} and show that this notion of independence rules out trivial contraction operators.

Forth, as an application, we employ contractionals to give alternative characterizations for natural contraction and moderate contraction \cite{KS_NayakGoebelOrgunPham2006,KS_NayakGoebelOrgun2007,KS_RamachandranNayakOrgun2012}.

\pagebreak[4]\noindent
In summary, the main contributions of this article 
    are:
\begin{itemize}
    	\item Iteration principles for contraction, including a contraction analogue for the notion of \emph{independence} known for revision \cite{KS_JinThielscher2007};
    \item Characterization theorems for all principles, connecting postulates from the viewpoints of changing \emph{beliefs}, of changing \emph{conditional beliefs}, and of changing \emph{relations over worlds};
    \item Introduction of the novel notion of \( \alpha \)-equivalence for the intuitive formulation of postulates;
    \item Succinct postulates for known iteration principles, including postulates for natural and moderate contraction by means of conditionals;
    \item Employment of contractionals for iteration principles for contraction and their connection to the \( \alpha \)-equivalence for epistemic states.
\end{itemize}
This article is a revised and largely extended version of our prior work from ECAI~\cite{KS_SauerwaldKern-IsbernerBeierle2020}.
     In particular, we consider and characterize a larger variety of iteration principles for AGM contraction; moreover, we provide a notion of independence for contraction, and characterizations of natural and moderate contraction with the means of conditionals.

The rest of the article is organised as follows. Section \ref{sec:prelim} provides the technical background, especially on the Darwiche-Pearl framework, iterated belief revision and iterated belief contraction.
We introduce Ramsey-test conditionals for revision and the Darwiche-Pearl postulates for the dynamics of conditional beliefs in Section \ref{sec:conditionals_revision}.
In Section \ref{sec:contractionsals} contractionals are introduced  as a contraction counterparts to Ramsey-test conditionals.
We investigate  in Section \ref{sec:syn_postulates} contraction analogues to the syntactic Darwiche-Pearl postulates.
Section \ref{sec:sem_postulates} considers semantic postulates for iterated contraction; 
in particular we characterise these semantic postulates by contractional based postulates.
In Section \ref{sec:ind_contraction} we consider a counter-part for contraction to the independence condition for revision by Jin and Thielscher \cite{KS_JinThielscher2007}.
Section \ref{sec:concrete_contraction_strategies} characterizes natural contraction and moderate contraction from the viewpoint of contractionals.
Finally, Section \ref{sec:conclusion} concludes and points out future work.

\section{Background on Logic and Preorders}
\label{sec:prelim}

Let $ \Sigma $ be a propositional signature (non-empty finite set of propositional variables) and $ \propLang $ a propositional language over $ \Sigma $. 
With lower Greek letters $ \alpha,\beta,\gamma,\ldots $ we denote formulas in $ \propLang $ and with lower case letters $ a,b,c,\ldots$ propositional variables from $ \Sigma $.
	The set of 
	propositional interpretations $ \Omega $, also called set of worlds, is identified with the set of corresponding complete conjunctions over $ \Sigma $.
Propositional entailment is denoted by $ \models $, the set of models of $ \alpha $ with $ \modelsOf{\alpha} $, and $ Cn(\alpha)=\{ \beta\mid \alpha\models \beta \} $ denotes the deductive closure of $ \alpha $.
For a set of formulas $ X\subseteq \propLang $ we define $ Cn(X)=\{ \beta \mid X\models\beta \}=\bigcap_{\alpha\in X}Cn(\alpha) $.
For a set of worlds $ \Omega'\subseteq \Omega $ and  a total preorder $ {\leq} \subseteq 
\Omega\times\Omega $ (total, reflexive and transitive relation) over $ \Omega $, we denote with 
$ \min(\Omega',\leq)=\{ \omega\in\Omega' \mid  \omega\leq \omega' \ksForAll \omega'\in\Omega' \} $
the set of all minimal worlds in $ \Omega' $  with respect to \( \leq \).
If $ \leq $ is a total preorder, then we denote with $ < $ its strict part, i.e. $ x < y $ iff $ x \leq y $ and $ y \not\leq x $. 
We lift \( \leq \) to formulas \( \alpha,\beta\in\propLang \)  by defining \( \alpha \leq \beta \) if for each \( \omega \in \min(\modelsOf{\beta},\leq) \) there is some \( \omega' \in \min(\modelsOf{\alpha},\leq) \) such that \( \omega' \leq \omega \). Consequently, we have \( \alpha < \beta \)  if for each \( \omega \in \min(\modelsOf{\beta},\leq) \) there is some \( \omega' \in \min(\modelsOf{\alpha},\leq) \) such that \( \omega' < \omega \).

\section{Belief Change and Epistemic States}
\label{sec:prelim_BC}

Classical belief change theory by Alchourrón, Gärdenfors and Makinson \cite{KS_AlchourronGaerdenforsMakinson1985} (AGM)
deals with belief sets as representation of belief states, i.e., deductively closed sets of
propositions. 
Change operators in this setup are often understood as operators for an individual belief set, i.e. formally, for a belief set \( K \)  the approach yields a function of type 
 \( \propLang \to \mathcal{P}(\propLang) \) which describes how to change \( K \) according to new information.

The area of iterated belief change investigates principles and their characterization for sequences of changes. 
In particular, changes in an agent's beliefs might depend on changes made in the past, and changes might have an impact on subsequent changes.
Therefore, one has to give up belief sets as representations \cite{KS_FriedmanHalpern1996,KS_DarwichePearl1997}.
To deal with these issues, the Darwiche-Pearl framework abstracts from the classical setup in two ways.

First, instead of using belief sets as representations, abstract entities are used, called \headword{epistemic states} \cite{KS_DarwichePearl1997}, 
in which the agent maintains necessary information for all possible subsequent belief changes. 
With $ \setAllES $ we denote the set of all epistemic states over $ \propLang $.
Every epistemic state $ \Psi\in\setAllES $ induces a set of beliefs $ \beliefsOf{\Psi}\subseteq \mathcal{L} $, which is a deductively closed set with respect to $ \propLang $.
We write $ \Psi\models\alpha $ if $ \alpha\in\beliefsOf{\Psi} $ and we use $ \modelsOf{\Psi}$ as shorthand for $\modelsOf{\beliefsOf{\Psi}}  $.
Furthermore, we assume satisfaction of the following principle for the set of epistemic states:
\begin{equation}
	\ksIF L\subseteq\propLang \ksAND L \text{ is consistent, then there exists } \Psi\in\setAllES \text{ with } \Cn(L)=\beliefsOf{\Psi}	\tag{unbiased}\label{pstl:unbiased}
\end{equation}
Intuitively, \ref{pstl:unbiased}ness guarantees that for every consistent belief set \( L \) there exists at least one epistemic state \( \Psi \) such that \( \beliefsOf{\Psi}=L \).
We make no explicit further requirements on $ \setAllES $. 
Consequently, \( \setAllES \) might contain two distinct epistemic states \( \Psi_1,\Psi_2 \) (or even more) inducing the same belief set, i.e.  \( \beliefsOf{\Psi_1} = \beliefsOf{\Psi_2} \), or might contain an epistemic state \( \Psi \) where \( \beliefsOf{\Psi} \) is inconsistent.

Second, because we are considering belief changes which depend upon prior belief changes, it is not sufficient to describe change operators for each epistemic state individually. 
In particular, a birds-eye perspective is required, because what a change operator does in one epistemic state depends on what it does in other epistemic states.
Consequently, we are modelling belief change operators for epistemic states as functions in the mathematical sense, describing globally, on all epistemic states from \( \setAllES \), how belief changes happen. The following definition describes this formally.
\begin{definition}\label{def:beliefchangeoperator}
{A \emph{belief change operator} (for epistemic states) over $ \setAllES $   is a function of type {$ \setAllES \times \propLang \to 
\setAllES $}.}
\end{definition}

For the rest of the article, we will always assume that belief change operators are given over some arbitrary but fixed chosen set of epistemic states \( \setAllES \), and we will  mention \( \setAllES \) only if it is necessary.
In the following, we will describe revision and contraction in this setting.
 
\subsection{Revision in Epistemic States}
\label{sec:prelim_revision}

Revision is the process of incorporating new beliefs into an agent's belief set, in a consistent way, whenever that is possible.
The postulates to revision given by Alchourr{\'{o}}n, G{\"{a}}rdenfors and Makinson  \cite{KS_AlchourronGaerdenforsMakinson1985} (AGM)
have counterparts in the framework of epistemic states, which were presented by Darwiche and Pearl \cite{KS_DarwichePearl1997}.

\begin{definition}[AGM Revision Operator for Epistemic States]
	A belief change operator \( \revision \) is called an \emph{AGM revision operator for epistemic states} if the following postulates are satisfied: %
	\begin{description}
	\item[\normalfont(\textlabel{R1}{pstl:R1})] 
	\( \alpha \in  \beliefsOf{\Psi \revision \alpha}  \)
\item[\normalfont(\textlabel{R2}{pstl:R2})] 
	\( \ksIF \beliefsOf{\Psi} + \alpha \text{ consistent, then }  \beliefsOf{\Psi \revision \alpha} = \Cn(\beliefsOf{\Psi} \cup \{\alpha\})  \)
\item[\normalfont(\textlabel{R3}{pstl:R3})] 
	\( \ksIF \alpha \text{ is consistent, then }  \beliefsOf{\Psi \revision \alpha} \text{ is consistent}  \)
\item[\normalfont(\textlabel{R4}{pstl:R4})] 
	\( \ksIF \alpha \equiv \beta \ksTHEN  \beliefsOf{\Psi \revision \alpha} = \beliefsOf{\Psi \revision \beta}  \)
\item[\normalfont(\textlabel{R5}{pstl:R5})] 
	\( \beliefsOf{\Psi \revision (\alpha\land\beta)} \subseteq \beliefsOf{\Psi \revision \alpha}+\beta \)
\item[\normalfont(\textlabel{R6}{pstl:R6})] 
	\( \ksIF \beliefsOf{\Psi\revision\alpha} + \beta \text{ is consistent, then }  \beliefsOf{\Psi \revision \alpha}+\beta \subseteq \beliefsOf{\Psi \revision (\alpha\land\beta)}  \)
	\end{description}
\end{definition}
The revision postulates \eqref{pstl:R1}--\eqref{pstl:R6} establish minimal change on beliefs when revising an epistemic state.

As shown by Katsuno and Mendelzon \cite{KS_KatsunoMendelzon1992} an AGM revision operator (in the classical AGM setting) is representable by a total preorder over the interpretations.
Darwiche and Pearl \cite{KS_DarwichePearl1997} adapt this approach to the framework of epistemic states, by equip each epistemic state $ \Psi $ with a total preorder $ \leq_{\Psi} $ over the worlds (interpretations) satisfying the so-called faithfulness condition. \pagebreak[3]
\begin{definition}[Faithful Assignment \cite{KS_DarwichePearl1997}]\label{def:faithful_assignment}
	A function $ \Psi\mapsto {\leq_\Psi} $ that maps each epistemic state to a total preorder on interpretations is said to be a faithful assignment if the following holds:
\begin{description}
    \item[\normalfont(\textlabel{FA1}{pstl:FA1})] \( \ksIF \omega_1 \in \modelsOf{\Psi} \ksAND \omega_2 \in \modelsOf{\Psi} \ksTHEN \omega_1 \simeq_\Psi \omega_2 \)
    \item[\normalfont(\textlabel{FA2}{pstl:FA2})] \( \ksIF \omega_1 \in \modelsOf{\Psi} \ksAND \omega_2\notin\modelsOf{\Psi} \ksTHEN \omega_1 <_\Psi \omega_2 \)
\end{description}
\end{definition}
Intuitively, the relation $ \leq_{\Psi} $ orders the worlds by plausibility, and the faithfulness condition guarantees that minimal worlds with respect to $ \leq_{\Psi} $ are the most plausible worlds, i.e., conformance of \( \modelsOf{\Psi} \) and \( \min(\Omega,\leq_\Psi) \), i.e., \( \modelsOf{\Psi}={\min(\Omega,\leq_\Psi)} \) if \( \beliefsOf{\Psi} \) is consistent.
For the case of an inconsistent belief set \( \beliefsOf{\Psi} \), faithfulness makes no assumptions about the minimal worlds of \( {\leq_\Psi} \).
To link assignments with belief revision operators, we use the following notion of compatibility \cite{KS_FalakhRudolphSauerwald2021}.
    \begin{definition}[Revision-Compatible%
        ]\label{def:revision-compatible}
        A faithful assignment \( \Psi \mapsto {\leq_{\Psi}} \) is called \emph{revision-compatible} with a belief change operator \( \revision \) if the following holds:
        \begin{equation}\tag{revision-compatible}\label{eq:repr_es_revision}
            \modelsOf{\Psi \revision \alpha} = \min(\modelsOf{\alpha},\leq_{\Psi})
        \end{equation}
\end{definition}

As discovered by Darwiche and Pearl, faithful assignments are a valid mean to characterise AGM revision operators for epistemic states.
\begin{proposition}[Characterization of AGM Revision Operators {\cite{KS_DarwichePearl1997}}]\label{prop:es_revision}
	A belief change operator $ \revision $ is an \emph{AGM revision operator for epistemic states} if and only if there is a faithful assignment $ \Psi\mapsto {\leq_\Psi} $ which is \ref{eq:repr_es_revision} with \( \revision \).
\end{proposition}	
Note that in the setting considered here, Proposition \ref{prop:es_revision} does not imply that epistemic states are equivalent (or isomorphic) to total preorders, i.e., the result \( \Psi \revision \alpha \) of a revision of \( \Psi \) by \( \alpha \) may not be completely determined by \( \leq_{\Psi} \) and \( \alpha \).
Due to Proposition \ref{prop:es_revision}, for each \( \alpha\in\propLang \) and for each \( \Psi_1,\Psi_2 \in \setAllES \) we have that \( {\leq_{\Psi_1}} = {\leq_{\Psi_2}} \) implies \( \beliefsOf{\Psi_1\revision\alpha}=\beliefsOf{\Psi_2\revision\alpha} \). 
However, this does not imply that the resulting epistemic states are the same, i.e., \( \Psi_1\revision\alpha\neq \Psi_2\revision\alpha \), and it could be the case\footnote{In the framework considered by Katsuno and Mendelzon for revision \cite{KS_KatsunoMendelzon1992}, this can not happen, because in their framework epistemic states are just belief sets, thus, \( \beliefsOf{\Psi_1\revision\alpha}=\beliefsOf{\Psi_2\revision\alpha} \) implies \( \Psi_1\revision\alpha=\Psi_2\revision\alpha \).} that subsequent revisions by some \( \alpha_1,\ldots,\alpha_n\in\propLang \) yields different belief sets, i.e., we could have that \( \beliefsOf{\Psi_1\revision\alpha\revision\alpha_1\revision\ldots\revision\alpha_n} \neq \beliefsOf{\Psi_2\revision\alpha\revision\alpha_1\revision\ldots\revision\alpha_n} \) holds.
The same observation was already implicitly made in \cite{KS_KoniecznyPinoPerez2008}.

Because we will stick to the Darwiche-Pearl framework for the rest of the article, we use \emph{AGM revision operator} as shorthand for AGM revision operator for epistemic states.
 
\subsection{Contraction in Epistemic States}
\label{sec:prelim_contraction}

Contraction is the process  of withdrawing beliefs.
Postulates for AGM contraction \cite{KS_AlchourronGaerdenforsMakinson1985} where first adapted by Caridroit, Konieczny and Marquis \cite{KS_CaridroitKoniecznyMarquis2015} to the setting of propositional logic.
Based on that reformulation for propositional logic of the AGM contraction postulates, Konieczny and Pino P{\'{e}}rez provided an adapted version of these postulates for the framework of epistemic states \cite{KS_KoniecznyPinoPerez2017}.
Independently, Chopra, Ghose, Meyer and  Wong \cite{KS_ChopraGhoseMeyerWong2008} gave a reformulation of AGM contraction for the framework of epistemic states, similar to the one used by Konieczny and Pino P{\'{e}}rez. 
There are subtle differences between these two similar, but slightly different groups of postulates\footnote{The subtle, but yet powerful difference is that Chopra, Ghose, Meyer and  Wong (CGMW) use a stronger syntax-independence postulate in comparison to the postulate \eqref{pstl:C5} by Konieczny and Pino P{\'{e}}rez (KPP).
        For an epistemic state \( \Psi \), the CGMW-postulate states that for equivalent formulas contraction on \( \Psi \) yields the same epistemic state, where the KPP-postulate \eqref{pstl:C5} states that for equivalent formulas contraction on \( \Psi \) yields the same belief set (but not necessarily the same epistemic state). 
        All other postulates are the same.
    Consequently, every belief change operator that satisfies the CGMW-postulates satisfies the KPP-postulates.
}.
Here we are using  the more general postulates by Konieczny and Pino P{\'{e}}rez adapted to our notation:
\pagebreak[3]
\begin{definition}[AGM Contraction Operator for Epistemic States]
		A belief change operator \( \contraction \) is called an \emph{AGM contraction operator for epistemic states} if the following postulates are satisfied: %
	\begin{description}
		\item[\normalfont(\textlabel{C1}{pstl:C1})] \( \beliefsOf{\Psi \contraction \alpha} \subseteq \beliefsOf{\Psi}  \) 
		
		\item[\normalfont(\textlabel{C2}{pstl:C2})] \( \ksIF  \alpha\notin\beliefsOf{\Psi} \ksTHEN \beliefsOf{\Psi}\subseteq \beliefsOf{\Psi \contraction\alpha} \) 
		
		\item[\normalfont(\textlabel{C3}{pstl:C3})] \( \ksIF \alpha \not\equiv \top \ksTHEN  \alpha \notin \beliefsOf{\Psi  \contraction  \alpha} \) 
		
		\item[\normalfont(\textlabel{C4}{pstl:C4})] \( \beliefsOf{\Psi} \subseteq Cn(\beliefsOf{\Psi  \contraction  \alpha} \cup \{\alpha\}) \) 
		
		\item[\normalfont(\textlabel{C5}{pstl:C5})] \( \ksIF \alpha \equiv\beta \ksTHEN \beliefsOf{\Psi  \contraction  \alpha} = \beliefsOf{\Psi  \contraction  \beta} \) 
		
		\item[\normalfont(\textlabel{C6}{pstl:C6})] \( \beliefsOf{\Psi \contraction \alpha} \cap \beliefsOf{\Psi \contraction \beta} \subseteq \beliefsOf{\Psi  \contraction  (\alpha\land\beta)} \) 
		
		\item[\normalfont(\textlabel{C7}{pstl:C7})] \( \ksIF \beta\notin\beliefsOf{\Psi   \contraction  (\alpha\land\beta)}  \ksTHEN \beliefsOf{\Psi  \contraction  (\alpha\land\beta)} \subseteq \beliefsOf{\Psi \contraction \beta} \) 
	\end{description}
\end{definition}
For an explanation of these postulates we refer to the article by Caridroit et al. \cite{KS_CaridroitKoniecznyMarquis2015}.
Note that the postulates \eqref{pstl:C1}--\eqref{pstl:C7} do not explicitly state how one should maintain the contraction strategy in the case of iteration. 
Like for revisions, we link contraction operators and faithful assignments by a notion of compatibility (c.f. Definition \ref{def:revision-compatible}).
\begin{definition}[contraction-compatibility]
A faithful assignment \( \Psi \mapsto {\leq_{\Psi}} \) is called \emph{contraction-compatible} with a belief change operator \( \contraction \) if the following holds:
			\begin{equation}\tag{contraction-compatible}\label{eq:repr_es_contraction}
			\modelsOfES{\Psi \contraction \alpha} = \modelsOfES{\Psi} \cup \min(\modelsOf{\negOf{\alpha}},\leq_{\Psi})
		\end{equation}
\end{definition}
\noindent A characterization in terms of total preorders on epistemic states is given by the following proposition.
\begin{proposition}[Characterization of AGM Contraction Operators {\cite{KS_KoniecznyPinoPerez2017}}]\label{prop:es_contraction}
	A belief change operator $ \contraction $ fulfils the postulates \eqref{pstl:C1}--\eqref{pstl:C7} if and only if there is a faithful assignment $ \Psi\mapsto {\leq_\Psi} $ \ref{eq:repr_es_contraction} with \( \contraction \).
\end{proposition}
 Because we will only consider belief change operators in the sense of Definition \ref{def:beliefchangeoperator}, for the purpose of this article we use \emph{AGM contraction operator} as shorthand for \emph{AGM contraction operator for epistemic states}.
 
\section{Principles for Iterated Revision and Conditionals}
\label{sec:conditionals_revision}
In this section, we present several iteration principles for revision.
The key idea is to connect belief revision with conditional beliefs, yielding a semantics for conditionals by change operators. Then, we use this connection to consider iteration principles for revision operators from different perspectives.

\subsection{Conditionals and the Ramsey-Test for Revisions}\label{sec:conditionals_revision_conditionals}
For the matter of this article, a conditional is a syntactic object \( \ramseyCond{\beta}{\alpha} \), where \( \alpha,\beta\in\propLang \).
A typical informal interpretation of a conditional \( \ramseyCond{\beta}{\alpha} \) is that \( \ramseyCond{\beta}{\alpha} \) stands for \enquote{if $ \alpha $, then usually $ \beta $}, establishing conditionals as fundamental objects in non-monotonic reasoning.
The so-called Ramsey test \cite{KS_Stalnaker1968} provides an alternative semantics for conditionals by stating that one should believe \( \ramseyCond{\beta}{\alpha} \) if and only if one believes \( \beta \) after revision by \( \alpha \).
We use the Ramsey test here in the following formulation:
\begin{description}
	\item[\normalfont(\textlabel{RT}{ramseytest})] \( \Psi \models \ramseyCond{\beta}{\alpha} \ksIFFlong \Psi * \alpha \models \beta \) 
\end{description}
One application of \eqref{ramseytest} in the setting of iterated belief change is that \eqref{ramseytest} makes it possible to use conditionals as a syntactic means to describe (partially) how a belief revision operator \( \revision \) behaves with respect to a particular epistemic state. More precisely, given the set \( C(\Psi) \) of all conditionals \( \ramseyCond{\beta}{\alpha} \) with \( \Psi\models \ramseyCond{\beta}{\alpha} \), in the presence of \eqref{ramseytest}, we can determine \( \beliefsOf{\Psi\revision\alpha} \) just from \( C(\Psi) \) and \( \alpha \).
Note that in general we can  not determine \( \Psi\revision\alpha \) just from set of conditionals \( C(\Psi) \) and \( \alpha \).

When considering an AGM revision operator \( \revision \), we obtain from Proposition \ref{prop:es_revision} that \( \leq_{\Psi} \) has similar properties as  \( C(\Psi) \), i.e., for each epistemic state \( \Psi \), the total preorder \( \leq_{\Psi} \) and \( \alpha \) determine \( \beliefsOf{\Psi\revision\alpha} \) entirely. 
This gives rationale to consider \( \leq_{\Psi} \) as a semantic counter-part of \( C(\Psi) \), which is indeed given by the following.
First, we formally introduce notions of acceptance of a conditional by an epistemic state and by a total preorder.
\begin{definition}
    Let  \( \Psi \) be an epistemic state and let \( \leq \) be a total preorder over $ \Omega $. 
    We say a conditional $ \ramseyCond{\beta}{\alpha} $ is \emph{accepted by \( \Psi \) (and \( \revision \))}, written $ \Psi \models \ramseyCond{\beta}{\alpha} $, if $ \Psi \revision \alpha \models \beta  $.
    Moreover, we say a conditional $ \ramseyCond{\beta}{\alpha} $ is \emph{accepted by $ {\leq} $} if for each $ \omega \in \modelsOf{\alpha\land\negOf{\beta}}  $ there exists $ \omega' \in \modelsOf{\alpha\land\beta} $ such that $ \omega' < \omega $.
\end{definition}
The definition of  $ \Psi \models \ramseyCond{\beta}{\alpha} $ is just the right-hand side of \eqref{ramseytest}.
Clearly, acceptance of conditional $ \ramseyCond{\beta}{\alpha} $ in \( \Psi \) depends on $ * $, but we write $ \Psi \models \ramseyCond{\beta}{\alpha} $ instead of $ \Psi \models^* \ramseyCond{\beta}{\alpha} $, omitting the superscript on \( \models^* \), since \( \revision \) will be always clear by the context.
Acceptance by a total preorder is chosen such that it is compatible with revision-compatibility (cf. Definition \ref{def:revision-compatible}).
In particular, this can be lifted to formulas (cf. Section \ref{sec:prelim}), the acceptance of $ \ramseyCond{\beta}{\alpha} $ by $ {\leq} $ is equivalent to having \( \alpha\land\beta <_\Psi \alpha\land\negOf{\beta} \).
The following proposition presents formally the correspondence between the acceptance by \( \Psi \) and acceptance by \( \leq_{\Psi} \).
\begin{proposition}\label{prop:revision_acceptance}
	Let $ * $ be an AGM revision operator and $ \Psi\mapsto{\leq_{\Psi}} $ be a faithful assignment \ref{eq:repr_es_revision} with \( \revision \). 
    For each \( \Psi\in\setAllES \) and \( \alpha,\beta\in\propLang \)
	the following statements are equivalent:
    \begin{enumerate}[(a)]
        \item  $ \Psi * \alpha \models \beta $
        \item  $  \ramseyCond{\beta}{\alpha} $ is accepted by $ \leq_{\Psi} $
        \item  $  \ramseyCond{\beta}{\alpha} $ is accepted by $ \Psi $ and \( \revision \)
    \end{enumerate}
\end{proposition}
\begin{proof}
    The equivalence of (a) and (c) is given by definition.
    From Proposition \ref{prop:es_revision} we obtain \( \beta\in\beliefsOf{\Psi\revision\alpha} \) if and only if \( \min(\modelsOf{\alpha},\leq_{\Psi}) \subseteq \modelsOf{\beta} \).
    The latter is exactly the case if and only if $  \ramseyCond{\beta}{\alpha} $ is accepted by $ \leq_{\Psi} $. We obtain that (a) and (b) are equivalent.
\end{proof}

Next, we present iteration principles for revision operators.

\subsection{Postulates for Iterated Revision}\label{sec:conditionals_revision_principle}

In this section, %
    we will present iteration principles for revision. We consider different postulates for these principles from the viewpoint of changing beliefs, of changing conditional beliefs, and of changing relations of worlds, and establish equivalences between them for AGM revision operators.
    
    \smallskip
    \emph{Note on the naming-pattern for postulates.} In the course of this paper, we consider several counter-parts for these principles for contraction. Because of the number of postulates and to underline the direct correspondence between revision and contraction postulates, we will use the following naming scheme. Postulates for \emph{iterated revision} are denoted by IR; postulates for \emph{iterated contraction} will be abbreviated by IC.
Moreover, we will consider postulates for the change of beliefs (formulas), change of conditional beliefs, and the change of relations. 
To highlight the difference, we denote iteration postulates for revision with a \emph{conditional} viewpoint by IR\( {}^\mathrm{cond} \) and IR\( {}^\mathrm{rel} \) stands for considering a \emph{relational} point of view. When considering change of beliefs, we denote them by IR without superscript.
\paragraph*{Minimal change of conditional beliefs}
As a first principle for revision, we consider minimal change of conditional beliefs, which is given by the following postulates\footnote{The subscript \( \mathrm{min} \) stands for \emph{minimization} of change in conditional beliefs.}:
\begin{itemize}
    \item[](\textlabel{IR${}^\mathrm{cond}_\mathrm{min}$}{pstl:CBcond}) \( \ksIF \Psi\revision\alpha \models \negOf{\beta} \ksTHEN  \Psi\revision\alpha \models \ramseyCond{\gamma}{\beta} \Leftrightarrow \Psi \models \ramseyCond{\gamma}{\beta} \)
    \item[](\textlabel{IR${}_\mathrm{min}$}{pstl:CB}) \( \ksIF \Psi\revision\alpha\models\negOf{\beta} \ksTHEN \beliefsOf{\Psi\revision\alpha\revision\beta} = \beliefsOf{\Psi\revision\beta} \) 
    \item[]{\normalfont(\textlabel{IR${}^\mathrm{rel}_\mathrm{min}$}{pstl:CBR})} \( \ksIF\omega_1,\omega_2 \notin \modelsOf{\Psi\revision\alpha} \ksTHEN \omega_1 \leq_{\Psi} \omega_2 \Leftrightarrow \omega_1 \leq_{\Psi \revision \alpha} \omega_2 \) 
\end{itemize}
Craig Boutilier proposed that minimality of change should be applied also to change of conditional beliefs \cite{KS_Boutilier1996}, leading to the principle
\eqref{pstl:CBcond},
which states that revision by \( \alpha \) does not influence conditional beliefs for which the antecedence is inconsistent with the result of revision by \( \alpha \).
AGM revision operators that satisfy \eqref{pstl:CBcond} are sometimes also  denoted as \emph{natural revision} \cite{KS_BoothMeyer2006}.
From Proposition \ref{prop:revision_acceptance} we obtain \eqref{pstl:CB}, which is a counter-part to \eqref{pstl:CBcond}, which was already given in the literature \cite{KS_DarwichePearl1997}.
By Darwiche and Pearl, absolute minimization of changes in conditional beliefs is not desirable for iterative belief revision \cite{KS_DarwichePearl1997}, as it prevents evolution of the change strategy itself.
The following proposition shows that when revising by \( \alpha \), satisfaction of \eqref{pstl:CBcond}  is equivalent to having that \( \leq_{\Psi * \alpha} \) changes only minimally with respect to \( \leq_{\Psi} \).
\begin{proposition}[\cite{KS_DarwichePearl1997}]
    Let \( \revision \) be an AGM revision operator and let \( \Psi \mapsto {\leq_{\Psi}} \) be a faithful assignment \ref{eq:repr_es_revision} with \( \revision \).
    The operator \( \revision \) satisfies \eqref{pstl:CBcond} if and only if 
\eqref{pstl:CBR}
    is satisfied.
\end{proposition}

\paragraph*{Cautious preservation of conditional beliefs}
While absolute minimization of changes in conditional beliefs is undesirable, arbitrary changes of conditional beliefs are likewise undesirable.
Therefore, Darwiche and Pearl propose the following 
iteration postulates\footnote{These principles are often denoted as (DP1)--(DP4), referring to Darwiche and Pearl.%
} for revision  \cite{KS_DarwichePearl1997}:
\begin{description}
	\item[\normalfont(\textlabel{IR1\( {}^\mathrm{cond} \)}{pstl:DP1cond})] \( \ksIF \beta\models\alpha \ksTHEN \Psi \models \ramseyCond{\gamma}{\beta} \Leftrightarrow \Psi * \alpha \models \ramseyCond{\gamma}{\beta} \) 
	\smallskip
	\item[\normalfont(\textlabel{IR2\( {}^\mathrm{cond} \)}{pstl:DP2cond})] \( \ksIF \beta\models\negOf{\alpha} \ksTHEN \Psi \models \ramseyCond{\gamma}{\beta} \Leftrightarrow \Psi * \alpha \models \ramseyCond{\gamma}{\beta} \) 
	\smallskip
	\item[\normalfont(\textlabel{IR3\( {}^\mathrm{cond} \)}{pstl:DP3cond})] \( \ksIF \Psi \models \ramseyCond{\alpha}{\beta} \ksTHEN \Psi * \alpha \models \ramseyCond{\alpha}{\beta}  \) 
	\smallskip
	\item[\normalfont(\textlabel{IR4\( {}^\mathrm{cond} \)}{pstl:DP4cond})] \( \ksIF \Psi \not\models \ramseyCond{\negOf{\alpha}}{\beta} \ksTHEN \Psi * \alpha \not\models \ramseyCond{\negOf{\alpha}}{\beta}  \) 
\end{description}
By the Ramsey test \eqref{ramseytest}, respectively by Proposition \ref{prop:revision_acceptance}, one obtains the following reformulation  of the Darwiche-Pearl postulates  in terms of two-step changes, which were also given already by Darwiche and Pearl \cite{KS_DarwichePearl1997}:
\begin{description}
	\item[\normalfont\normalfont(\textlabel{IR1}{pstl:DP1})] \( \ksIF \beta\models\alpha \ksTHEN \beliefsOf{\Psi * \alpha * \beta} = \beliefsOf{\Psi * \beta} \) 
	\smallskip
	\item[\normalfont\normalfont(\textlabel{IR2}{pstl:DP2})] \( \ksIF \beta\models\negOf{\alpha} \ksTHEN \beliefsOf{\Psi * \alpha * \beta} = \beliefsOf{\Psi * \beta} \) 
	\smallskip
	\item[\normalfont\normalfont(\textlabel{IR3}{pstl:DP3})] \( \ksIF \Psi\revision\beta \models \alpha \ksTHEN \Psi\revision\alpha\revision\beta \models \alpha  \) 
	\smallskip
	\item[\normalfont\normalfont(\textlabel{IR4}{pstl:DP4})] \( \ksIF \Psi\revision\beta \not\models \negOf{\alpha} \ksTHEN \Psi\revision\alpha\revision\beta \not\models \negOf{\alpha}  \) 
\end{description}
Intuitively, \eqref{pstl:DP1} and \eqref{pstl:DP2} establish that revision by a more general belief \( \alpha \), respectively \( \negOf{\alpha} \), beforehand does not influences revision by a more specific belief \( \beta \).
The postulates \eqref{pstl:DP3} and \eqref{pstl:DP4} together state that revision by \( \alpha \) does not influence the credibility of \( \alpha \) with respect to subsequent changes.

It is well-known that these operators can be characterized in the semantic framework of total preorders.
\begin{proposition}[Iterated Revision{\cite{KS_DarwichePearl1997}}]\label{prop:it_es_revision}
	Let $ * $ be an AGM revision operator. Then $ * $ satisfies \eqref{pstl:DP1} to \eqref{pstl:DP4} if and only there exists a faithful assignment $ \Psi\mapsto{\leq_{\Psi}} $ \ref{eq:repr_es_revision} with \( \revision \) such that the following postulates\footnote{These principles were denoted as (CR1)--(CR4) by Darwiche and Pearl \cite{KS_DarwichePearl1997}.}
     are satisfied:
	\begingroup
	\begin{description}
		\item[\normalfont(\textlabel{IR1\( {}^\mathrm{rel} \)}{pstl:RR8})] \( \ksIF \omega_1,\omega_2 \in \modelsOf{\alpha} \ksTHEN \omega_1 \leq_{\Psi} \omega_2 \Leftrightarrow \omega_1 \leq_{\Psi * \alpha} \omega_2 \) 
		\smallskip
		\item[\normalfont(\textlabel{IR2\( {}^\mathrm{rel} \)}{pstl:RR9})] \( \ksIF \omega_1,\omega_2 \in \modelsOf{\negOf{\alpha}} \ksTHEN \omega_1 \leq_{\Psi} \omega_2 \Leftrightarrow \omega_1 \leq_{\Psi * \alpha} \omega_2 \)
		\smallskip
		\item[\normalfont(\textlabel{IR3\( {}^\mathrm{rel} \)}{pstl:RR10})] \( \ksIF \omega_1 \in \modelsOf{\alpha} \ksAND \omega_2 \in \modelsOf{\negOf{\alpha}}  \ksTHEN    \omega_1 <_{\Psi} \omega_2  \Rightarrow  \omega_1 <_{\Psi * \alpha} \omega_2 \)
		\smallskip
		\item[\normalfont(\textlabel{IR4\( {}^\mathrm{rel} \)}{pstl:RR11})] \( \ksIF \omega_1  \in \modelsOf{\alpha} \ksAND \omega_2 \in \modelsOf{\negOf{\alpha}}  \ksTHEN    \omega_1 \leq_{\Psi} \omega_2  \Rightarrow  \omega_1 \leq_{\Psi * \alpha} \omega_2 \)
	\end{description}
	\endgroup
\end{proposition}%

In addition to Proposition \ref{prop:it_es_revision} we like to remark that there is a direct correspondence between the postulates considered in this section. An AGM revision operator satisfies \eqref{pstl:DP1} if and only if it satisfies \eqref{pstl:DP1cond} if and only if it satisfies \eqref{pstl:RR8}.
The same holds for \eqref{pstl:DP2}, \eqref{pstl:DP2cond} and \eqref{pstl:RR9}, and so on for the others postulates.
An advantage of these postulates is that they are very intuitive, regarding which of the three perspectives given here is considered.
Most of the Darwiche-Pearl postulates for revision are widely accepted and have wide applications. 
However, the postulate \eqref{pstl:DP2} has sometimes been criticized \cite{KS_Rott2003}.

\paragraph*{The independence principle}
We consider independence, which is given by the following postulates\footnote{ind stands for \emph{independence}.}:
\begin{description}
    \item[\normalfont(\textlabel{IR\( {}_\mathrm{ind} \)}{pstl:Ind})] \( \ksIF \Psi\revision\beta\not\models \negOf{\alpha} \ksTHEN \Psi\revision\alpha\revision\beta\models\alpha   \) 
    \item[\normalfont(\textlabel{IR\( {}^\mathrm{cond}_\mathrm{ind} \)}{pstl:IndCond})] \( \ksIF \Psi\revision\beta\not\models \negOf{\alpha} \ksTHEN \Psi\revision\alpha \models \ramseyCond{\alpha}{\beta}   \) 
    \item[\normalfont\normalfont(\textlabel{IR\( {}^\mathrm{rel}_\mathrm{ind} \)}{pstl:IndR})] \( \ksIF \omega_1  \in \modelsOf{\alpha} \ksAND \omega_2 \in \modelsOf{\negOf{\alpha}}  \ksTHEN    \omega_1 \leq_{\Psi} \omega_2  \Rightarrow  \omega_1 <_{\Psi * \alpha} \omega_2 \) 
\end{description}
The so-called \emph{independence} condition was introduced by Jin and Thielscher to avoid unwanted dependencies between beliefs in the context of revision \cite{KS_JinThielscher2007}.
Due to Proposition \ref{prop:revision_acceptance} we obtain easily that \eqref{pstl:IndCond} is an analogue to \eqref{pstl:Ind}. %
It has been shown \cite{KS_BoothMeyer2006} that in the framework we consider here, \eqref{pstl:Ind} is equivalent to \eqref{pstl:IndR}
whenever \( \revision \) is an AGM revision operator, which reveals that independence is a stronger version of \eqref{pstl:DP4}.
AGM revision operators that satisfy \eqref{pstl:DP1}, \eqref{pstl:DP2} and \eqref{pstl:Ind} are called also \emph{admissible} revision operators \cite{KS_BoothMeyer2006}.

\paragraph*{Lexicographic revision} As last principle, we consider lexicographic revisions which are given by \eqref{pstl:DP1}, \eqref{pstl:DP2} and by one of the following postulates (which are all equivalent)\footnote{\emph{lex} stands for \emph{lexicographic}.}:
\begin{description}
    \item[\normalfont(\textlabel{IR\( {}_\mathrm{lex} \)}{pstl:Lex})] \( \ksIF \beta\not\models\negOf{\alpha} \ksTHEN \Psi\revision\alpha\revision\beta\models\alpha   \) 
    \item[\normalfont(\textlabel{IR\( {}^\mathrm{lex}_\mathrm{ind} \)}{pstl:LexCond})] \( \ksIF \Psi\revision\beta\not\models \negOf{\alpha} \ksTHEN \Psi\revision\alpha \models \ramseyCond{\alpha}{\beta}   \) 
    \item[\normalfont\normalfont(\textlabel{IR\( {}^\mathrm{rel}_\mathrm{lex} \)}{pstl:LexR})] \( \ksIF \omega_1  \in \modelsOf{\alpha} \ksAND \omega_2 \in \modelsOf{\negOf{\alpha}}  \ksTHEN    \omega_1 \leq_{\Psi} \omega_2  \Rightarrow  \omega_1 <_{\Psi * \alpha} \omega_2 \) 
\end{description}
Lexicographic revision was introduced by Nayak as an approach to revision which gives higher priority to beliefs compatible with new information than to other beliefs, while staying sensitive to the priorities given by the revision history \cite{KS_Nayak1994}. 
We present here the so-called simple version \cite{KS_NayakPagnuccoPeppas2003}.
This approach has been identified as yields the most conservative admissible revision operators \cite{KS_BoothMeyer2006}.
Equivalence of \eqref{pstl:Lex} and \eqref{pstl:LexCond} is given by Proposition \ref{prop:revision_acceptance} and the equivalence between \eqref{pstl:Lex} and \eqref{pstl:LexR} has been shown by Booth and Meyer~\cite{KS_BoothMeyer2006}.

\section{Iterated Contraction and Conditionals}
\label{sec:contractionsals}

In this section we prepare the ground for our investigations on iteration principles for contraction.
Therefore, we will use AGM contraction operators to provide a semantics for conditionals in the same fashion as revision operators provide a semantics of conditionals via the Ramsey-test (cf. Section \ref{sec:conditionals_revision_conditionals}).
The approach we follow is that we replace the revision operator in \eqref{ramseytest} by a contraction operator.
To distinguish conditionals which have a semantics by revision operators, sometimes called \emph{RT-conditionals} in the sequel, from conditionals which use contraction operators as semantics, we will call the latter \emph{contractionals} \cite{KS_Bochman2001} and denote them by $ \contractionCond{\beta}{\alpha} $.
We obtain the following reformulation of \eqref{ramseytest} for contractionals:
\begingroup%
\begin{description}
	\item[\normalfont(\textlabel{Contractional}{contractionConditional})] \(  \Psi \models \contractionCond{\beta}{\alpha} \ksIFFlong \Psi \contraction \alpha \models \beta \) 
\end{description}
\endgroup

For the rest of this article we assume that \( \contraction \) is always an AGM contraction operator,
and given this class of operators, we will investigate the properties of contractionals.
We start with some intuition for contractionals
by suggesting to read a {contractional} $ \contractionCond{\beta}{\alpha} $ as:%
\begin{center}
	\emph{believe $ \beta $ even in the absence of $ \alpha $}.
\end{center}
More specifically, a contractional $ \contractionCond{\beta}{\alpha} $ encodes two aspects: (1) \( \beta \) is believed, and (2) believing of \( \beta \) is independent of believing in \( \alpha \).
The rationale for this interpretation is given by the AGM contraction postulates \eqref{pstl:C1}--\eqref{pstl:C7}.
In particular, the rationale for (1) is because \( \contraction \) satisfies inclusion \eqref{pstl:C1}, by which a contraction never introduces new beliefs. 
Thus, it is unavoidable that \( \Psi \models \contractionCond{\beta}{\alpha} \) holds only if \( \beta\in\beliefsOf{\Psi} \).
Aspect (2) is given by the success condition \eqref{pstl:C3}, encoding the overall goal of contraction by \( \alpha \) to remove all beliefs that imply \( \alpha \).
Consequently, if \( \Psi \models \contractionCond{\beta}{\alpha} \), then \( \beta \) is a belief that is unrelated to \( \alpha \) in the sense that believing in \( \beta \) gives no rationale to believe in \( \alpha \).

Consider the following example for a comparison of the meaning of contractionals to Ramsey test conditionals.
\begin{example}
	Let $ f $ have the intended meaning that something is \enquote{able to fly} and $ p $ the intended meaning that something is a \enquote{penguin}.
	Then the acceptance of a (Ramsey test) conditional $ \ramseyCond{\neg f}{p} $ states that if the agent 
	learns
	that something is a penguin, she will believe that it is not able to fly.
	In contrast, the acceptance of a contractional $ \contractionCond{\neg f}{p} $ states that the agent keeps the belief that something is not able to fly, even if the agent 
	gives up her belief that it is a penguin.
\end{example}

We continue by investigating contractionals more formally.
Like in the previous section, we define a notion of acceptance of a contractional by an epistemic state \( \Psi \) and a notion of acceptance by a total preorder over worlds.
\begin{definition}
	Let  \( \Psi \) be an epistemic state and let \( \preceq \) be a total preorder over $ \Omega $. 
	We say a contractional $ \contractionCond{\beta}{\alpha} $ is \emph{accepted by \( \Psi \)}, written $ \Psi \models \contractionCond{\beta}{\alpha} $, if $ \Psi \contraction \alpha \models \beta  $.
    Moreover, we say a contractional $ \contractionCond{\beta}{\alpha} $ is \emph{accepted by $ {\leq} $} if $ {\min(\Omega,\leq)} \subseteq {\modelsOf{\beta}} $ and for every $ \omega_1 \in \modelsOf{\negOf{\alpha}\land\negOf{\beta}}  $ there exists $ \omega_2 \in \modelsOf{\negOf{\alpha}\land\beta} $ such that $ \omega_2 \leq \omega_1 $ and \( \omega_1 \not\leq \omega_2 \).
\end{definition}

Note that the notion acceptance of $ \contractionCond{\beta}{\alpha} $ by a total preorder is chosen to be compliant with the notion of contraction-compatibility used in Proposition \ref{prop:es_contraction}.
Analogously to Proposition \ref{prop:revision_acceptance}, we obtain that acceptance by an epistemic state and acceptance by a total preorder of a contractional are equivalent. 
\begin{proposition}\label{prop:contractional_acceptance}
    Let $ \contraction $ be an AGM contraction operator and let $ \Psi\mapsto{\leq_{\Psi}} $ be a faithful assignment \ref{eq:repr_es_contraction} with \( \contraction \).
    For each \( \Psi\in\setAllES \) and \( \alpha,\beta\in\propLang \) the following statements are equivalent:
    \begin{enumerate}[(a)]
        \item  $ \Psi \contraction \alpha \models \beta $
        \item  $  \contractionCond{\beta}{\alpha} $ is accepted by $ \leq_{\Psi} $
        \item  $  \contractionCond{\beta}{\alpha} $ is accepted by $ \Psi $ and \( \contraction \)
    \end{enumerate}
\end{proposition}
\begin{proof}
The equivalence of (a) and (c) is given by definition.
From Proposition \ref{prop:es_contraction}, we obtain \( \beta\in\beliefsOf{\Psi\revision\alpha} \) if and only if \( \modelsOf{\Psi} \cup \min(\modelsOf{\alpha},\leq_{\Psi}) \subseteq \modelsOf{\beta} \).
The latter is exactly the case if and only if $  \ramseyCond{\beta}{\alpha} $ is accepted by $ \leq_{\Psi} $. We obtain that (a) and (b) are equivalent.
\end{proof}

As a first application of Proposition \ref{prop:contractional_acceptance}, we consider the inclusion postulate \eqref{pstl:C1} from the set of basic postulates for AGM contraction, stating that a contraction does not yield new beliefs. %
The following property is mainly a consequence of the inclusion property.

\begin{proposition}\label{prop:inclusion_direct_consequence}
	Let \( - \) be a belief change operator satisfying \eqref{pstl:C1} and \eqref{pstl:C4}.
    If \( \alpha,\beta,\gamma\in\propLang \), then \( \gamma \in \beliefsOf{\Psi - \alpha - \beta}   \) implies \( \gamma \in \beliefsOf{\Psi - \top}  \).
\end{proposition}
\begin{proof}
    Because \( - \) satisfies \eqref{pstl:C1}, we obtain \( \gamma \in \beliefsOf{\Psi}   \) from \( \gamma \in \beliefsOf{\Psi - \alpha - \beta}   \).
    Now note that \( \top\in\beliefsOf{\Psi - \top}  \), because \( \top \) is element of every deductively closed set.
    Consequently, we have  \( Cn(\beliefsOf{\Psi - \top}\cup\{\top\})=\beliefsOf{\Psi - \top} \).
    From this last observation and \eqref{pstl:C4} we obtain \( \beliefsOf{\Psi} \subseteq \beliefsOf{\Psi - \top}  \).
\end{proof}

By employing Proposition \ref{prop:contractional_acceptance}, we obtain from Proposition \ref{prop:inclusion_direct_consequence} that every AGM contraction operator satisfies the following:
\begin{equation*}
	\ksIF \alpha,\beta,\gamma\in\propLang \ksTHEN \Psi \contraction \alpha \models \contractionCond{\gamma}{\beta}  \Rightarrow \Psi \models \contractionCond{\gamma}{\top}
\end{equation*}

To get an impression what a contractional encodes with respect to a precedence on formulas, consider the following consequence of Proposition \ref{prop:contractional_acceptance}.
\begin{corollary}\label{col:contraction_formula}
Let $ \Psi\mapsto{\leq_{\Psi}} $ be a faithful assignment \ref{eq:repr_es_contraction} with an AGM contraction operator \( \contraction \).
	Then $ \contractionCond{\beta}{\alpha} $ is accepted by \( \Psi \) if $ \beta <_{\Psi} \negOf{\beta} $ and $ \negOf{\alpha}\land \beta <_{\Psi} \negOf{\alpha}\land\negOf{\beta} $.
\end{corollary}
Corollary \ref{col:contraction_formula} highlights that a contractional is not simple replaceable by a single Ramsey test conditional. 
More insights on relation between formulas and iterated contraction are given in the literature, e.g., \cite{KS_RamachandranNayakOrgun2012}.

In the following sections we investigate and evaluate iteration principles for AGM contraction operators.
Therefore, we consider mainly two groups of iteration principles which are analogues to the Darwiche-Pearl iteration principles for AGM revision operators (cf. Section \ref{sec:conditionals_revision_principle}).
The first group are counter-parts for AGM contraction operators to the syntactic versions  \eqref{pstl:DP1}--\eqref{pstl:DP4}, respectively \eqref{pstl:DP1cond}--\eqref{pstl:DP4cond}, of the Darwiche-Pearl iteration principles.
Secondly, we consider counter-parts  for AGM contraction operators to the semantic versions \eqref{pstl:RR8}--\eqref{pstl:RR11} of the Darwiche-Pearl iteration principles.
For both groups of principles we consider the impact of conditional beliefs, i.e., how acceptance of contractions evolves in the process of contraction.
Finally, we consider conditional beliefs in the context of concrete strategies for AGM contraction operators. 
In particular, we consider minimization in the change of conditional beliefs.

\section{Syntactic Analogues to \eqref{pstl:DP1}--\eqref{pstl:DP4} for Contraction}
\label{sec:syn_postulates}

\begin{figure}[t]
   \begin{center}
       \begin{tabular}{rlll}
            \toprule
            \multirow{3}{0.3cm}{\rotatebox{90}{Revision}}&RT-Conditional & \eqref{pstl:DP1cond}   & \( \ksIF \beta\models\alpha \ksTHEN \Psi \models \ramseyCond{\gamma}{\beta} \Leftrightarrow \Psi * \alpha \models \ramseyCond{\gamma}{\beta} \)  \\[0.2em] 
            &Belief & \eqref{pstl:DP1}   & \( \ksIF \beta\models\alpha \ksTHEN \beliefsOf{\Psi * \alpha * \beta} = \beliefsOf{\Psi * \beta} \)  \\[0.2em] 
            &Relational & \eqref{pstl:RR8}   & \( \ksIF \omega_1,\omega_2 \in \modelsOf{\alpha} \ksTHEN \omega_1 \!\leq_{\Psi}\! \omega_2 \Leftrightarrow \omega_1 \!\leq_{\Psi * \alpha}\! \omega_2 \)   \\[0.2em]  \midrule
            
            \multirow{15}{0.3cm}{\rotatebox{90}{Contraction}}&Contractional & \eqref{pstl:Prop1}   & \( \ksIF \neg\alpha\models\beta \ksTHEN \Psi \contraction \alpha \models \contractionCond{\gamma}{\beta}  \Leftrightarrow \Psi \models \contractionCond{\gamma}{\beta} \) \\[0.2em]
            &Belief     &  \eqref{pstl:C8prop}                    & \( \ksIF \neg\alpha\models\beta \ksTHEN \beliefsOf{\Psi \contraction \alpha \contraction \beta}    = \beliefsOf{\Psi \contraction \beta} \)                               \\[0.2em]
            &Relational      &                      & no counterpart (Proposition \ref{prop:nonexist_prop1})                                                                                                                    \\[0.2em] \cmidrule{2-4}
            
            &Contractional & \eqref{pstl:C1left}   & \( \ksIF \neg\alpha\models\beta \ksTHEN \Psi  \models \contractionCond{\gamma}{\beta}  \Rightarrow \Psi \contraction \alpha \models \contractionCond{\gamma}{\beta} \) \\[0.2em]
            &Belief     &                       & \( \ksIF \neg\alpha\models\beta \ksTHEN \Psi  \contraction\beta \models \gamma  \Rightarrow \Psi \contraction \alpha \contraction\beta \models \gamma \)                                                                                                                                                                 \\[0.2em]
            &Semantic      &                       & no counterpart (Proposition \ref{prop:nonexist_prop1})                                                                                                                 \\[0.2em] \cmidrule{2-4}
            &Contractional & \eqref{pstl:C1right}  & \( \ksIF \neg\alpha\models\beta \ksTHEN \Psi \contraction \alpha \models \contractionCond{\gamma}{\beta}  \Rightarrow \Psi \models \contractionCond{\gamma}{\beta} \)  \\[0.2em]
            &Belief     &                       & \( \ksIF \neg\alpha\models\beta \ksTHEN \Psi \contraction \alpha \contraction\beta \models \gamma  \Rightarrow \Psi \models\contraction\beta \models \gamma \)                                                                                                                                                                 \\[0.2em]
            &Semantic      & \eqref{pstl:CR1right} & \( \ksIF \omega_1,\omega_2 \in \modelsOf{\alpha} \ksTHEN \omega_1 \leq_{\Psi} \omega_2 \Rightarrow \omega_1 \leq_{\Psi\contraction\alpha} \omega_2 \)                  \\[0.2em] \cmidrule{2-4}
            &Contractional   & \eqref{pstl:C8cond}     & \( \ksIF \negOf{\alpha}\models\beta    \ksTHEN  \Psi\!\contraction\!\alpha \!\models\! \contractionCond{\alpha\to\gamma}{\beta}   \!\Leftrightarrow\! \Psi \!\models\! \contractionCond{\alpha\to\gamma}{\beta}   \)  \\[0.2em]
            &Belief       & \eqref{pstl:C8}         & \( \ksIF \negOf{\alpha}\models\beta  \ksTHEN \beliefsOf{\Psi\contraction\alpha\contraction\beta} =_\alpha \beliefsOf{\Psi\contraction\beta}  \)     \\[0.2em] 
            &Semantic        & \eqref{pstl:CR8}        & \( \ksIF \omega_1,\omega_2 \in \modelsOf{\alpha} \ksTHEN \omega_1 \leq_{\Psi} \omega_2 \Leftrightarrow \omega_1 \leq_{\Psi\contraction\alpha} \omega_2 \)                                                             \\[0.2em]\bottomrule
    \end{tabular}
   \end{center}
    \caption{Overview of the different principles proposed for AGM contraction operators which are counter-parts to \eqref{pstl:DP1cond} and their different representations.}\label{fig:syn_postulate_overview_dp1}
\end{figure}

\begin{figure}[t]
    \begin{center}
        \begin{tabular}{rlll}
            \toprule
            \multirow{3}{0.3cm}{\rotatebox{90}{Revision}}&RT-Conditional & \eqref{pstl:DP2cond}   & \( \ksIF \beta\models\negOf{\alpha} \ksTHEN \Psi \models \ramseyCond{\gamma}{\beta} \Leftrightarrow \Psi * \alpha \models \ramseyCond{\gamma}{\beta} \)   \\[0.2em] 
            &Belief & \eqref{pstl:DP2}   & \( \ksIF \beta\models\negOf{\alpha} \ksTHEN \beliefsOf{\Psi * \alpha * \beta} = \beliefsOf{\Psi * \beta} \)   \\[0.2em] 
            &Relational & \eqref{pstl:RR9}   &  \( \ksIF \omega_1,\omega_2 \in \modelsOf{\negOf{\alpha}} \ksTHEN \omega_1 \!\leq_{\Psi}\! \omega_2 \Leftrightarrow \omega_1 \!\leq_{\Psi * \alpha}\! \omega_2 \)   \\[0.2em]  \midrule
            
            \multirow{15}{0.3cm}{\rotatebox{90}{Contraction}}            
            &Contractional & \eqref{pstl:Prop2}   & \( \ksIF \alpha\models\beta \ksTHEN \Psi \contraction \alpha \models \contractionCond{\gamma}{\beta}  \Leftrightarrow \Psi \models \contractionCond{\gamma}{\beta} \)     \\[0.2em]
            &Belief     &    \eqref{pstl:C9prop}     & \( \ksIF \alpha\models\beta \ksTHEN \beliefsOf{\Psi \contraction \alpha \contraction \beta}    = \beliefsOf{\Psi \contraction \beta} \)                                   \\[0.2em]
            &Relational      &                      & no counterpart (Proposition \ref{prop:nonexist_prop2})                                                                                                                    \\[0.2em] \cmidrule{2-4}
            
           &Contractional & \eqref{pstl:C2left}   & \( \ksIF \alpha\models\beta \ksTHEN \Psi  \models \contractionCond{\gamma}{\beta}  \Rightarrow \Psi \contraction \alpha \models \contractionCond{\gamma}{\beta} \)     \\[0.2em]
           &Belief     &                       & \( \ksIF \alpha\models\beta \ksTHEN \Psi  \contraction\beta \models \gamma  \Rightarrow \Psi \contraction \alpha \contraction\beta \models \gamma \)                                                                                                                                                                 \\[0.2em]
           &Semantic      &                       & no counterpart (Proposition \ref{prop:nonexist_prop2})                                                                                                                 \\[0.2em] \cmidrule{2-4}
           &Contractional & \eqref{pstl:C2right}  & \( \ksIF \alpha\models\beta \ksTHEN \Psi \contraction \alpha \models \contractionCond{\gamma}{\beta}  \Rightarrow \Psi \models \contractionCond{\gamma}{\beta} \)      \\[0.2em]
           &Belief     &                       & \( \ksIF \alpha\models\beta \ksTHEN \Psi \contraction \alpha \contraction\beta \models \gamma  \Rightarrow \Psi \contraction\beta \models \gamma \)                                                                                                                                                                 \\[0.2em]
           &Semantic      & \eqref{pstl:CR2right} & \( \ksIF \omega_1,\omega_2 \in \modelsOf{\neg\alpha} \ksTHEN \omega_1 \leq_{\Psi} \omega_2 \Rightarrow \omega_1 \leq_{\Psi\contraction\alpha} \omega_2  \)             \\[0.2em] \cmidrule{2-4}

            &Semantic        & \eqref{pstl:CR9}        & \( \ksIF \omega_1,\omega_2 \in \modelsOf{\negOf{\alpha}} \ksTHEN \omega_1 \leq_{\Psi} \omega_2 \Leftrightarrow \omega_1 \leq_{\Psi\contraction\alpha} \omega_2 \)                                                     \\[0.2em]
            &Contractional   & \eqref{pstl:C9cond}     & \( \ksIF \alpha\models\beta             \ksTHEN   \Psi\!\contraction\!\alpha \models \contractionCond{\negOf{\beta}\to\gamma}{\beta}  \Leftrightarrow \Psi \models \contractionCond{\negOf{\beta}\to\gamma}{\beta} \) \\[0.2em]
            &Belief       & \eqref{pstl:C9}         & \( \ksIF \alpha\models\beta	\ksTHEN \beliefsOf{\Psi\contraction\alpha\contraction\beta} =_\negOf{\beta} \beliefsOf{\Psi\contraction\beta} \)                                                                          \\[0.2em]\bottomrule
    \end{tabular}
    \end{center}
    \caption{Overview of the different principles proposed for AGM contraction operators which are counter-parts to \eqref{pstl:DP2cond} and their different representations.}\label{fig:syn_postulate_overview_dp2}
\end{figure}

{In this section, we consider contraction analogues to the (syntactic) iteration principles given by Darwiche and Pearl.
As a first result,  we will obtain that \eqref{pstl:DP1}--\eqref{pstl:DP3} are not plausible principles for AGM contraction operators, as they are, in a very general way, incompatible with any AGM contraction operator.
Second, we consider slightly modified versions of \eqref{pstl:DP1}--\eqref{pstl:DP4}. 
Again, we will see that these postulates are only partially compatible with AGM contraction operators.

\smallskip\emph{Remark on the naming-pattern.} In this and the following sections we consider several iteration principles of contractions and versions thereof. 
    As already motivated and outlined in Section \ref{sec:conditionals_revision_principle},
     postulates for \emph{iterated contraction} are denoted by IC; recall that postulates for \emph{iterated revision} are abbreviated by IR.
    Moreover, to highlight the difference among the
 postulates for the change of beliefs, change of conditional beliefs, and the change of relations, 
    iteration postulates for contractions with a \emph{contractional} viewpoint are denoted by IC\( {}^\mathrm{cond} \) and IC\( {}^\mathrm{rel} \) stands for considering a \emph{relational} point of view. When considering change of beliefs, we denote them by IC without superscript.
Furthermore, numbering and naming will be chosen to emphasize the correspondence of the contraction postulates to their corresponding postulates  for revision.
Figure \ref{fig:syn_postulate_overview_dp1} to Figure \ref{fig:syn_postulate_overview_dp4}  give an overview of the postulates considered.}

\subsection{The Darwiche-Pearl Postulates \eqref{pstl:DP1}--\eqref{pstl:DP4} as Contraction Postulates}\label{sec:dpasContraction}
A compelling approach is to consider the syntactic Darwiche-Pearl postulates as contraction postulates.
However, the following proposition shows that this is a hopeless endeavour, as each of the postulates \eqref{pstl:DP1}--\eqref{pstl:DP3} is individually incompatible with AGM contraction operators.

\begin{proposition}\label{prop:dp12dp3noncontraction}
    There is no AGM contraction operator \( \contraction \) that satisfies one or more postulates from  \eqref{pstl:DP1}--\eqref{pstl:DP3}.
\end{proposition}
We skip the proof of Proposition \ref{prop:dp12dp3noncontraction} here, and refer to the full proof of Proposition \ref{prop:dp12dp3noncontraction} in Appendix \ref{adx:proofs}.
While there is no AGM contraction operator that is compatible with one of the postulates \eqref{pstl:DP1}--\eqref{pstl:DP3}, the situation if different for \eqref{pstl:DP4}.
The following proposition characterizes \eqref{pstl:DP4} semantically for AGM contraction operators.
\begin{proposition}\label{prop:dp4contraction}
        Let \( \contraction \) be an AGM contraction operator and let \( \Psi\mapsto{\leq_{\Psi}} \) be a faithful-assignment which is \ref{eq:repr_es_contraction}  with \( \contraction \).
The belief change operator \( \contraction \) satisfies \eqref{pstl:DP4} if and only if the following is satisfied:
\begin{description}
    \item[\normalfont(\textlabel{IR4\( {}^\mathrm{rel}_\mathrm{\contraction} \)}{pstl:RR4C})] \( \ksIF \omega_1 \in\modelsOf{\alpha} \ksAND \omega_2\in\modelsOf{\negOf{\alpha}} \ksAND \modelsOf{\Psi}\subseteq\modelsOf{\negOf{\alpha}} \ksTHEN \omega_1 \leq_{\Psi} \omega_2 \Rightarrow \omega_1 \leq_{\Psi \contraction \alpha} \omega_2 \)
\end{description}
\end{proposition}
\begin{proof}We consider both directions of Proposition \ref{prop:dp4contraction} independently.

\begin{itemize}
    \item[]\hspace{-4.4ex}\emph{From \eqref{pstl:RR4C} to \eqref{pstl:DP4}}. 
Towards a contradiction we suppose that \eqref{pstl:DP4} is not satisfied, i.e.,  \( \Psi\contraction\beta\not\models\negOf{\alpha} \) and \( \Psi\contraction\alpha\contraction\beta\models\negOf{\alpha} \) holds.
If \(  \modelsOf{\Psi}\cap\modelsOf{\alpha}\neq\emptyset \), we obtain \eqref{pstl:DP4} directly from \eqref{pstl:C1}.
Thus, we have \( \modelsOf{\Psi}\subseteq\modelsOf{\negOf{\alpha}} \).
From \( \Psi\contraction\beta\not\models\negOf{\alpha} \) and Proposition \ref{prop:es_contraction} we obtain that there is some interpretation \( \omega_1 \in \min(\modelsOf{\negOf{\beta}},\leq_{\Psi}) \) such that \( \omega_1\in\modelsOf{\alpha} \).
Likewise, 
from \( \Psi\contraction\alpha\contraction\beta\models\negOf{\alpha} \) and Proposition \ref{prop:es_contraction} we obtain that there is some interpretation \( \omega_2 \in \min(\modelsOf{\negOf{\beta}},\leq_{\Psi}) \) such that \( \omega_1\in\modelsOf{\negOf{\alpha}} \).
From the latter, we obtain that \( \omega_2 <_{\Psi \contraction \alpha} \omega_1 \) holds.
Application of \eqref{pstl:RR4C} contra-positively yields \( \omega_2 <_{\Psi} \omega_1 \).
This is a contradiction to the minimality of \( \omega_1 \) with respect \( \modelsOf{\negOf{\beta}} \) in \( \leq_{\Psi} \).
    
\item[]\hspace{-4.4ex}\emph{From \eqref{pstl:DP4} to \eqref{pstl:RR4C}}. 
Let \( \omega_1 \in\modelsOf{\alpha} \) and \( \omega_2\in\modelsOf{\negOf{\alpha}} \) and \( \modelsOf{\Psi}\subseteq\modelsOf{\negOf{\alpha}} \).
Towards a contradiction we suppose that we have \( \omega_1 \leq_{\Psi} \omega_2 \) and \( \omega_2 <_{\Psi \contraction \alpha} \omega_1 \).
Because \eqref{pstl:C1} is satisfied, we obtain  \( \omega_1,\omega_2\in\modelsOf{\Psi} \) from  \( \omega_2 <_{\Psi \contraction \alpha} \omega_1 \) and faithfulness of \( \Psi\mapsto{\leq_{\Psi}} \).
Now let \( \beta \) be such that \( \modelsOf{\negOf{\beta}}=\{ \omega_1,\omega_2 \} \) holds.
Since \( \Psi\mapsto{\leq_{\Psi}} \) is \ref{eq:repr_es_contraction}  with \( \contraction \), our assumptions \( \omega_1 \leq_{\Psi} \omega_2 \) and \( \omega_2 <_{\Psi \contraction \alpha} \omega_1 \) together yield that \( \omega_1\in \modelsOf{\Psi\contraction\beta} \) and \( \omega_1\notin \modelsOf{\Psi\contraction\beta} \) holds.
We obtain the contradiction \( \Psi\contraction\beta\not\models\negOf{\alpha} \), but \( \Psi\contraction\alpha\contraction\beta\models\negOf{\alpha} \).\qedhere
\end{itemize}
\end{proof}

Proposition \ref{prop:dp4contraction} yields that \eqref{pstl:DP4} is an iteration principle which is compatible with AGM contraction operators.
Furthermore, Proposition \ref{prop:dp4contraction} points out that  \eqref{pstl:DP4} prevents strict improvement of counter-worlds for \( \alpha \) when contracting by \( \alpha \), which could be considered as counter-intuitive for the process of contraction.
In the next section we consider postulates that are similar but different from \eqref{pstl:DP1}--\eqref{pstl:DP4}.

\subsection{Adapting the Darwiche-Pearl Postulates to a Contraction Attitude}
In the previous section, we established that the postulates \eqref{pstl:DP1}--\eqref{pstl:DP3} are non-valid iteration principles for AGM contraction operators.
We will now consider postulates that are similar to the Darwiche-Pearl Postulates for revision, but reflect better the intention of AGM contraction.
However, we obtain again that these principles are either incompatible with AGM contraction operators or are only partially satisfactory.

We start with the first two postulates%
, which are obtainable from \eqref{pstl:DP1cond} and \eqref{pstl:DP2cond} by replacing the preconditions of these postulates as follows\footnote{\emph{sa} stands for \emph{syntactic analogue}.}:
\begin{description}
	\item[\normalfont(\textlabel{IC1\( {}^\mathrm{cond}_\textrm{sa} \)}{pstl:Prop1})] \( \ksIF \neg\alpha\models\beta \ksTHEN \Psi \contraction \alpha \models \contractionCond{\gamma}{\beta}  \Leftrightarrow \Psi \models \contractionCond{\gamma}{\beta} \)\\
	\emph{Explanation:} If \( \beta \) is believed whenever the negation of \( \alpha \) is believed, then contraction by \( \alpha \) does not influence whether an agent believes \( \gamma \), even in the absence of \( \beta \).
	\item[\normalfont(\textlabel{IC2\( {}^\mathrm{cond}_\textrm{sa} \)}{pstl:Prop2})] \( \ksIF \alpha\models\beta \ksTHEN \Psi \contraction \alpha \models \contractionCond{\gamma}{\beta}  \Leftrightarrow \Psi \models \contractionCond{\gamma}{\beta} \)\\
	\emph{Explanation:}  If \( \beta \) is believed whenever \( \alpha \) is believed, then contraction by \( \alpha \) does not influence whether an agent believes \( \gamma \), even in the absence of \( \beta \).
\end{description}
In revision, the postulates \eqref{pstl:DP1cond} and \eqref{pstl:DP2cond} claim that revision of a belief does not affect revisions by more specific beliefs. 
The postulates \eqref{pstl:Prop1} and \eqref{pstl:Prop2} make a similar, but dual, claim, that a contraction of specific beliefs does not affect a contraction by a more general belief.

We continue with analogues\footnote{We are reading \eqref{pstl:DP4cond} contrapositively.} to \eqref{pstl:DP3cond} and \eqref{pstl:DP4cond}, which are obtainable from \eqref{pstl:DP3cond} and \eqref{pstl:DP4cond} by replacing the consequence of conditionals in these postulates by its negation as in the following:
\begin{description}
	\item[\normalfont(\textlabel{IC3\( {}^\mathrm{cond}_\textrm{sa} \)}{pstl:Prop3})]\( \ksIF \Psi \models \contractionCond{\negOf{\alpha}}{\beta} \ksTHEN \Psi \contraction\alpha \models \contractionCond{\negOf{\alpha}}{\beta} \)\\
	\emph{Explanation:} If \( \neg\alpha \) is believed even in the absence of \( \beta \), then contraction by \( \alpha \) does not revoke this relation between \( \neg\alpha \) and \( \beta \).
	\item[\normalfont(\textlabel{IC4\( {}^\mathrm{cond}_\textrm{sa} \)}{pstl:Prop4})] \( \ksIF \Psi\contraction\alpha \models \contractionCond{\negOf{\alpha}}{\beta} \ksTHEN \Psi \models \contractionCond{\negOf{\alpha}}{\beta} \)\\
	\emph{Explanation:} If posterior to a contraction with \( \alpha \) an agent is convinced that \( \negOf{\alpha} \) should be believed even in the absence of \( \beta \), then the agent was convinced previously about this relation between \( \negOf{\alpha} \) and \( \beta \).
\end{description}
The Postulates \eqref{pstl:DP3cond} and \eqref{pstl:DP4cond} could be read as the claim that a revision by \( \alpha \) does not affect the credibility of \( \alpha \) in subsequent revisions.
Similarly, \eqref{pstl:Prop3} and \eqref{pstl:Prop4} state that 
a contraction by \( \alpha \) does not influence the credibility of the negation of \( \alpha \), and, moreover, contracting by \( \alpha \) does not influence whether believing in \( \negOf{\alpha} \) depends on some other belief \( \beta \).

In the remaining part of this section, we discuss the postulates \eqref{pstl:Prop1} to \eqref{pstl:Prop4}.
We will not only consider a technical viewpoint on contractionals, but also discuss meaning of these postulates.

\begin{figure}[t]
\begin{center}
        \begin{tabular}{rlll}
            \toprule
            \multirow{3}{0.3cm}{\rotatebox{90}{Revision}}&RT-Conditional & \eqref{pstl:DP3cond}   & \( \ksIF \Psi \models \ramseyCond{\alpha}{\beta} \ksTHEN \Psi * \alpha \models \ramseyCond{\alpha}{\beta}  \)   \\[0.2em] 
            &Belief & \eqref{pstl:DP3}   &  \( \ksIF \Psi\revision\beta \models \alpha \ksTHEN \Psi\revision\alpha\revision\beta \models \alpha  \)   \\[0.2em] 
            &Relational & \eqref{pstl:RR10}   &  \( \ksIF \omega_1 \!\in\! \modelsOf{\alpha} \ksAND \omega_2 \!\in\! \modelsOf{\negOf{\alpha}}  \ksTHEN    \omega_1 \!<_{\Psi}\! \omega_2 \! \Rightarrow \! \omega_1 \!<_{\Psi * \alpha}\! \omega_2 \)   \\[0.2em]  \midrule
            
            \multirow{8}{0.3cm}{\rotatebox{90}{Contraction}}  & Contractional & \eqref{pstl:Prop3}   & \( \ksIF \Psi \models \contractionCond{\negOf{\alpha}}{\beta} \ksTHEN \Psi \contraction\alpha \models \contractionCond{\negOf{\alpha}}{\beta} \)                          \\[0.2em]
            &Belief     &                      & \( \ksIF \Psi \contraction\beta \models \negOf{\alpha} \ksTHEN \Psi \contraction\alpha \contraction\beta \models \negOf{\alpha} \)                                        \\[0.2em]
            &Relational      & \eqref{pstl:CR3weak} & \( \ksIF \omega_1 \in \modelsOf{\neg\alpha} , \omega_2 \in \modelsOf{\alpha} \ksAND \modelsOf{\Psi}\subseteq \modelsOf{\negOf{\alpha}}  \),                               \\
            &              &                      & then \(  \omega_1 <_\Psi \omega_2 \Rightarrow \omega_1 <_{\Psi\contraction\alpha} \omega_2 \)                                                                             \\[0.2em]  \cmidrule{2-4}

            &Contractional   & \eqref{pstl:C10condnew} & \( \ksIF \gamma\models\beta \ksTHEN \Psi \models \contractionCond{\alpha\to\gamma}{\beta}   \Rightarrow
            \Psi\contraction\alpha  \models \contractionCond{\alpha\to\gamma}{\beta} \)                        \\[0.2em]
            &Belief       & \eqref{pstl:C10new}     &  \( \ksIF \gamma\models\beta  \ksTHEN  \Psi \contraction \beta \models \alpha \to \gamma \Rightarrow \Psi\contraction\alpha \contraction\beta \models \alpha \to \gamma   \)                                                                                                                                                                                                                     \\[0.2em]
            &Relational        & \eqref{pstl:CR10}       & \( \ksIF \omega_1\!\in\!\modelsOf{\negOf{\alpha}} \ksAND \omega_2\!\in\!\modelsOf{\alpha}  \ksTHEN    \omega_1 \!<_{\Psi}\! \omega_2 \Rightarrow \omega_1 \!<_{\Psi\contraction\alpha}\! \omega_2 \)              \\[0.2em]
            
            \bottomrule
    \end{tabular}
\end{center}
    \caption{Overview of the different principles proposed for AGM contraction operators which are counter-parts to \eqref{pstl:DP3cond} and their different representations.}\label{fig:syn_postulate_overview_dp3}
\end{figure}

\begin{figure}[t]
\begin{center}
        \begin{tabular}{rlll}
            \toprule
            \multirow{3}{0.3cm}{\rotatebox{90}{Revision}}&RT-Conditional & \eqref{pstl:DP4cond}   & \( \ksIF \Psi \models \ramseyCond{\alpha}{\beta} \ksTHEN \Psi * \alpha \models \ramseyCond{\alpha}{\beta}  \)   \\[0.2em] 
            &Belief & \eqref{pstl:DP4}   & \( \ksIF \Psi\revision\beta \not\models \negOf{\alpha} \ksTHEN \Psi\revision\alpha\revision\beta \not\models \negOf{\alpha}  \)   \\[0.2em] 
            &Relational & \eqref{pstl:RR11}   &  \( \ksIF \omega_1  \!\in\! \modelsOf{\alpha} \ksAND \omega_2 \!\in\! \modelsOf{\negOf{\alpha}}  \ksTHEN    \omega_1 \!\leq_{\Psi}\! \omega_2 \! \Rightarrow \! \omega_1 \!\leq_{\Psi * \alpha}\! \omega_2 \)   \\[0.2em]  \midrule
            
            \multirow{8}{0.3cm}{\rotatebox{90}{Contraction}} 
            & Contractional & \eqref{pstl:Prop4}   & \( \ksIF \Psi\contraction\alpha \models \contractionCond{\negOf{\alpha}}{\beta} \ksTHEN \Psi \models \contractionCond{\negOf{\alpha}}{\beta} \)                           \\[0.2em]
            & Belief     &                      & \( \ksIF \Psi\contraction\alpha \contraction\beta \models \negOf{\alpha} \ksTHEN \Psi \contraction\beta \models \negOf{\alpha} \)                                         \\[0.2em]
            &  Relational      & \eqref{pstl:CR4weak} & \( \ksIF \omega_1 \in \modelsOf{\neg\alpha}, \omega_2 \in \modelsOf{\alpha} \ksAND \modelsOf{\Psi}{\subseteq} \modelsOf{\negOf{\alpha}} \),                               \\
            &                &                      & then \( \omega_1 <_{\Psi\contraction\alpha} \omega_2 \Rightarrow {\omega_1 <_\Psi \omega_2} \)                                                                            \\[0.2em]  \cmidrule{2-4}
            
            &Contractional   & \eqref{pstl:C11condnew} & \( \ksIF \gamma\models\beta \ksTHEN \Psi\contraction\alpha \models \contractionCond{\neg\alpha\to\gamma}{\beta}  		  \Rightarrow
            \Psi  \models \contractionCond{\neg\alpha\to\gamma}{\beta} \)             \\[0.2em]
            &Belief       & \eqref{pstl:C11new}     &                \( 
            \ksIF 
            \gamma\models\beta 		\ksTHEN  \Psi \contraction \alpha \contraction \beta \models \neg\alpha \to \gamma \Rightarrow \Psi \contraction \beta \models \neg\alpha \to \gamma \)    \\[0.2em]                                                                                                                                                                                             &Relational        & \eqref{pstl:CR11}       & \( \ksIF \omega_1\!\in\!\modelsOf{\negOf{\alpha}} \ksAND \omega_2\!\in\!\modelsOf{\alpha}  \ksTHEN    \omega_1 \!\leq_{\Psi}\! \omega_2 \!\Rightarrow\! \omega_1 \!\leq_{\Psi\contraction\alpha}\! \omega_2 \)   \\[0.2em]
            \bottomrule
    \end{tabular}
\end{center}
    \caption{Overview of the different principles proposed for AGM contraction operators which are counter-parts to \eqref{pstl:DP4cond} and their different representations.}\label{fig:syn_postulate_overview_dp4}
\end{figure} 

\subsection{Analogues to \eqref{pstl:DP1cond} and \eqref{pstl:DP2cond}}

We will now argue that \eqref{pstl:Prop1} is a too strong postulate to be applicable in every case. 
Consider the following example, which provides a case where \eqref{pstl:Prop1} is too strong.
\begin{example}\label{ex:against_prop1}
	Consider the following three propositions
	\begin{align*}
		\alpha : & \text{Alice attends the party} \\
		\beta  : & \text{Bernd attends the party} \\
		\gamma : & \text{Gavin attends the party}
	\end{align*}
about believing whether Alice, Bernd and Gavin attending to a  party.
Bernd loves parties but seems not to like Alice, thus he is surely joining the party if Alice is not attending to the party (\( \negOf{\alpha} \models \beta \)). Moreover, Gavin seems not to like parties, but Gavin likes Alice, and he is only attending to the party when Alice attends to the party (\( \gamma \models \alpha \)).
	Suppose we believe initially that Gavin is attending to the party (\( \gamma \in \beliefsOf{\Psi} \)). Moreover, we believe that Gavin does not know Bernd, and thus Gavin will join the party, even if Bernd does not attend to the party (\( \Psi \models \contractionCond{\gamma}{\beta} \)).
	
	We arrive at the party, and one of the other guests lets us know that Alice will not join the party. 
	In this new situation (\( \Psi\contraction\alpha \)) it becomes clear that Gavin will not attend to the party (\( \gamma \notin \beliefsOf{\Psi\contraction\alpha} \)), and thus the connection between Gavin and Bernd becomes irrelevant (\( \Psi \contraction \alpha \not\models \contractionCond{\gamma}{\beta} \)).
\end{example}

In particular, Example \ref{ex:against_prop1} indicates that the \enquote{\( \Rightarrow \)}-part of the postulate \eqref{pstl:Prop1} is implausible. 
Moreover, the following proposition shows that this is even reflected by the formal framework.
\begin{proposition}\label{prop:nonexist_prop1} 
	There is no AGM contraction operator \( \contraction \) that satisfies:
	\begin{description}
		\item[\normalfont(\textlabel{IC1\( {}^\mathrm{cond}_{\Leftarrow} \)}{pstl:C1left})] \( \ksIF \neg\alpha\models\beta \ksTHEN \Psi  \models \contractionCond{\gamma}{\beta}  \Rightarrow \Psi \contraction \alpha \models \contractionCond{\gamma}{\beta} \)
	\end{description}
\end{proposition}
\begin{proof}
	We choose two arbitrary but distinct interpretations \( \omega,\omega' \) from \( \Omega \).
	Because that the set of epistemic states \( \setAllES \) is \ref{pstl:unbiased}, there is some epistemic state \( \Psi \) such that \( \modelsOf{\Psi} = \{ \omega \} \).
	Let \( \alpha \) be such that \( \modelsOf{\alpha}=\Omega\setminus\{\omega'\} \) and \( \beta \) such that \( \modelsOf{\beta}=\Omega\setminus \{ \omega \} \).
	Therefore, we have \( \negOf{\alpha} \models \beta \) and \( \modelsOf{\negOf{\beta}} = \{ \omega \} \). By Proposition \ref{prop:es_contraction},  for every contraction operator \( \contraction \) we obtain \( \modelsOf{\Psi\contraction\beta}=\{\omega\} \) and \( \modelsOf{\Psi\contraction\alpha\contraction\beta}=\{ \omega,\omega' \} \). Thus, for every \( \gamma\in\beliefsOf{\Psi} \) with \( \omega'\not\models\gamma \) we obtain \( \Psi  \models \contractionCond{\gamma}{\beta} \) and \( \Psi \contraction \alpha \not\models \contractionCond{\gamma}{\beta} \), which contradicts \eqref{pstl:C1left}.
\end{proof}

Note that the non-existence of an operator stated as in Proposition \ref{prop:nonexist_prop1} does not imply that there is no contraction satisfying \eqref{pstl:C1left} on every epistemic state, but there are certain epistemic states where the postulate \eqref{pstl:C1left} is incompatible with AGM contractions. Consequently, there can not be an AGM contraction operator which satisfies \eqref{pstl:C1left}, as an operator has to yield, for every epistemic state and every input, a new epistemic state.

While one direction of \eqref{pstl:Prop1} is incompatible with AGM contraction operators, the other direction of \eqref{pstl:Prop1} seems to be reasonable and has an intuitive semantic counterpart, as the following proposition shows.
\begin{proposition} 
	Let \( \contraction \) be an AGM contraction operator and let \( \Psi\mapsto{\leq_{\Psi}} \) be a faithful assignment \ref{eq:repr_es_contraction} with \( \contraction \). The following holds
	\begin{description}
		\item[\normalfont(\textlabel{IC1\( {}^\mathrm{cond}_{\Rightarrow} \)}{pstl:C1right})] \( \ksIF \neg\alpha\models\beta \ksTHEN \Psi \contraction \alpha \models \contractionCond{\gamma}{\beta}  \Rightarrow \Psi \models \contractionCond{\gamma}{\beta} \)\\
		\emph{Explanation:} If the negation of \( \alpha \) is more specific than \( \beta \), then a contraction with \( \alpha \) does not influence whether \( \gamma \) is believed in the absence of  \( \beta \) or not.
	\end{description}
	if and only if the following holds:
	\begin{description}
		\item[\normalfont(\textlabel{IC1\( {}^\mathrm{rel}_{\Rightarrow} \)}{pstl:CR1right})] \( \ksIF \omega_1,\omega_2 \in \modelsOf{\alpha} \ksTHEN \omega_1 \leq_{\Psi} \omega_2 \Rightarrow \omega_1 \leq_{\Psi\contraction\alpha} \omega_2 \)
	\end{description}
\end{proposition}
\begin{proof} We consider both directions of the claim independently.
    \begin{itemize}
        \item[]\hspace{-4.4ex}
	\emph{From \eqref{pstl:C1right} to \eqref{pstl:CR1right}.}
	Let \( \omega_1,\omega_2 \in \modelsOf{\alpha} \) and let \( \omega_1 \leq_{\Psi} \omega_2 \). 
	Towards a contradiction, we suppose that \( {\omega_2 <_{\Psi\contraction\alpha} \omega_1} \) holds.
	Now choose \( \beta \in\propLang \) such that \( \modelsOf{\beta}=\Omega\setminus\{ \omega_1,\omega_2 \} \) holds, and choose \( \gamma \) such that \( \modelsOf{\gamma}=\Omega\setminus\{\omega_1\} \) holds.
    Because \( \Psi\mapsto{\leq_{\Psi}} \) is \ref{eq:repr_es_contraction} with \( \contraction \), we obtain \( \omega_1 \in \modelsOf{\Psi\contraction\beta} \) from  \( {\omega_2 <_{\Psi\contraction\alpha} \omega_1} \), and from  \( {\omega_2 <_{\Psi\contraction\alpha} \omega_1} \) we obtain \( \omega_1 \notin \modelsOf{\Psi\contraction\alpha\contraction\beta} \).
	Consequently, we have \( \gamma\notin\beliefsOf{\Psi\contraction\beta} \) and \( \gamma\in\beliefsOf{\Psi\contraction\alpha\contraction\beta} \).
	This contradicts the satisfaction of \eqref{pstl:C1right}.
	
    \item[]\hspace{-4.4ex}
    \emph{From \eqref{pstl:CR1right} to \eqref{pstl:C1right}.}
	Now assume that \eqref{pstl:CR1right} is satisfied. We show that \eqref{pstl:C1right} holds.
	Let \( \negOf{\alpha} \models \beta \)  and \( \Psi \contraction\alpha \models \contractionCond{\gamma}{\beta} \). 
	By Proposition \ref{prop:es_contraction}, we obtain \( \modelsOf{\Psi} \cup \min(\modelsOf{\negOf{\beta}},\leq_{\Psi\contraction\alpha})  \subseteq \modelsOf{\gamma} \).
	Towards a contradiction assume \( \Psi \not\models \contractionCond{\gamma}{\beta} \), i.e. \( \gamma \notin \beliefsOf{\Psi\contraction\beta} \).
	Proposition \ref{prop:es_contraction} yields \( \min(\modelsOf{\negOf{\beta}},\leq_{\Psi}) \not\subseteq \modelsOf{\gamma} \).
	Therefore, there exist some \( \omega_1 \in \min(\modelsOf{\negOf{\beta}},\leq_{\Psi})  \) such that \( \omega_1\notin\modelsOf{\gamma} \) and \( \omega_1 \notin {\min(\modelsOf{\negOf{\beta}},\leq_{\Psi\contraction\alpha})} \).
	Because \( {\min(\modelsOf{\negOf{\beta}},\leq_{\Psi\contraction\alpha})} \) is non-empty, there is some \( \omega_2\in {\min(\modelsOf{\negOf{\beta}},\leq_{\Psi\contraction\alpha})} \). 
    From \( \omega_1 \in \min(\modelsOf{\negOf{\beta}},\leq_{\Psi}) \) and \( \omega_2\in\modelsOf{\negOf{\beta}} \) we immediately obtain \( \omega_1 \leq_{\Psi} \omega_2 \).
        Moreover, from \( \omega_2\in \min(\modelsOf{\negOf{\beta}},\leq_{\Psi\contraction\alpha}) \) and \( \omega_1 \notin \min(\modelsOf{\negOf{\beta}},\leq_{\Psi\contraction\alpha}) \) we conclude that \( \omega_2 <_{\Psi\contraction\alpha} \omega_1 \) holds.
    Recall that    \( \omega_1,\omega_2 \in \modelsOf{\alpha} \), and thus, because \( \contraction \) satisfies \eqref{pstl:CR1right}, we obtain \( {\omega_2 <_{\Psi} \omega_1} \) from the contraposition of \eqref{pstl:CR1right} and \( {\omega_2 <_{\Psi\contraction\alpha} \omega_1} \).
    However, this last observation contradicts our previous observation that we have \( \omega_1 \leq_{\Psi} \omega_2 \).\qedhere
    \end{itemize}
\end{proof}

For the postulate \eqref{pstl:Prop2} we obtain similar results. We begin with the compatible half of the postulate.

\begin{proposition} 
	Let \( \contraction \) be an AGM contraction operator and \( \Psi\mapsto{\leq_{\Psi}} \) compatible with \( \contraction \). The following holds
	\begin{description}
			\item[\normalfont(\textlabel{IC2\( {}^\mathrm{cond}_{\Rightarrow} \)}{pstl:C2right})] \( \ksIF \alpha\models\beta \ksTHEN \Psi \contraction \alpha \models \contractionCond{\gamma}{\beta}  \Rightarrow \Psi \models \contractionCond{\gamma}{\beta} \)\\
		\emph{Explanation:} If \( \alpha \) is more specific than \( \beta \), then a contraction with \( \alpha \) does not influence whether \( \gamma \) is believed in the absence of  \( \beta \) or not.
	\end{description}
if and only if the following holds:
\begin{description}
	\item[\normalfont(\textlabel{IC2\( {}^\mathrm{rel}_{\Rightarrow} \)}{pstl:CR2right})] \( \ksIF \omega_1,\omega_2 \in \modelsOf{\neg\alpha} \ksTHEN \omega_1 \leq_{\Psi} \omega_2 \Rightarrow \omega_1 \leq_{\Psi\contraction\alpha} \omega_2 \)
	\end{description}
\end{proposition}
\begin{proof}We consider both directions of the claim independently.
    \begin{itemize}
        \item[]\hspace{-4.4ex}\emph{From  \eqref{pstl:C2right} to \eqref{pstl:CR2right}.}
	Assume satisfaction of \eqref{pstl:C2right}. We show satisfaction of \eqref{pstl:CR2right} by contradiction.
	Let \( \omega_1,\omega_2 \in \modelsOf{\negOf{\alpha}} \) with \( \omega_1 \leq_{\Psi} \omega_2 \). 
	Towards a contradiction, assume \( \omega_2 <_{\Psi\contraction\alpha} \omega_1 \).
	First, choose \( \beta\in\propLang \) such that \( \modelsOf{\beta}=\Omega\setminus\{\omega_1,\omega_2\} \) and \( \gamma \) such that \( \modelsOf{\gamma}=\Omega\setminus\{ \omega_1 \} \). 
	From our assumptions about \( \leq_{\Psi\contraction\alpha} \) and the faithfulness of \(  \leq_{\Psi\contraction\alpha} \) we obtain \( \omega_1 \notin \modelsOf{\Psi\contraction\alpha\contraction\beta} \) and \( \omega_1\in \modelsOf{\Psi\contraction\beta} \).
	By Proposition \ref{prop:es_contraction}, we obtain \( \gamma \in \beliefsOf{\Psi\contraction\alpha\contraction\beta} \) and \( \gamma\notin\beliefsOf{\Psi\contraction\beta} \).
	Note that \( \alpha\models\beta \), and therefore we obtain a contradiction to \eqref{pstl:C2right}.

        \item[]\hspace{-4.4ex}\emph{From  \eqref{pstl:CR2right} to \eqref{pstl:C2right}.}
	For the other direction, assume satisfaction of \eqref{pstl:CR2right}. We show satisfaction of \eqref{pstl:C2right}.
	Let \( \alpha\models\beta \) and \( \gamma\in\beliefsOf{\Psi\contraction\alpha\contraction\beta} \).
	Now suppose \( \gamma\notin \beliefsOf{\Psi\contraction\beta} \).
	This implies the existence of interpretations \( \omega_1,\omega_2 \) such that   \( \omega_1 \in  \min(\modelsOf{\negOf{\beta}},\leq_{\Psi\contraction\alpha}) \) and \( \omega_2\notin \min(\modelsOf{\negOf{\beta}},\leq_{\Psi\contraction\alpha})  \) and \( \omega_2 \in \min(\modelsOf{\negOf{\beta}},\leq_{\Psi})  \). We obtain \( \omega_1 <_{\Psi\contraction\alpha} \omega_2 \) and \( \omega_2 \leq_{\Psi} \omega_1 \).
	Now note that \( \modelsOf{\negOf{\beta}} \subseteq \modelsOf{\negOf{\alpha}} \), which implies contradiction to \eqref{pstl:CR2right}.\qedhere
\end{itemize}
\end{proof}

The following proposition shows that the other half of \eqref{pstl:Prop2} is  incompatible with AGM contraction operators.
\begin{proposition}\label{prop:nonexist_prop2}
	There is no AGM contraction operator \( \contraction \) that satisfies:
	\begin{description}		
		\item[\normalfont(\textlabel{IC2\( {}^\mathrm{cond}_{\Leftarrow} \)}{pstl:C2left})] \( \ksIF \alpha\models\beta \ksTHEN \Psi  \models \contractionCond{\gamma}{\beta}  \Rightarrow \Psi \contraction \alpha \models \contractionCond{\gamma}{\beta} \)\\
	\end{description}
\end{proposition}
\begin{proof}
	Let \( \contraction \) be an arbitrary contraction operator and let \( \omega \) be an arbitrary interpretation from \( \Omega \).	
Because we assumed that the set of epistemic states satisfies \eqref{pstl:unbiased}, there is some epistemic state \( \Psi_\omega \) such that \( \modelsOf{\Psi_\omega} = \{ \omega \} \).
Assume that \( \Psi\mapsto{\leq_{\Psi}} \) is a faithful assignment compatible with \( \contraction \) as guaranteed by Proposition \ref{prop:es_contraction}. 
Now choose \( \omega'\in\Omega \) such that \( \omega' \in  \min(\Omega\setminus\{\omega\},\leq_{\Psi_\omega}) \). 

Let \( \alpha \) be such that \( \modelsOf{\alpha}=\{\omega\} \) and \( \beta \) such that \( \modelsOf{\beta}=\{\omega, \omega' \} \).
Therefore, we have \( \alpha \models \beta \) and \( \modelsOf{\negOf{\beta}} = \Omega \setminus \{\omega, \omega' \} \). By Proposition \ref{prop:es_contraction},  we obtain \( \modelsOf{\Psi\contraction\beta}=\{\omega\} \cup M \) and \( \modelsOf{\Psi\contraction\alpha\contraction\beta}=\{ \omega,\omega' \}  \cup M \) for some \( M \subseteq \Omega \setminus \{\omega, \omega' \} \). Thus, for every \( \gamma\in\beliefsOf{\Psi} \) with \( \omega'\not\models\gamma \) and \( M \subseteq \modelsOf{\gamma} \) we obtain \( \Psi  \models \contractionCond{\gamma}{\beta} \) and \( \Psi \contraction \alpha \not\models \contractionCond{\gamma}{\beta} \), which contradicts \eqref{pstl:C2left}.
\end{proof}

To illustrate the incompatibility given by Proposition \ref{prop:nonexist_prop2} consider a variation of Example \ref{ex:against_prop1}.
\begin{example}\label{ex:against_prop2}
	Reconsider Example \ref{ex:against_prop1} about Alice, Bernd and Gavin. 
	The specific personal relationship between Alice and Bernd seems to be not very important for Example \ref{ex:against_prop1}.
	Thus, the example scenario also holds when replacing \( \negOf{\alpha} \models \beta \) by \( \alpha \models \beta \).
	In terms of the example scenario: assume that Bernd is indifferent about parties, but seems to like Alice; then he is surely joining the party if Alice is attending to the party (\( \alpha \models \beta \)). 
\end{example}

\subsection{Analogues to \eqref{pstl:DP3} and \eqref{pstl:DP4}}\label{sec:dp34analoguescontraction}

In the following, we characterise and discuss \eqref{pstl:Prop3} and \eqref{pstl:Prop4}.
Remember that if a contractional \( \contractionCond{\alpha}{\beta} \) is accepted by \( \Psi \), then a contraction by \( \beta \) does not revoke believing in \( \alpha \).
Moreover, informally, the contractional \( \contractionCond{\alpha}{\beta} \) states that in particular that believing in \( \alpha \) is somewhat independent of believing in \( \beta \).
In the light of this interpretation, \eqref{pstl:Prop3} and \eqref{pstl:Prop4} establish generally reasonable principles for contraction, stating together that a contraction of a belief \( \alpha \) does not change the independence of believing in \( \negOf{\alpha} \) from believing in \( \beta \), i.e. \eqref{pstl:Prop3} and \eqref{pstl:Prop4} both together are equivalent to:
\begin{equation*}
	 \Psi  \models \contractionCond{\negOf{\alpha}}{\beta} \ksIFFlong  \Psi\contraction\alpha \models \contractionCond{\negOf{\alpha}}{\beta}
\end{equation*}
To illustrate \eqref{pstl:Prop3} and \eqref{pstl:Prop4} we present the following example, which makes use of the contraposition of \eqref{pstl:Prop4}:
\begin{equation*}
	\ksIF \Psi  \not\models \contractionCond{\negOf{\alpha}}{\beta} \ksTHEN  \Psi\contraction\alpha \not\models \contractionCond{\negOf{\alpha}}{\beta}
\end{equation*}
\begin{example}\label{ex:prop3_prop4}
	Let \( \alpha,\beta \) be the following propositions:
	\begin{align*}
		\alpha : & \text{ Arnold likes pop music} \\
		\beta : & \text{ Arnold likes classical music}
	\end{align*}
about Arnold taste of music. Suppose that an agent believes that Arnold does not like Pop music (\( \negOf{\alpha} \in \beliefsOf{\Psi} \)) and likes classical music (\( \beta\in\beliefsOf{\Psi} \)). 
Consider in the following two approaches to the dependences between believing in \( \negOf{\alpha} \) and \( \beta \).
\begin{itemize}
    \item[]\hspace{-4.4ex}For \eqref{pstl:Prop3}: Suppose that the agent is convinced that Arnold tastes of music for different music genres are \emph{independent} of each other, in particular, for the agent believing that Arnold does not like pop music does \emph{not depend} on whether Arnold likes classical music or not (\( \Psi \models \contractionCond{\negOf{\alpha}}{\beta} \)). 

\item[]\hspace{-4.4ex}For \eqref{pstl:Prop4}: Suppose that the agent is convinced that Arnold tastes of music for different music genres are \emph{dependent} from each other. In particular, believing that Arnold does not like pop music depends on believing that Arnold likes classical music (\( \Psi \not\models \contractionCond{\negOf{\alpha}}{\beta} \)).
\end{itemize}
In both cases, when for some reason the agent becomes even more sure that Arnold does not like pop music (\( \Psi\contraction\alpha \)), it would be implausible that the agent changes generally her opinion on the interrelation between tastes of music.
This is reflected by \eqref{pstl:Prop3}, respectively \eqref{pstl:Prop4},  and we would obtain  \( \Psi \contraction \alpha \models \contractionCond{\negOf{\alpha}}{\beta} \), respectively  \( \Psi \contraction \alpha \not\models \contractionCond{\negOf{\alpha}}{\beta} \).
\end{example}

The following two propositions characterize \eqref{pstl:Prop3} and \eqref{pstl:Prop4} semantically.

\begin{proposition}\label{prop:representation_prop3}
	Let \( \contraction \) be an AGM contraction operator and \( \Psi\mapsto{\leq_{\Psi}} \) compatible with \( \contraction \). 
	Then, the operator \( \contraction \) satisfies \eqref{pstl:Prop3}
	if and only if the following holds:
	\begin{description}
		\item[\normalfont(\textlabel{IC3\( {}^\mathrm{rel}_{\textrm{weak}} \)}{pstl:CR3weak})] \( \ksIF \omega_1 \in \modelsOf{\neg\alpha} \ksAND \omega_2 \in \modelsOf{\alpha} \ksAND \modelsOf{\Psi}\subseteq \modelsOf{\negOf{\alpha}}  \), then \(  \omega_1 <_\Psi \omega_2 \Rightarrow \omega_1 <_{\Psi\contraction\alpha} \omega_2 \)
\end{description}
\end{proposition}
\begin{proof}We consider both directions of the claim independently.
    \begin{itemize}
        \item[]\hspace{-4.4ex}\emph{From  \eqref{pstl:Prop3} to \eqref{pstl:CR3weak}.}
	Assume satisfaction of \eqref{pstl:Prop3}. We show satisfaction of \eqref{pstl:CR3weak} by contradiction.
	Let \( \omega_1 \in \modelsOf{\neg\alpha} \) and \( \omega_2 \in \modelsOf{\alpha} \) and \( \modelsOf{\Psi}\subseteq \modelsOf{\negOf{\alpha}} \). Suppose that \( \omega_1 <_\Psi \omega_2 \) and \( \omega_2 \leq_{\Psi\contraction\alpha} \omega_1 \).
	Now choose \( \beta \) such that \( \modelsOf{\beta}=\Omega\setminus\{\omega_1,\omega_2\} \).
	Using Proposition \ref{prop:es_contraction}, we obtain \( \negOf{\alpha} \in \beliefsOf{\Psi\contraction\beta} \) and \( \negOf{\alpha} \notin \beliefsOf{\Psi\contraction\alpha\contraction\beta} \), a direct contradiction to   \eqref{pstl:Prop3}.

    \item[]\hspace{-4.4ex}\emph{From  \eqref{pstl:CR3weak} to \eqref{pstl:Prop3}.}
	For the other direction, assume satisfaction of \eqref{pstl:CR3weak}. We show satisfaction of \eqref{pstl:Prop3}.
	Let \( \negOf{\alpha} \in \beliefsOf{\Psi \contraction \beta} \) and \( \negOf{\alpha} \notin \beliefsOf{\Psi \contraction\alpha\contraction \beta} \). Because of Proposition \ref{prop:es_contraction} we obtain 
	\( \modelsOf{\Psi} \subseteq \modelsOf{\negOf{\alpha}} \) and \( \min(\negOf{\beta},\leq_{\Psi}) \subseteq \modelsOf{\negOf{\alpha}} \). 
	Moreover, Proposition \ref{prop:es_contraction} implies the existence of an interpretation \( \omega_2 \) such that \( \omega_2\notin\modelsOf{\negOf{\alpha}} \) and \( \omega_2 \in \min(\modelsOf{\negOf{\beta}},\leq_{\Psi\contraction\alpha})  \).
	Because \( \omega_2 \) exists, there exists another interpretation \( \omega_1 \) such that \( \omega_1 \in \min(\negOf{\beta},\leq_{\Psi})  \) and thus, \( \omega_1 \in\modelsOf{\negOf{\alpha}} \).
	We obtain that \( \omega_1 <_\Psi \omega_2 \) and \( \omega_2 \leq_{\Psi\contraction\alpha} \omega_1 \), which contradicts \eqref{pstl:CR3weak}.\qedhere
\end{itemize}
\end{proof}

\begin{proposition}\label{prop:representation_prop4}
	Let \( \contraction \) be an AGM contraction operator and \( \Psi\mapsto{\leq_{\Psi}} \) compatible with \( \contraction \). 
Then, the operator \( \contraction \) satisfies \eqref{pstl:Prop4}
	if and only if the following holds:
	\begin{description}
		\item[\normalfont(\textlabel{IC4\( {}^\mathrm{rel}_{\textrm{weak}} \)}{pstl:CR4weak})] \( \ksIF \omega_1 \in \modelsOf{\neg\alpha} \ksAND \omega_2 \in \modelsOf{\alpha} \ksAND \modelsOf{\Psi}\subseteq \modelsOf{\negOf{\alpha}} \), then \( \omega_1 <_{\Psi\contraction\alpha} \omega_2 \Rightarrow \omega_1 <_\Psi \omega_2 \)
	\end{description}
\end{proposition}
\begin{proof}We consider both directions of the claim independently.
    \begin{itemize}
        \item[]\hspace{-4.4ex}\emph{From  \eqref{pstl:Prop4} to \eqref{pstl:CR4weak}.}
	Assume satisfaction of \eqref{pstl:Prop4}. We show satisfaction of \eqref{pstl:CR4weak} by contradiction.
	Let \( \omega_1 \in \modelsOf{\neg\alpha} \) and \( \omega_2 \in \modelsOf{\alpha} \) and \( \modelsOf{\Psi}\subseteq \modelsOf{\negOf{\alpha}} \). Suppose that \( \omega_1 <_{\Psi\contraction\alpha} \omega_2 \) and \( \omega_2 \leq_{\Psi} \omega_1 \).
	Now choose \( \beta \) such that \( \modelsOf{\beta}=\Omega\setminus\{\omega_1,\omega_2\} \).
	Using Proposition \ref{prop:es_contraction}, we obtain \( \negOf{\alpha} \in \beliefsOf{\Psi\contraction\alpha\contraction\beta} \) and \( \negOf{\alpha} \notin \beliefsOf{\Psi\contraction\beta} \), a direct contradiction to   \eqref{pstl:Prop4}.

    \item[]\hspace{-4.4ex}\emph{From  \eqref{pstl:CR4weak} to \eqref{pstl:Prop4}.}
	For the other direction, assume satisfaction of \eqref{pstl:CR4weak}. We show satisfaction of \eqref{pstl:Prop4}.
	Let \( \negOf{\alpha} \in \beliefsOf{\Psi \contraction\alpha\contraction \beta} \) and \( \negOf{\alpha} \notin \beliefsOf{\Psi \contraction \beta} \). Because of Proposition \ref{prop:es_contraction} we obtain 
	\( \modelsOf{\Psi} \subseteq \modelsOf{\negOf{\alpha}} \) and \( \min(\negOf{\beta},\leq_{\Psi\contraction\alpha}) \subseteq \modelsOf{\negOf{\alpha}} \). 
	Moreover, Proposition \ref{prop:es_contraction} implies the existence of an interpretation \( \omega_2 \) such that \( \omega_2\notin\modelsOf{\negOf{\alpha}} \) and \( \omega_2 \in \min(\modelsOf{\negOf{\beta}},\leq_{\Psi})  \).
	Because \( \omega_2 \) exists, there exists another interpretation \( \omega_1 \) such that \( \omega_1 \in \min(\negOf{\beta},\leq_{\Psi\contraction\alpha})  \) and thus, \( \omega_1 \in\modelsOf{\negOf{\alpha}} \).
	We obtain that \( \omega_1 <_{\Psi\contraction\alpha} \omega_2 \) and \( \omega_2 \leq_{\Psi} \omega_1 \), which contradicts \eqref{pstl:CR4weak}.\qedhere
\end{itemize}
\end{proof}

Proposition \ref{prop:representation_prop3} and Proposition \ref{prop:representation_prop4} reveal that \eqref{pstl:Prop3} and \eqref{pstl:Prop4} only apply to a contraction with \( \alpha \) when \( \negOf{\alpha} \) is already believed, i.e., \( \negOf{\alpha} \in \beliefsOf{\Psi} \).

In summary, this section presented and discussed postulates based on the translation of the syntactic Darwiche-Pearl postulates for revision to contraction. 
In the following section, we will consider analogue semantic postulates and describe syntactic counterparts for them. 

\section{Semantic Analogues to \eqref{pstl:RR8}--\eqref{pstl:RR11} for Contraction}
\label{sec:sem_postulates}

In this section, we consider contraction analogues to the semantic Darwiche-Pearl postulates \eqref{pstl:RR8}--\eqref{pstl:RR11} and propose a syntactic characterization for them.
We will also present equivalent formulations of these new syntactic characterizations of the contraction postulates with and without employing contractionals. 
The postulates are summarized in Figure \ref{fig:postulate_overview}.
The semantic contraction postulates which we are consider here, are the following ones, which are given by Chopra, Ghose, Meyer and Wong \cite{KS_ChopraGhoseMeyerWong2008}:
\begin{description}
	\item[\normalfont(\textlabel{IC1\( {}^\mathrm{rel} \)}{pstl:CR8})] \( \ksIF \omega_1,\omega_2 \in \modelsOf{\alpha} \ksTHEN \omega_1 \leq_{\Psi} \omega_2 \Leftrightarrow \omega_1 \leq_{\Psi\contraction\alpha} \omega_2 \)
	\smallskip
	\item[\normalfont(\textlabel{IC2\( {}^\mathrm{rel} \)}{pstl:CR9})] \( \ksIF \omega_1,\omega_2 \in \modelsOf{\negOf{\alpha}} \ksTHEN \omega_1 \leq_{\Psi} \omega_2 \Leftrightarrow \omega_1 \leq_{\Psi\contraction\alpha} \omega_2 \)
	\smallskip
	\item[\normalfont(\textlabel{IC3\( {}^\mathrm{rel} \)}{pstl:CR10})] \( \ksIF \omega_1\in\modelsOf{\negOf{\alpha}} \ksAND \omega_2\in\modelsOf{\alpha}  \ksTHEN    \omega_1 <_{\Psi} \omega_2 \Rightarrow \omega_1 <_{\Psi\contraction\alpha} \omega_2 \)
	\smallskip
	\item[\normalfont(\textlabel{IC4\( {}^\mathrm{rel} \)}{pstl:CR11})] \( \ksIF \omega_1\in\modelsOf{\negOf{\alpha}} \ksAND \omega_2\in\modelsOf{\alpha}  \ksTHEN    \omega_1 \leq_{\Psi} \omega_2 \Rightarrow \omega_1 \leq_{\Psi\contraction\alpha} \omega_2 \)
\end{description}
		The postulates \eqref{pstl:CR8} and \eqref{pstl:CR9} ensure that the order of worlds does not
		change if they are equivalent with respect to the new information received. 
		The postulates \eqref{pstl:CR10} and \eqref{pstl:CR11} enforce that no world that contradicts the contracted information
		is getting relatively more plausible after the contraction than a world that was already more plausible
		before and does not contradict the contracted information \cite{KS_Kern-IsbernerBockSauerwaldBeierle2017a}.
        
        In the next section, we consider characterizations of \eqref{pstl:CR8}--\eqref{pstl:CR11} present in the literature.

\begin{figure}
\begin{center}
    \begin{tabular}{lll}
		\toprule
		Semantic        & \eqref{pstl:CR8}        & \( \ksIF \omega_1,\omega_2 \in \modelsOf{\alpha} \ksTHEN \omega_1 \leq_{\Psi} \omega_2 \Leftrightarrow \omega_1 \leq_{\Psi\contraction\alpha} \omega_2 \)                                                             \\[0.2em]
		Contractional   & \eqref{pstl:C8cond}     & \( \ksIF \negOf{\alpha}\models\beta    \ksTHEN  \Psi\contraction\alpha \models \contractionCond{\alpha\to\gamma}{\beta}   \Leftrightarrow \Psi \models \contractionCond{\alpha\to\gamma}{\beta}   \)  \\[0.2em]
		Syntactic       & \eqref{pstl:C8}         & \( \ksIF \negOf{\alpha}\models\beta  \ksTHEN \beliefsOf{\Psi\contraction\alpha\contraction\beta} =_\alpha \beliefsOf{\Psi\contraction\beta}  \)                                                                       \\[0.2em] %
		\midrule
		Semantic        & \eqref{pstl:CR9}        & \( \ksIF \omega_1,\omega_2 \in \modelsOf{\negOf{\alpha}} \ksTHEN \omega_1 \leq_{\Psi} \omega_2 \Leftrightarrow \omega_1 \leq_{\Psi\contraction\alpha} \omega_2 \)                                                     \\[0.2em]
		Contractional   & \eqref{pstl:C9cond}     & \( \ksIF \alpha\models\beta             \ksTHEN   \Psi\contraction\alpha \models \contractionCond{\negOf{\beta}\to\gamma}{\beta}  \Leftrightarrow \Psi \models \contractionCond{\negOf{\beta}\to\gamma}{\beta} \) \\[0.2em]
		Syntactic       & \eqref{pstl:C9}         & \( \ksIF \alpha\models\beta	\ksTHEN \beliefsOf{\Psi\contraction\alpha\contraction\beta} =_\negOf{\beta} \beliefsOf{\Psi\contraction\beta} \)                                                                          \\[0.2em] %
		\midrule
		Semantic        & \eqref{pstl:CR10}       & \( \ksIF \omega_1\in\modelsOf{\negOf{\alpha}} \ksAND \omega_2\in\modelsOf{\alpha}  \ksTHEN    \omega_1 <_{\Psi} \omega_2 \Rightarrow \omega_1 <_{\Psi\contraction\alpha} \omega_2 \)              \\[0.2em]
		Contractional   & \eqref{pstl:C10condnew} & \( \ksIF \gamma\models\beta \ksTHEN \Psi \models \contractionCond{\alpha\to\gamma}{\beta}   \text{  implies  }
		  \Psi\contraction\alpha  \models \contractionCond{\alpha\to\gamma}{\beta} \)                        \\[0.2em]
		Syntactic       & \eqref{pstl:C10new}     &  \( \ksIF \gamma\models\beta  \ksTHEN  \Psi \contraction \beta \models \alpha \to \gamma \Rightarrow \Psi\contraction\alpha \contraction\beta \models \alpha \to \gamma   \)                                                                                                                                                                                                                     \\[0.2em] %
		\midrule
		Semantic        & \eqref{pstl:CR11}       & \( \ksIF \omega_1\in\modelsOf{\negOf{\alpha}} \ksAND \omega_2\in\modelsOf{\alpha}  \ksTHEN    \omega_1 \leq_{\Psi} \omega_2 \Rightarrow \omega_1 \leq_{\Psi\contraction\alpha} \omega_2 \)        \\[0.2em]
		Contractional   & \eqref{pstl:C11condnew} & \( \ksIF \gamma\models\beta \ksTHEN \Psi\contraction\alpha \models \contractionCond{\neg\alpha\to\gamma}{\beta}  		  \text{  implies  }
		  \Psi  \models \contractionCond{\neg\alpha\to\gamma}{\beta} \)             \\[0.2em]
		Syntactic       & \eqref{pstl:C11new}     &                \( 
		\ksIF 
		\gamma\models\beta 		\ksTHEN  \Psi \contraction \alpha \contraction \beta \models \neg\alpha \to \gamma \Rightarrow \Psi \contraction \beta \models \neg\alpha \to \gamma \)                                                                                                                                                                                                       \\ \bottomrule
	\end{tabular}
\end{center}
\caption{Overview of the different principles proposed in Section \ref{sec:sem_postulates} and their different representations shown here.}\label{fig:postulate_overview}
\end{figure} 
\subsection{Characterization by Konieczny and Pino P{\'{e}}rez}
\label{sec:kpp}

The first syntactic characterization of the postulates \eqref{pstl:CR8} and \eqref{pstl:CR11} was given by involving a revision operator \cite{KS_ChopraGhoseMeyerWong2008}. 
A characterization, which does not depend on other operators, was given by Konieczny and Pino P{\'{e}}rez \cite{KS_KoniecznyPinoPerez2017}.
Therefore, Konieczny and Pino P{\'{e}}rez proposed the following postulates\footnote{\emph{KPP} stands for \emph{Konieczny and Pino P{\'{e}}rez}. The original formulation of these postulates makes use of a formula $ B(\Psi) $ instead of a belief set $ \beliefsOf{\Psi} $ \cite{KS_KoniecznyPinoPerez2017}.} for the iteration of contraction \cite{KS_KoniecznyPinoPerez2017}:

\pagebreak[3]
\begin{description}
    \item[\normalfont(\textlabel{IC1\( {}_\mathrm{KPP} \)}{pstl:KPP8})] \( \ksIF \negOf{\alpha} \models \gamma \ksTHEN 
        \beliefsOf{\Psi \contraction \alpha} \subseteq   \beliefsOf{\Psi \contraction (\alpha\lor\beta)} 
        \Leftrightarrow
        \beliefsOf{\Psi \contraction \gamma \contraction \alpha} \subseteq   \beliefsOf{\Psi \contraction \gamma \contraction (\alpha\lor\beta)}
 \) 
    \smallskip
    \item[\normalfont(\textlabel{IC2\( {}_\mathrm{KPP} \)}{pstl:KPP9})] \( \ksIF \gamma  \models \alpha \ksTHEN
\beliefsOf{\Psi \contraction \alpha} \subseteq  \beliefsOf{\Psi \contraction (\alpha\lor\beta)} 
        \Leftrightarrow
        \beliefsOf{\Psi \contraction \gamma \contraction \alpha} \subseteq  \beliefsOf{\Psi \contraction \gamma \contraction (\alpha\lor\beta)}
 \) 
    \smallskip
    \item[\normalfont(\textlabel{IC3\( {}_\mathrm{KPP} \)}{pstl:KPP10})] \( \ksIF \negOf{\beta}\models \gamma \ksTHEN 
\beliefsOf{\Psi \contraction \gamma \contraction \alpha} \subseteq  \beliefsOf{\Psi \contraction \gamma \contraction (\alpha\lor\beta)} \  \Rightarrow \beliefsOf{\Psi \contraction \alpha} \subseteq \beliefsOf{\Psi \contraction (\alpha\lor\beta)}
 \) 
    \smallskip
    \item[\normalfont(\textlabel{IC4\( {}_\mathrm{KPP} \)}{pstl:KPP11})] \( \ksIF \gamma \models \beta \ksTHEN 
     \beliefsOf{\Psi \contraction \gamma \contraction \alpha} \subseteq  \beliefsOf{\Psi \contraction \gamma \contraction (\alpha\lor\beta)} \Rightarrow \beliefsOf{\Psi \contraction \alpha} \subseteq  \beliefsOf{\Psi \contraction (\alpha\lor\beta)}
    \) 
\end{description}

\noindent For an explanation of \eqref{pstl:KPP8}--\eqref{pstl:KPP11} we refer to Konieczny and Pino P{\'{e}}rez \cite{KS_KoniecznyPinoPerez2017}. The class of operators fulfilling these postulates is captured semantically by the following characterization theorem.
		\begin{proposition}[{\cite{KS_KoniecznyPinoPerez2017}}]\label{prop:it_es_contraction}
	Let $ \contraction $ be an AGM contraction operator and let $ \Psi\mapsto{\leq_{\Psi}} $ be a faithful assignment \ref{eq:repr_es_contraction} with \( \contraction \). Then $ \contraction $ satisfies \eqref{pstl:KPP8}--\eqref{pstl:KPP11} if and only 
	\eqref{pstl:CR8}--\eqref{pstl:CR11} are satisfied.
\end{proposition}

The postulates \eqref{pstl:KPP8}--\eqref{pstl:KPP11} focus on the role of disjunctive beliefs. However, they do not give a direct overview of the impact of \eqref{pstl:CR8}--\eqref{pstl:CR11} on the dynamics of conditional beliefs in the process of iterated contraction.
In the following sections, we consider postulates which actually do provide insight on the dynamics of conditional beliefs, and we will show that these postulates characterize \eqref{pstl:CR8}--\eqref{pstl:CR11}.
Therefore, next, we provide a notion of equivalence, which will be a formal means to describe these postulates. 
\subsection{\( \alpha \)-Equivalence}
	\label{sec:alpha_eq}
For the case of iterated contraction, we need to constrain the notion of equivalence of formulas to specific cases.
For expressing such constraints, 
we propose a notion of equivalence which is relative to a proposition $ \alpha $, which we call $ \alpha $-equivalence.
\begin{definition}[$ \alpha $-equivalence]
	For two sets of interpretations $ \Omega_1,\Omega_2\subseteq\Omega $ and a formula $ \alpha $ we say $ \Omega_1 $ is \emph{$ \alpha $-equivalent} to $ \Omega_2 $, written $ \Omega_1 =_\alpha \Omega_2 $, if $ \Omega_1 $ and $ \Omega_2 $ contain the same set of models of $ \alpha $, 
	i.e. $ \Omega_1 \cap \modelsOf{\alpha} = \Omega_2 \cap \modelsOf{\alpha} $.
\end{definition}
\noindent This is lifted to sets of formulas $ X,Y $, by saying $ X $ is \emph{$ \alpha $-equivalent} to $ Y $, written $ X =_\alpha Y $, if $ \modelsOf{X}=_\alpha \modelsOf{Y} $.
For sets of formulas an alternative formulation is possible:
\begin{proposition}
	Two sets of formulas $ X$ and $Y $ are $ \alpha $-equivalent if and only if $ Cn(X\cup \{\alpha\} ) = Cn(Y\cup \{\alpha\} ) $.
\end{proposition}
 Intuitively, $ X $ and $ Y $ are $ \alpha $-equivalent if they agree on everything about $ \alpha $.  
In the following, we give an example which demonstrates $ \alpha $-equivalence.
\begin{example}
	Suppose a scenario about birds ($ b $), penguins ($ p $) and flying ($ f $).
	Let $ X=Cn(b \land f, p \to f) $ and $ Y=Cn(b \land f, p \to \neg f) $ be belief sets which differ mainly in their beliefs about whether a penguin can fly or not.
	The models of these two belief sets are	$ \modelsOf{X}=\{ bfp, bf\overline{p} \} $ and $ \modelsOf{Y}=\{ bf\overline{p} \} $.
	Then $ X $ and $ Y $ agree in their view on birds that are no penguins, $ X =_{b\land \neg p} Y $, but they do not agree in everything about birds, $ X \neq_{b} Y $.
\end{example}

We will use the notion of $ \alpha $-equivalence as a tool to describe invariants for belief changes. 
As an example, consider the following proposition, holding for every AGM contraction.
\begin{proposition}
	For every AGM contraction operator $ \contraction $ and all propositions $ \alpha,\beta $ the following postulate holds: 
	\begin{equation*}
		\ksIF \negOf{\alpha}\land\beta \equiv \bot \ksTHEN \beliefsOf{\Psi} =_\beta \beliefsOf{\Psi \contraction \alpha}
	\end{equation*}
\end{proposition}
\begin{proof}
	Assume $ \alpha,\beta $ such that $ \neg\alpha\land\beta\equiv\bot $. 
	By Proposition \ref{prop:es_contraction} there is a faithful assignment $ \Psi\mapsto\leq_{\Psi} $ such that \eqref{eq:repr_es_contraction} is fulfilled.
	Since $ \negOf{\alpha}\land\beta \equiv \bot $ holds, $ \negOf{\alpha} $ and $ \beta $ have no models in common.
	Therefore, we can infer that the set $ \min(\modelsOf{\negOf{\alpha}},\leq_{\Psi}) $ contains no models of $ \beta $. 
	Thus from $ \modelsOfES{\Psi\contraction\alpha}=\modelsOfES{\Psi} \cup \min(\modelsOf{\negOf{\alpha}},\leq_{\Psi}) $ we can derive $ \modelsOfES{\Psi\contraction\alpha} =_\beta  \modelsOfES{\Psi} $, which is equivalent to $ \beliefsOf{\Psi} =_\beta \beliefsOf{\Psi \contraction \alpha} $.
\end{proof}

The following proposition relates $ \alpha $-equivalence of beliefs to the acceptance of contractionals.	
\begin{proposition}\label{prop:eq_contractional_correspondence}
	Let $ \contraction $ be an AGM contraction operator, $ \Psi,\Phi $ be epistemic states and $ \alpha,\beta $ propositions. 
	Then $ \beliefsOf{\Psi\contraction\beta} =_\alpha \beliefsOf{\Phi\div\beta} $ holds if and only if for all propositions $ \gamma $  we have:
	\begin{equation*}
		\Psi\models\contractionCond{\alpha \rightarrow \gamma}{\beta} \text{ if and only if } \Phi\models\contractionCond{\alpha \rightarrow \gamma}{\beta}
	\end{equation*}
	
\end{proposition}
\begin{proof}
We consider both directions of the claim independently.

\begin{itemize}
    \item[]\hspace{-4.4ex}\emph{The \enquote{\( \Rightarrow \)}-direction.} For this direction let $ \beliefsOf{\Psi\contraction\beta} =_\alpha \beliefsOf{\Phi\contraction\beta} $. This is equivalent to:
\begin{equation}
\modelsOfES{\Psi\contraction\beta}\cap \modelsOf{\alpha} = \modelsOf{\Phi\contraction\beta} \cap \modelsOf{\alpha} \label{eq:p11:aequivalence}
\end{equation}
Assume now (without loss of generality) that $ \Psi \models \contractionCond{\gamma \lor \negOf{\alpha}}{\beta} $ and $  \Phi \not\models \contractionCond{\gamma \lor \negOf{\alpha}}{\beta} $.
Then we get a contradiction, since there must be a world $ \omega\in\modelsOf{ \negOf{\gamma}\land \alpha } $ such that $ \omega\in\modelsOf{\Phi\contraction\beta} $, which contradicts the assumption in combination with Equation \eqref{eq:p11:aequivalence}.

\item[]\hspace{-4.4ex}\emph{The \enquote{\( \Leftarrow \)}-direction.} For 
all propositions $ \gamma $  it holds hat $ \Psi\models\contractionCond{\gamma \lor \negOf{\alpha}}{\beta} \Leftrightarrow \Phi\models\contractionCond{\gamma \lor \negOf{\alpha}}{\beta} $.
Towards a contradiction assume now that  $ \modelsOfES{\Psi\contraction\beta}\cap \modelsOf{\alpha} \neq \modelsOf{\Phi\contraction\beta} \cap \modelsOf{\alpha} $.
This implies (without loss of generality)  that there is a world $ \omega $ such that $ \omega\notin \modelsOfES{\Psi\contraction\beta}\cap \modelsOf{\alpha} $ but $ \omega \in \modelsOfES{\Phi\contraction\beta}\cap \modelsOf{\alpha} $.
Now let $\gamma$ be a formula such that $ \modelsOf{\gamma}=\modelsOfES{\Psi\contraction\beta}\cap \modelsOf{\alpha} $. 
Clearly, it holds that $ \Psi \contraction \beta \models \gamma\lor\negOf{\alpha} $ and $ \Phi \contraction \beta \not\models \gamma\lor\negOf{\alpha} $.
By the correspondence between contractionals and contractions this is a contradiction the our assumption.\qedhere
\end{itemize} \end{proof}
We introduced \( \alpha \)-equivalence in this section and will employ it in the following section as a means for the succinct formulation of postulates. 
\subsection{Syntactic Characterization of \eqref{pstl:CR8} and \eqref{pstl:CR9}}
\label{sec:relative_change_itr}

In the following, we present two principles \eqref{pstl:C8} and \eqref{pstl:C9}, which will be postulates in the fashion of \eqref{pstl:DP1} and \eqref{pstl:DP2}, specifying situations in which beliefs after a contraction are not influenced by specific prior contractions.
Recall that the postulates \eqref{pstl:Prop1} and \eqref{pstl:Prop2} already describe such situations, but are shown to be incompatible with AGM contraction operators (cf. Section \ref{sec:syn_postulates}). Due to Proposition \ref{prop:contractional_acceptance}, the following postulates are equivalent to \eqref{pstl:Prop1} and \eqref{pstl:Prop2}:
\begin{description}
    \item[\normalfont(\textlabel{IC1\( {}_\mathrm{sa} \)}{pstl:C8prop})]\( \ksIF \negOf{\alpha}\models\beta  \ksTHEN \beliefsOf{\Psi\contraction\alpha\contraction\beta} = \beliefsOf{\Psi\contraction\beta}  \) 
    \smallskip
    \item[\normalfont(\textlabel{IC2\( {}_\mathrm{sa} \)}{pstl:C9prop})]\( \ksIF \alpha\models\beta	\ksTHEN \beliefsOf{\Psi\contraction\alpha\contraction\beta} = \beliefsOf{\Psi\contraction\beta} \) 
\end{description}\pagebreak[3]
The postulates \eqref{pstl:C8} and \eqref{pstl:C9} we propose below are similar to \eqref{pstl:C8prop} and \eqref{pstl:C9prop}, but make use of \( \alpha \)-equivalence to restrict \eqref{pstl:C8prop} and \eqref{pstl:C9prop} to specific cases such that \eqref{pstl:C8} and \eqref{pstl:C9} are compatible with AGM contraction operators.
Moreover, we chose this restriction such that \eqref{pstl:C8} and \eqref{pstl:C9} are equivalent to \eqref{pstl:CR8} and \eqref{pstl:CR9}.

Remember that \eqref{pstl:CR8} and \eqref{pstl:CR9} require that all interpretations of a certain kind should not change their plausibility relative to each other in the process of contraction.
The postulate \eqref{pstl:CR8} enforces this condition for the models of  $ \alpha $, and \eqref{pstl:CR9} enforces the same condition for models of $ \negOf{\alpha} $, thus it is natural to specify the following postulates:
\begin{description}
	\item[\normalfont(\textlabel{IC1}{pstl:C8})]\( \ksIF \negOf{\alpha}\models\beta  \ksTHEN \beliefsOf{\Psi\contraction\alpha\contraction\beta} =_\alpha \beliefsOf{\Psi\contraction\beta}  \) \\
	\emph{Explanation:} The beliefs about $ \alpha $ after a contraction with $ \beta $ are independent from whether $ \alpha $ was contracted previously or not, if $ \beta $ is more general than the negation of $ \alpha $.
	\item[\normalfont(\textlabel{IC2}{pstl:C9})]\( \ksIF \alpha\models\beta	\ksTHEN \beliefsOf{\Psi\contraction\alpha\contraction\beta} =_\negOf{\beta} \beliefsOf{\Psi\contraction\beta} \) \\
	\emph{Explanation:} The beliefs about $ \neg\beta $ after a contraction with $ \beta $ are independent from whether $ \alpha $ was contracted previously or not, if $ \beta $ is more general than $ \alpha $.
\end{description}
We will now show that \eqref{pstl:C8} and \eqref{pstl:C9} are equivalent to \eqref{pstl:CR8} and \eqref{pstl:CR9} for AGM contraction operators. 
Note that \eqref{pstl:C9} makes use of \( \negOf{\beta} \)-equivalence, and the following proposition shows also that this is the right condition to capture \eqref{pstl:CR9}.
\begin{proposition}\label{prop:eqrelated_c8_c9}
    Let $ \contraction $  be an AGM contraction operator and let $ \Psi\mapsto{\leq_{\Psi}} $ be a faithful assignment \ref{eq:repr_es_contraction} with \( \contraction \).
    The operator $ \contraction $ satisfies the postulate \eqref{pstl:C8}, respectively \eqref{pstl:C9}, if and only if \eqref{pstl:CR8}, respectively \eqref{pstl:CR9}, is satisfied.
\end{proposition}
Because of its length, we skip here the proof of Proposition \ref{prop:eqrelated_c8_c9} and present the proof of Proposition \ref{prop:eqrelated_c8_c9} in Appendix \ref{adx:proofs}.
As \eqref{pstl:CR9} explicitly restricts the dynamics of models of \( \negOf{\alpha} \), one might wonder why we do not use the following postulate as counter-part to 
\eqref{pstl:CR9}:
	\begin{description}
	\item[\normalfont(\textlabel{IC2$ ^\prime $}{pstl:C9dash})]\( \ksIF \alpha\models\beta	\ksTHEN \beliefsOf{\Psi\contraction\alpha\contraction\beta} =_\negOf{\alpha} \beliefsOf{\Psi\contraction\beta} \) 
\end{description}
The rationale is that \eqref{pstl:CR9} is more permissive than \eqref{pstl:C9dash}.
Note that \(  \alpha\models\beta \) implies \( \modelsOf{\negOf{\beta}} \subseteq \modelsOf{\negOf{\alpha}} \).
Thus, a careful reading of \eqref{pstl:C9dash} shows that, semantically, \eqref{pstl:C9dash} imposes the additional condition that all minimal models of $ \negOf{\alpha} $ with respect to \( \leq_{\Psi} \) have to be models of $ \negOf{\beta} $ .
The drastic consequence is that there is no AGM contraction operator which is compatible with \eqref{pstl:C9dash}.
The following proposition and its proof provide this as a technical statement.
\begin{proposition}
     There is no AGM contraction operator that satisfies \eqref{pstl:C9dash}.
\end{proposition}
\begin{proof}
    
    Let \( \contraction \) be an AGM contraction operator that satisfies \eqref{pstl:C9dash}. From Proposition \ref{prop:es_contraction} be obtain a faithful assignment \( \Psi\mapsto{\leq_{\Psi}} \) which is \ref{eq:repr_es_contraction} with \( \contraction \).

Let \( \alpha,\beta,\gamma \in\propLang \) be formulas such that \( \modelsOf{\negOf{\alpha}}=\{\omega,\omega'\} \) and \( \modelsOf{\negOf{\beta}}=\{
\omega' \} \) and \( \modelsOf{\negOf{\gamma}}=\{\omega\} \).
Because \( \setAllES \) is \ref{pstl:unbiased}, there exists some epistemic state \( \Psi\in\setAllES \) such that \( \omega,\omega'\notin\modelsOf{\Psi} \).
As \( \Psi\mapsto{\leq_{\Psi}} \) is \ref{eq:repr_es_contraction} with \( \contraction \), we obtain \( \modelsOf{\Psi\contraction\beta} =_{\negOf{\alpha}} \{\omega'\} \) and \( \modelsOf{\Psi\contraction\gamma} =_{\negOf{\alpha}} \{\omega'\} \).
Since \( \contraction \) satisfies \eqref{pstl:C9dash}, we have \( \modelsOf{\Psi\contraction\alpha\contraction\beta} =_{\negOf{\alpha}} \{\omega'\} \). Likewise, we obtain \( \modelsOf{\Psi\contraction\alpha\contraction\gamma} =_{\negOf{\alpha}} \{\omega'\} \) from \eqref{pstl:C9dash}.
Note that contraction-compatibility implies that we have either \( \modelsOf{\Psi\contraction\alpha}=_{\negOf{\alpha}}\{\omega\} \) or \( \modelsOf{\Psi\contraction\alpha}=_{\negOf{\alpha}}\{\omega'\} \) or \( \modelsOf{\Psi\contraction\alpha}=_{\negOf{\alpha}}\{\omega,\omega'\} \).
For all of these three cases we obtain a contradiction, because \( \contraction \) satisfies inclusion \eqref{pstl:C1}. Consequently, there is no AGM contraction operator that satisfies \eqref{pstl:C9dash}.~\qedhere
\end{proof}

	The correspondence between contractionals and contractions from Section \ref{sec:alpha_eq}, in particular Proposition \ref{prop:eq_contractional_correspondence}, allows us to give a conditional formulation of the postulates \eqref{pstl:C8} and \eqref{pstl:C9}:
\begin{description}
	\item[\normalfont(\textlabel{IC1\( {}^\mathrm{cond} \)}{pstl:C8cond})]\( \ksIF \negOf{\alpha}\models\beta    \ksTHEN  \Psi\!\contraction\!\alpha \!\models\! \contractionCond{\alpha\to\gamma}{\beta}   \!\Leftrightarrow\! \Psi \!\models\! \contractionCond{\alpha\to\gamma}{\beta}   \)
	\smallskip
	\item[\normalfont(\textlabel{IC2\( {}^\mathrm{cond} \)}{pstl:C9cond})]\( \ksIF \alpha\models\beta             \ksTHEN   \Psi\!\contraction\!\alpha \models \contractionCond{\negOf{\beta}\to\gamma}{\beta}  \Leftrightarrow \Psi \models \contractionCond{\negOf{\beta}\to\gamma}{\beta} \) 				
\end{description}

\noindent  We close this subsection with a formal statement about the interrelationship between the conditional and non-conditional variant for these postulates.
\begin{proposition}\label{cor:c8_c9_conditional}
	If $ \contraction $ is an AGM contraction operator, then, \eqref{pstl:C8}, respectively \eqref{pstl:C9}, is satisfied by $ \contraction $ if and only if $ \eqref{pstl:C8cond} $, respectively \eqref{pstl:C9cond}, is satisfied.
\end{proposition}

Next we consider postulates which will turn out to be syntactic counter-parts to \eqref{pstl:CR10} and \eqref{pstl:CR11}. 

\subsection{Syntactic Characterization of \eqref{pstl:CR10} and \eqref{pstl:CR11}}
\label{sec:itr_contractionals_1011}

The postulates \eqref{pstl:CR10} and \eqref{pstl:CR11}
both ensure that by a contraction with $ \alpha $,
models of $ \alpha $ should not be improved with respect to models of $ \negOf{\alpha} $.
We have seen that the postulates \eqref{pstl:Prop3} and \eqref{pstl:Prop4} already capture \eqref{pstl:CR10} and \eqref{pstl:CR11} for certain epistemic states (cf. Section \ref{sec:dp34analoguescontraction}, and also consult Figure \ref{fig:syn_postulate_overview_dp3} and Figure \ref{fig:syn_postulate_overview_dp4}).
In this section, we will show that in the framework we are using here, \eqref{pstl:CR10} and \eqref{pstl:CR11} are  fully characterized by the following postulates:
\begin{description}
	\item[\normalfont(\textlabel{IC3\( {}^\mathrm{cond} \)}{pstl:C10condnew})]\( \ksIF \gamma\models\beta \ksTHEN \Psi \models \contractionCond{\alpha\to\gamma}{\beta}  %
	\text{  implies  }
	\Psi\contraction\alpha  \models \contractionCond{\alpha\to\gamma}{\beta} \)\\[\smallskipamount]
	\emph{Explanation:} A contraction with $ \alpha $ preserves that the implication $ \alpha \to \gamma $ is believed even in absence of \( \beta \), if \( \beta \) is more general than %
	$ \gamma $.
	\smallskip
	\item[\normalfont(\textlabel{IC4\( {}^\mathrm{cond} \)}{pstl:C11condnew})]\( \ksIF \gamma\models\beta \ksTHEN \Psi\contraction\alpha \models \contractionCond{\neg\alpha\to\gamma}{\beta}  %
	\text{  implies  }
	\Psi  \models \contractionCond{\neg\alpha\to\gamma}{\beta} \)\\[\smallskipamount]
	\emph{Explanation:} If \( \beta \) is more general than %
	$ \gamma $, then if the implication $ \negOf{\alpha} \to \gamma $ is believed even in absence of \( \beta \) after contraction with $ \alpha $, then this was priorly also true.
\end{description}
By using contraposition and the correspondence between contractionals and contractions given by Proposition \ref{prop:contractional_acceptance}, we obtain the following non-conditional formulation of the principles \eqref{pstl:C11condnew} and \eqref{pstl:C11condnew}:
\begin{description}
	\item[\normalfont(\textlabel{IC3}{pstl:C10new})]\( \ksIF \gamma\models\beta  \ksTHEN  \Psi \contraction \beta \models \alpha \to \gamma \ksIMPLIES \Psi\contraction\alpha \contraction\beta \models \alpha \to \gamma   \)
	\smallskip
	\item[\normalfont(\textlabel{IC4}{pstl:C11new})]\( 
		\ksIF 
		\gamma\models\beta
		\ksTHEN  \Psi \contraction \alpha \contraction \beta \models \neg\alpha \to \gamma \ksIMPLIES \Psi \contraction \beta \models \neg\alpha \to \gamma \)
\end{description}
Note that AGM contraction operators fulfil the inclusion postulate \eqref{pstl:C1}, and therefore no contraction can add additional beliefs. 
The postulates \eqref{pstl:C10new} and \eqref{pstl:C11new} constrain further which beliefs should be retained. The postulate \eqref{pstl:C10new} ensures that a contraction with $ \alpha $ does not internally give up beliefs. 
The postulate \eqref{pstl:C11new} is more difficult, stating that if two sequential contractions do not withdraw a belief $ \gamma $, then the second contraction only does not withdraw $ \gamma $.
The following proposition states equivalence between \eqref{pstl:C10condnew} and \eqref{pstl:C10new}, and equivalence between \eqref{pstl:C11condnew} and \eqref{pstl:C11new}. 
\begin{proposition}\label{cor:c10_c11_conditional}
	Let $ \contraction $ be an AGM contraction operator. Then \eqref{pstl:C10condnew}, respectively \eqref{pstl:C11condnew}, is satisfied $ \contraction $ if and only if \eqref{pstl:C10condnew}, respectively \eqref{pstl:C11condnew}, is fulfilled.
\end{proposition}

We show for the non-conditional postulates \eqref{pstl:C10new} and \eqref{pstl:C10new} that they characterize \eqref{pstl:CR10} and \eqref{pstl:CR11}.
\begin{proposition}\label{prop:eqrelated_c10_C11new}
	Let $ \contraction $  be an AGM contraction operator and let $ \Psi\mapsto{\leq_{\Psi}} $ be a faithful assignment \ref{eq:repr_es_contraction} with \( \contraction \).
    The operator $ \contraction $ satisfies the postulate \eqref{pstl:C10new}, respectively \eqref{pstl:C11new}, if and only if \eqref{pstl:CR10}, respectively \eqref{pstl:CR11}, is satisfied.
\end{proposition}
Because of its length we skip the proof of Proposition \ref{prop:eqrelated_c10_C11new} here and refer to the proof of Proposition \ref{prop:eqrelated_c10_C11new} in Appendix \ref{adx:proofs}.
Next, we consider alternatives to \eqref{pstl:C10condnew} and \eqref{pstl:C11condnew} when \eqref{pstl:C8} and \eqref{pstl:C9} are given. 
\subsection{Alternatives for \eqref{pstl:CR10} and \eqref{pstl:CR11} in the Context of \eqref{pstl:C8} and \eqref{pstl:C9}}
\label{sec:old_itr_contractionals_1011}

One might observe that \eqref{pstl:C10condnew} and \eqref{pstl:C11condnew} are more complex than \eqref{pstl:Prop3} and \eqref{pstl:Prop4}, and are more complex than their counter-parts for revision operators \eqref{pstl:DP3} and \eqref{pstl:DP4}.
In the following, we establish that if \eqref{pstl:C8cond}  and \eqref{pstl:C9cond} are given, then the postulates \eqref{pstl:C10condnew} and \eqref{pstl:C11condnew} can be further simplified.
Consider the following postulates\footnote{The \enquote{alt} in IC3\( {}^\mathrm{cond}_\mathrm{alt} \) stands for \emph{alternative}.}:
	\begin{description}
		\item[\normalfont(\textlabel{IC3\( {}^\mathrm{cond}_\mathrm{alt} \)}{pstl:C10cond})]\( \ksIF \negOf{\alpha}\models\gamma \ksTHEN \Psi \models \contractionCond{\gamma}{\beta}  %
		\text{  implies  }
		\Psi\contraction\alpha  \models \contractionCond{\gamma}{\beta} \)\\[0.2em]
		\emph{Explanation:} A contraction with $ \alpha $ preserves the acceptance of a contractional if its conclusion $ \gamma $ is more general than %
		$ \negOf{\alpha} $.
				\smallskip
	\item[\normalfont(\textlabel{IC4\( {}^\mathrm{cond}_\mathrm{alt} \)}{pstl:C11cond})]\( \ksIF \alpha\models\gamma \ksTHEN  \Psi\contraction\alpha  \models \contractionCond{\gamma}{\beta} %
	\text{ implies }
	\Psi \models \contractionCond{\gamma}{\beta} \) \\[0.2em]
	\emph{Explanation:} If a contractional whose conclusion $ \gamma $ is more general than $ \alpha $ is accepted after a contraction with $ \alpha $, then the contractional should be accepted previously.
\end{description}
By using contraposition and the correspondence between contractionals and contractions, the following non-conditional formulation of the principles \eqref{pstl:C10cond} and \eqref{pstl:C11cond} can be obtained:
		\begin{description}
		\item[\normalfont(\textlabel{IC3\( {}_\mathrm{alt} \)}{pstl:C10})]\( \ksIF \negOf{\alpha}\models\gamma \ksTHEN  \Psi\contraction\beta \models \gamma \ksIMPLIES \Psi\contraction\alpha \contraction\beta \models \gamma \) 	
		\smallskip
		\item[\normalfont(\textlabel{IC4\( {}_\mathrm{alt} \)}{pstl:C11})]\( \ksIF 
		\alpha\models\gamma
		\ksTHEN \Psi \contraction\alpha \contraction\beta \models \gamma \ksIMPLIES \Psi\contraction\beta \models \gamma  \) 
	\end{description}
The following proposition establishes equivalence of \eqref{pstl:C10cond} and \eqref{pstl:C10}, respectively equivalence between \eqref{pstl:C11cond} and \eqref{pstl:C11}. 
\begin{proposition}%\label{cor:c10_c11_conditional}
	Let $ \contraction $ be an AGM contraction operator. Then \eqref{pstl:C10cond}, respectively \eqref{pstl:C11cond}, is satisfied $ \contraction $ if and only if \eqref{pstl:C10cond}, respectively \eqref{pstl:C11cond}, is fulfilled.
\end{proposition}

The following proposition shows that \eqref{pstl:C10new} and \eqref{pstl:C11new} are indeed equivalent to \eqref{pstl:C10} and \eqref{pstl:C11} in the context of \eqref{pstl:C8} and \eqref{pstl:C9}.
\begin{proposition}\label{prop:eqrelated_c10_c11_old}
	For a AGM contraction operator $ \contraction $ the following two statements hold:
	\begin{enumerate}[(a)]
		\item If \( \contraction \)  satisfies \eqref{pstl:C8}, then $ \contraction $ satisfies \eqref{pstl:C10}  if and only if \( \contraction \) satisfies \eqref{pstl:C10new}.
		\item If \( \contraction \)  satisfies \eqref{pstl:C9}, then $ \contraction $ satisfies \eqref{pstl:C11}  if and only if \( \contraction \) satisfies \eqref{pstl:C11new}.
	\end{enumerate}
\end{proposition}
    Again, because of its length, we skip the proof of Proposition \ref{prop:eqrelated_c10_c11_old} here and refer to the full proof of Proposition \ref{prop:eqrelated_c10_c11_old} in Appendix \ref{adx:proofs}.
In the following, we summarize all results of this section.

\subsection{Extended Characterization Theorem}
\label{sec:extended_representation}

We started this section by considering the semantic postulates \eqref{pstl:CR8}--\eqref{pstl:CR11} and showed step by step that \eqref{pstl:C8}--\eqref{pstl:C11new} are appropriate counter-parts that highlight also the role of contractional beliefs.
To summarize the results of this section, we present the main characterization theorem for the sets of postulates presented in this section.

\begin{theorem}[Extended Characterization Theorem]\label{thm:contraction_ext_representation}
	Let $ \contraction $  be an AGM contraction operator and let $ \Psi\mapsto{\leq_{\Psi}} $ be a faithful assignment which is \ref{eq:repr_es_contraction} with \( \contraction \). Then the following statements are equivalent:
\begin{itemize}%
    \item The operator $ \contraction $ satisfies \eqref{pstl:CR8}--\eqref{pstl:CR11}.
	\item The operator $ \contraction $ satisfies \eqref{pstl:C8}--\eqref{pstl:C11new}.
	\item The operator $ \contraction $ satisfies \eqref{pstl:C8cond}--\eqref{pstl:C11condnew}.
	\item The operator $ \contraction $ satisfies \eqref{pstl:C8}, \eqref{pstl:C9}, \eqref{pstl:C10} and \eqref{pstl:C11}.
	\item The operator $ \contraction $ satisfies \eqref{pstl:C8cond}, \eqref{pstl:C9cond}, \eqref{pstl:C10cond} and \eqref{pstl:C11cond}.
	\item The operator $ \contraction $ satisfies \eqref{pstl:KPP8}--\eqref{pstl:KPP11}.
\end{itemize}
\end{theorem}

We have shown that all the groups of postulates considered in Theorem \ref{thm:contraction_ext_representation} are equivalent to \eqref{pstl:CR8}--\eqref{pstl:CR11} in the context of AGM contraction operators.
In particular, the class of AGM contraction operators which fulfil the iteration postulates 
\eqref{pstl:KPP8} to \eqref{pstl:KPP11}
from Konieczny and Pino P{\'{e}}rez \cite{KS_KoniecznyPinoPerez2017}
for contraction can be expressed equivalently by any of the two groups of postulates \eqref{pstl:C8} to \eqref{pstl:C11new} and \eqref{pstl:C8cond} to \eqref{pstl:C11condnew} considered here in this section and summarized in Figure \ref{fig:postulate_overview}.
Moreover, results closely related to the equivalences between \eqref{pstl:C8}--\eqref{pstl:C11new} and \eqref{pstl:CR8}--\eqref{pstl:CR11} stated in Theorem \ref{thm:contraction_ext_representation} have been obtained independently by Eric Raidl and Hans Rott \cite{KS_RaidlRott2021}. %

We want to highlight that each of \eqref{pstl:C8}--\eqref{pstl:C11new} corresponds exactly to one of \eqref{pstl:CR8}--\eqref{pstl:CR11}, e.g., for AGM contraction operators, we have that \eqref{pstl:C8} is satisfied if and only if \eqref{pstl:CR8} is satisfied. 
    Thus, we have a situation similar as for AGM revision operators, where each \eqref{pstl:DP1}--\eqref{pstl:DP4} corresponds exactly to one of \eqref{pstl:RR8}--\eqref{pstl:RR11}.

Finally, we want to remark that one could extend Theorem \ref{thm:contraction_ext_representation} by including also the syntactic postulates for iterated contractions by Chopra, Ghose, Meyer and Wong \cite{KS_ChopraGhoseMeyerWong2008}.
The rationale to not include their result is two-fold. First, their contraction postulates depend on a revision; this complicates specifying a class of operators, since for instance in the iterative case there are more revisions than contractions \cite{KS_KoniecznyPinoPerez2017}. Moreover, the contraction postulates they are considering are slightly stronger, thus the exact relation is open to discover (cf. Section \ref{sec:prelim_contraction}).

\section{An Independence Analogue for Contraction}
\label{sec:ind_contraction}
In this section we consider a contraction analogue to the independence condition for revision by Jin and Thielscher \cite{KS_JinThielscher2007}  (see Section \ref{sec:conditionals_revision_principle}). 
Like in the last section, we start from the semantic side, and we
propose to adapt \( \eqref{pstl:IndR} \), the semantic analogue to \eqref{pstl:Ind},  to the following semantic postulate:
\begin{description}
    \item[\normalfont(\textlabel{IC\( {}^\mathrm{rel}_\mathrm{ind} \)}{pstl:IndCR})] \( \ksIF \omega_1\in\modelsOf{\negOf{\alpha}} \ksAND \omega_2\in\modelsOf{\alpha}\setminus\modelsOf{\Psi}  \ksTHEN    \omega_1 \leq_{\Psi} \omega_2 \Rightarrow \omega_1 <_{\Psi\contraction\alpha} \omega_2 \)
\end{description}
Similar to \eqref{pstl:IndR} for revision, the postulate \eqref{pstl:IndCR} demands strict improvement of counter-models of \( \alpha \) over models of \( \alpha \) when contracting by \( \alpha \). But to deal with the inclusion postulate of contraction \eqref{pstl:C1}, strict improvement has to be restricted to non-models of \( \Psi \).
A contractional counter-part to \eqref{pstl:IndCR} is the following postulate:
\begin{description}
    \item[\normalfont(\textlabel{IC\( {}_\mathrm{ind} \)}{pstl:IndC})] \( 
    \ksIF 
    \negOf{\alpha}\models\gamma 
    \ksAND \Psi \contraction \beta \not\models \negOf{\alpha}\to\beta
    \ksTHEN  
    \Psi\contraction\alpha\models\gamma 
    \allowbreak\Rightarrow \Psi \contraction\alpha\contraction\beta \models \gamma  \)\\[\smallskipamount]
        \emph{Explanation:} If contraction by \( \beta \) ensures that \( \negOf{\alpha} \) does not imply \( \beta \), then every consequence \( \gamma \) of \( \negOf{\alpha} \), which is considered to be true after contraction with \( \alpha \), is not discarded by a subsequent contraction with \( \beta \).
\end{description}
As before, we obtain an equivalent postulate to \eqref{pstl:IndCR} for contractionals by employing Proposition \ref{prop:contractional_acceptance}:
\begin{description}
    \item[{\normalfont(\textlabel{IC\( {}^\mathrm{cond}_\mathrm{ind} \)}{pstl:IndCCond})}] 
    \(     \ksIF 
    \negOf{\alpha}\models\gamma 
    \ksAND \Psi \not\models \contractionCond{\negOf{\alpha}\to\beta}{\beta} 
    \ksTHEN  
    \Psi\models\contractionCond{\gamma}{\alpha}
    \allowbreak\Rightarrow \Psi \contraction\alpha \models \contractionCond{\gamma}{\beta}   \)
\end{description}
In the following, we will show that \eqref{pstl:IndCR} is equivalent to \eqref{pstl:IndCR} for AGM contraction operators.
\begin{proposition}\label{prop:ind_contraction}
    Let $ \contraction $  be an AGM contraction operator and let $ \Psi\mapsto{\leq_{\Psi}} $ be a faithful assignment \ref{eq:repr_es_contraction} with \( \contraction \). 
    The operator $ \contraction $ satisfies the postulate \eqref{pstl:IndCR} if and only if \eqref{pstl:IndC} is satisfied.
\end{proposition}
\begin{proof}
We show the direction from \eqref{pstl:IndCR} to \eqref{pstl:IndC} and the direction from \eqref{pstl:IndC} to \eqref{pstl:IndCR} independently.
\begin{description}    
    \item[\normalfont\emph{From \eqref{pstl:IndCR} to \eqref{pstl:IndC}}.]
    Let \( \negOf{\alpha}\models\gamma  \)
     and \( \Psi\contraction\alpha\models\gamma \) and \( \Psi \contraction \beta \not\models \negOf{\alpha}\to\beta \).
    Towards a contradiction, assume that \(  \Psi \contraction\alpha\contraction\beta \not\models \gamma \) holds.
    From this assumption and Proposition \ref{prop:es_contraction} we obtain that there exists some \( \omega_2\in\Omega \) such that \( \omega_2\not\models\gamma \) and
    \begin{equation*}\tag{\( \star \)}\label{eq:ind_contraction_eq1}
   \omega_2 \in     \modelsOf{\Psi \contraction\alpha\contraction\beta}  = \modelsOf{\Psi\contraction\alpha}\cup\minOf{\modelsOf{\negOf{\beta}}}{\leq_{\Psi\contraction\alpha}}
    \end{equation*}
holds. From \eqref{eq:ind_contraction_eq1} and \(  \Psi\contraction\alpha\models\gamma \), we obtain \( \omega_2 \in \minOf{\modelsOf{\negOf{\beta}}}{\leq_{\Psi\contraction\alpha}} \).
Because we have \( \omega_2\not\models\gamma \), and we have \( \negOf{\alpha}\models\gamma  \), we directly obtain \( \omega_2\models\negOf{\beta}\land\alpha \).

From \( \Psi \contraction \beta \not\models \negOf{\alpha}\to\beta \) we obtain that there exists some \( \omega_1\in\Omega \) such that \( \omega_1 \in \modelsOf{\Psi \contraction \beta} \) and \( \omega_1\models \negOf{\alpha}\land\negOf{\beta} \) holds. 
From the observation that \( \negOf{\beta} \) is consistent, we obtain, by using Proposition \ref{prop:es_contraction} again, that \( \omega_1 \in     {\minOf{\modelsOf{\negOf{\beta}}}{\leq_{\Psi}}} \subseteq \modelsOf{\Psi \contraction \beta} \) holds. 

In summary, we have \( \omega_1 \in     {\minOf{\modelsOf{\negOf{\beta}}}{\leq_{\Psi}}}\cap\modelsOf{\negOf{\alpha}} \) and \( \omega_2 \in {\minOf{\modelsOf{\negOf{\beta}}}{\leq_{\Psi\contraction\alpha}}}\allowbreak\cap\modelsOf{\alpha} \).
The latter yields, together with \( \omega_2\not\models\gamma \) and \( \Psi\contraction\alpha\models\gamma \), that \( \omega_2\notin\modelsOf{\Psi} \) holds.
From the first we obtain \( \omega_1 \leq_{\Psi} \omega_2 \), and thus by \eqref{pstl:IndCR}, we obtain  \( \omega_1 <_{\Psi\contraction\alpha} \omega_2 \). This last observation contradicts the minimality of \( \omega_2 \).

\item[\normalfont\emph{From \eqref{pstl:IndR} to \eqref{pstl:IndCR}}.]
Let \( \omega_1\in\modelsOf{\negOf{\alpha}} \) and \( \omega_2\in\modelsOf{\alpha}\setminus\modelsOf{\Psi} \)  and    \( \omega_1 \leq_{\Psi} \omega_2 \). We show that  \( \omega_1 <_{\Psi\contraction\alpha} \omega_2 \) holds.

Choose \( \beta,\gamma\in\propLang \) such that \( \modelsOf{\gamma}=\Omega\setminus\{\omega_2\} \) and \( \modelsOf{\negOf{\beta}}= \{\omega_1,\omega_2\} \).
By definition, we obtain \( \negOf{\alpha}\models\gamma \).
Because of \( \omega_2\notin\modelsOf{\Psi} \) and \( \omega_2\models\alpha \) and contraction-compatibility of $ \Psi\mapsto{\leq_{\Psi}} $ with \( \contraction \) from Proposition \ref{prop:es_contraction}, we have \( \Psi\contraction\alpha\models\gamma \). Using contraction-compatibility again together with \( \omega_1 \leq_{\Psi} \omega_2 \) yields \( \omega_1\in\modelsOf{\Psi\contraction\beta} \) and thus \( \Psi\contraction\beta \not\models\negOf{\alpha}\to\beta \), as \( \omega_1\not\models\negOf{\alpha}\to\beta \).
Since all preconditions are satisfied, we can apply \eqref{pstl:IndC} and obtain \( \Psi \contraction\alpha\contraction\beta \models \gamma \).
Because of \( \omega_2\not\models\gamma \), we obtain \( \omega_2 \notin \Psi \contraction\alpha\contraction\beta \), implying \( \omega_2 \notin \minOf{\modelsOf{\negOf{\beta}}}{\leq_{\Psi \contraction \alpha}} \). 
Because \( \negOf{\beta} \) has only two models, we obtain that \( \omega_1 \in \minOf{\modelsOf{\negOf{\beta}}}{\leq_{\Psi \contraction \alpha}} \) holds. 
Consequently, we obtain the desired result \( \omega_1 <_{\Psi\contraction\alpha}~\omega_2 \).\qedhere
\end{description}
\end{proof}

Next, we show that independence of contraction \eqref{pstl:IndC} rules out certain operators.
For the remaining parts of this article, an AGM contraction operator \( \contraction \) is said to be \emph{trivial} if the following is satisfied:
\begin{equation*}\tag{triviality}\label{def:triviality}
    \beliefsOf{\Psi\contraction\alpha}=\begin{cases}
        \beliefsOf{\Psi}  & \ksIF \alpha\notin \beliefsOf{\Psi}\\
        \beliefsOf{\Psi}\cap \Cn(\negOf{\alpha})  & \ksOtherwise
    \end{cases}
\end{equation*}
Thus, if \( \alpha \) is believed priorly, then a trivial contraction by \( \alpha \) leads to a situation where the agent only retains beliefs that are consequences of \( \negOf{\alpha} \).
Semantically, a trivial contraction operator includes as much as possible counter-models of \( \alpha \) in the process of contraction by \( \alpha \), i.e., \ref{def:triviality} is the same as:
\begin{equation*}
    \modelsOf{\Psi\contraction\alpha}=\begin{cases}
        \modelsOf{\Psi}  & \ksIF \modelsOf{\Psi}\cap\modelsOf{\negOf{\alpha}}\neq\emptyset\\
        \modelsOf{\Psi}\cup \modelsOf{\negOf{\alpha}}  & \ksOtherwise
    \end{cases}
\end{equation*}
Note that all iteration principles for contraction considered in the previous sections are compatible with trivial contraction operators (if they are compatible with AGM contraction operators at all).
We now show that independence for contraction and trivial AGM contraction operators are not compatible.
\begin{proposition}\label{prop:ind_trivial}
    Every trivial AGM contraction operator violates \eqref{pstl:IndC}.
\end{proposition}
\begin{proof}
Every trivial AGM contraction operator \( \contraction \) is \ref{eq:repr_es_contraction} with the following flat faithful assignment $ \Psi\mapsto{\leq_{\Psi}} $ given by
\begin{equation*}
    \omega_1 \leq_{\Psi} \omega_2 \ \ksIF \omega_1\in\modelsOf{\Psi} \ksOR \omega_2\notin\modelsOf{\Psi}
\end{equation*}
for each \( \Psi \in\setAllES\). 
Consequently, for each \( \omega_1,\omega_2\notin\modelsOf{\Psi} \) we have \( \omega_1 \leq_{\Psi} \omega_2 \) and \(  \omega_2 \leq_{\Psi} \omega_1 \) for each \( \Psi\in\setAllES \).
This is impossible when \eqref{pstl:IndCR} is satisfied. By Proposition \ref{prop:ind_contraction} the operator \( \contraction \) violates \eqref{pstl:IndC}.
\end{proof}

In the next section, we consider specific contraction approaches. 

\section{Conditional Perspective for Contraction Approaches}
\label{sec:concrete_contraction_strategies}
In the previous parts of this article, we considered principles for the process of iteration for AGM contraction operators.
In this section, we consider specific approaches to contraction and investigate the role of contractionals for them.

A fundamental investigation of strategies for iterated contraction is due to Nayak, Orgun et al. \cite{KS_NayakGoebelOrgunPham2006,KS_NayakGoebelOrgun2007,KS_RamachandranNayakOrgun2012}.
In particular, they consider different types of iterated contraction approaches, defined from a semantic viewpoint.
We consider natural contraction and moderate contraction, which are summarized as follows:
\begin{itemize}
	\item \emph{Natural Contraction}: Shift the relation between worlds as few as possible when contracting. %
	This is also known as \emph{conservative contraction}.
	\item \emph{Moderate Contraction}: After contraction of \( \alpha \), strictly prefer counter models of \( \alpha \) over  models of \( \alpha \) (whenever possible). This is also known as \emph{priority contraction}.
\end{itemize}

Characterizations in syntactic terms of these contraction operations are a given in the literature \cite{KS_NayakGoebelOrgunPham2006,KS_NayakGoebelOrgun2007,KS_RamachandranNayakOrgun2012}. 
In the following, we will present elegant alternative characterizations in terms of contractionals.

\subsection{Natural Contraction}
We  define natural contraction as operators keeping as much of the prior total preorder when changing.

\begin{definition}
	Let \( \contraction \) be an AGM contraction operator and \( \Psi \mapsto {\leq_{\Psi}} \) a faithful assignment \ref{eq:repr_es_contraction} with \( \contraction \)
	The operator $ \contraction $ is said to be a \emph{natural contraction} if the following holds\footnote{\emph{nc} stands for \emph{natural contraction}.}:
	\begin{description}
		\item[\normalfont(\textlabel{IC\( {}^\mathrm{rel}_\mathrm{nc} \)}{pstl:NCR})]  \( \ksIF \omega_1,\omega_2 \notin \modelsOf{\Psi\contraction\alpha} \ksTHEN \omega_1 \leq_{\Psi} \omega_2 \ksIFFlong \omega_1 \leq_{\Psi\contraction\alpha} \omega_2  \)
	\end{description}
\end{definition}
Briefly explained, natural contraction is contraction where preferences between worlds change minimally, i.e., \eqref{pstl:NCR} demands to change the total preorder \( \leq_{\Psi} \) as minimal as possible.
It has been shown that natural contraction is characterized by the following postulate
\cite{KS_NayakGoebelOrgun2007,KS_RamachandranNayakOrgun2012}:
\begin{description}
	\item[\normalfont(\textlabel{Insertion}{pstl:insertion})] \( \ksIF \beta \in \beliefsOf{\Psi\contraction\alpha} \ksTHEN \beliefsOf{\Psi\contraction\alpha\contraction\beta} = \beliefsOf{\Psi\contraction\alpha} \cap \beliefsOf{\Psi\contraction\beta}  \)
\end{description}

We like to contribute an additional viewpoint by contractionals, respectively by $\alpha$-equivalence. 
Consider therefore the following proposition.
\begin{proposition}[natural contraction]\label{prop:natural_contraction}
	Let \( \contraction \) be an AGM contraction operator. Then \( \contraction \) is a natural contraction if and only if the following is satisfied:
	\begin{description}
		\item[\normalfont(\textlabel{IC\( {}^\mathrm{cond}_\mathrm{nc} \)}{pstl:NCcond})] \( \ksIF \Psi \contraction\alpha \models \beta \ksTHEN \Psi\contraction\alpha \models \contractionCond{\negOf{\beta}\to\gamma}{\beta} \ksIFFlong \Psi \models \contractionCond{\negOf{\beta}\to\gamma}{\beta}  \)
	\end{description}
\end{proposition}
\begin{proof} Let  \( \Psi \mapsto {\leq_{\Psi}} \) be a faithful assignment \ref{eq:repr_es_contraction} with \( \contraction \).
	First, note that due to Proposition \ref{prop:eq_contractional_correspondence} the postulate \eqref{pstl:NCcond} is equivalent to:
	\begin{description}
	\item[\normalfont(\textlabel{IC\( {}_\mathrm{nc} \)}{pstl:NC})] \( \ksIF \Psi \contraction \alpha \models \beta \ksTHEN \beliefsOf{\Psi\contraction\alpha\contraction\beta} \equiv_{\negOf{\beta}} \beliefsOf{\Psi\contraction\beta}  \)
    \end{description}
We show equivalence of \eqref{pstl:NC} and\eqref{pstl:NCR} by considering both directions independently.
\begin{itemize}
    \item[]\hspace{-4.4ex}\emph{\eqref{pstl:NC} implies \eqref{pstl:NCR}.}
	Therefore, let \( \omega_1,\omega_2 \notin \modelsOf{\Psi\contraction\alpha} \) and \( \omega_1 <_{\Psi} \omega_2 \).
	Now choose \( \beta \) such that \( \modelsOf{\beta}=\Omega\setminus\{\omega_1,\omega_2\} \).
	Then by Proposition \ref{prop:es_contraction} we obtain \( \Psi\contraction\alpha\models\beta \), i.e. \( \modelsOf{\Psi\contraction\alpha}\cap\modelsOf{\negOf{\beta}}=\emptyset \).
	Moreover, because of \( \omega_1 <_{\Psi} \omega_2 \) and Proposition \ref{prop:es_contraction} 
	we obtain \( \modelsOf{\Psi\contraction\beta}\cap\modelsOf{\negOf{\beta}} = \{ \omega_1 \} \).
	Employing \eqref{pstl:NC} yields \( \modelsOf{\Psi\contraction\alpha\contraction\beta}\cap\modelsOf{\negOf{\beta}} = \{ \omega_1 \} \).
	In summary, together with Proposition \ref{prop:es_contraction}, we obtain \( \omega_1 <_{\Psi\contraction\alpha} \omega_2 \).
	
	For the other direction assume \( \omega_1 <_{\Psi\contraction\alpha} \omega_2 \) and choose \( \beta \) as before.
	We obtain  \( \modelsOf{\Psi\contraction\alpha\contraction\beta}\cap\modelsOf{\negOf{\beta}} = \{ \omega_1 \} \).
	Because of \eqref{pstl:NC}, we also have \( \modelsOf{\Psi\contraction\beta}\cap\modelsOf{\negOf{\beta}} = \{ \omega_1 \} \).
	This implies \( \omega_1 <_{\Psi} \omega_2 \).
	
        \item[]\hspace{-4.4ex}\emph{\eqref{pstl:NCR} implies \eqref{pstl:NC}.}
	From \( \Psi\contraction\alpha\models\beta \) and Proposition \ref{prop:es_contraction} conclude \( \modelsOf{\Psi}\cap\modelsOf{\negOf{\beta}} = \emptyset \) and \( {\min(\modelsOf{\negOf{\alpha}},\leq_{\Psi})} \cap\modelsOf{\negOf{\beta}} = \emptyset \).
	Consequently, by Proposition \ref{prop:es_contraction}, we obtain that \( \beliefsOf{\Psi\contraction\alpha\contraction\beta} \equiv_{\negOf{\beta}} \beliefsOf{\Psi\contraction\beta} \) if and only if \( \min(\modelsOf{\negOf{\beta}},\leq_{\Psi}) = \min(\modelsOf{\negOf{\beta}},\leq_{\Psi\contraction\alpha}) \).
	Towards a contradiction suppose \( \min(\modelsOf{\negOf{\beta}},\leq_{\Psi}) \neq \min(\modelsOf{\negOf{\beta}},\leq_{\Psi\contraction\alpha}) \).
	Then there are two interpretations \( \omega_1,\omega_2\in\modelsOf{\negOf{\beta}} \) such that \( \omega_1 \leq_{\Psi} \omega_2 \) if and only if \( \omega_2 <_{\Psi\contraction\alpha} \omega_1 \).
	Note that \( \omega_1,\omega_2 \notin\modelsOf{\Psi\contraction\alpha} \).
	We obtain a direct contradiction to \eqref{pstl:NCR}.\qedhere
\end{itemize}
\end{proof}

Proposition \ref{prop:natural_contraction} shows that natural contraction leads to minimal change on the level of contractionals, as the total preorder changes only minimally. We want to emphasize that in particular \eqref{pstl:NC}, i.e.,
\begin{description}
    \item[\normalfont\eqref{pstl:NC}] \( \ksIF \Psi \contraction \alpha \models \beta \ksTHEN \beliefsOf{\Psi\contraction\alpha\contraction\beta} \equiv_{\negOf{\beta}} \beliefsOf{\Psi\contraction\beta}  \)
\end{description}
used in the proof of Proposition \ref{prop:natural_contraction} shows the conceptual similarity to Boutilier's proposal \cite{KS_Boutilier1996} to minimize changes of conditional beliefs in the process of iterated revision (cf. Postulate \eqref{pstl:CB} in Section \ref{sec:conditionals_revision_principle}).

\subsection{Moderate Contraction}

We define moderate contraction as an operation which gives precedence to counter-models of \( \alpha \) when contracting by \( \alpha \).
\begin{definition}[moderate contraction]
	Let \( \contraction \) be an AGM contraction operator and \( \Psi \mapsto {\leq_{\Psi}} \) a faithful assignment \ref{eq:repr_es_contraction} with \( \contraction \).
	Then $ \contraction $ is called a \emph{moderate contraction} if \eqref{pstl:CR8}, \eqref{pstl:CR9} and the following holds\footnote{\emph{mc} stands for \emph{moderate contraction}.}:
	\begin{description}
		\item[\normalfont(\textlabel{IC\( {}^\mathrm{rel}_\mathrm{mc} \)}{pstl:MCR})] \( \ksIF \omega_1{\in}\modelsOf{\negOf{\alpha}},\ \omega_2{\in}\modelsOf{\alpha} \ksAND \omega_2 \notin \modelsOf{\Psi\contraction\alpha} \ksTHEN \omega_1 <_{\Psi\contraction\alpha} \omega_2  \)
	\end{description}
\end{definition}

We give a characterization theorem for moderate contraction.
\begin{proposition}\label{prop:moderatecontractio}
	Let \( \contraction \) be an AGM contraction operator. 
	The operator \( \contraction \) is a moderate contraction if and only if \eqref{pstl:C8}, \eqref{pstl:C9} and the following is satisfied:
	\begin{description}
		\item[\normalfont(\textlabel{IC\( {}^\mathrm{cond}_\mathrm{mc} \)}{pstl:MCcond})] \( \ksIF \negOf{\alpha}\models\gamma \ksAND \alpha\lor\beta\not\equiv\top  \ksTHEN \Psi\models\contractionCond{\gamma}{\alpha} \ksIMPLIES \Psi\contraction\alpha \models \contractionCond{\gamma}{\beta} \)
	\end{description}
\end{proposition}
\begin{proof} Let  \( \Psi \mapsto {\leq_{\Psi}} \) be a faithful assignment \ref{eq:repr_es_contraction} with \( \contraction \).
	The correspondence between \eqref{pstl:CR8}, \eqref{pstl:CR9} and \eqref{pstl:C8}, \eqref{pstl:C9} is given in Proposition \ref{prop:eqrelated_c8_c9}. 
	Thus, it suffices to show the correspondence between \eqref{pstl:MCcond} and \eqref{pstl:MCR}.
	Note that \eqref{pstl:MCcond} is equivalent to the following postulate due to Proposition \ref{prop:contractional_acceptance}:
	\begin{description}
		\item[\normalfont(\textlabel{IC\( {}_\mathrm{mc} \)}{pstl:MC})] \( \ksIF \negOf{\alpha}\models\gamma \ksAND \alpha\lor\beta\not\equiv\top  \ksTHEN \Psi\contraction\alpha\models\gamma \ksIMPLIES \Psi\contraction\alpha \contraction\beta \models \gamma\)
	\end{description}
We show equivalence of \eqref{pstl:MC} and \eqref{pstl:MCR} by considering both directions independently. In the following, let \( \Psi \mapsto {\leq_{\Psi}} \) be a faithful assignment compatible with \( \contraction \).
\begin{itemize}
    \item[]\hspace{-4.4ex}\emph{\eqref{pstl:MC} implies \eqref{pstl:MCR}.}
	Let \( \omega_1\in\modelsOf{\negOf{\alpha}} \) and \( \omega_2\in\modelsOf{\alpha}\setminus\modelsOf{\Psi\contraction\alpha} \).  Towards a contradiction assume \( \omega_2 \leq_{\Psi\contraction\alpha} \omega_1 \).
	Choose \( \beta,\gamma\in\propLang \) such that \( \modelsOf{\beta}=\Omega\setminus \{ \omega_1,\omega_2 \} \) and \( \modelsOf{\gamma}=\modelsOf{\Psi\contraction\alpha}\cup\modelsOf{\negOf{\alpha}} \).
	We obtain that \( \modelsOf{\negOf{\beta}}=\{\omega_1,\omega_2\} \). Moreover, we have \( \negOf{\alpha}\models\gamma \), \( \gamma\in\beliefsOf{\Psi\contraction\alpha} \) and \( \alpha\lor\beta \not\equiv\top \).
	Therefore, by \eqref{pstl:MC} we obtain \( \gamma\in\beliefsOf{\Psi\contraction\alpha\contraction\beta} \). 
	This implies that \( \omega_2\notin \modelsOf{\Psi\contraction\alpha\contraction\beta} \). However, by Proposition \ref{prop:es_contraction} we obtain the contradiction \( \omega_2\in \modelsOf{\Psi\contraction\alpha\contraction\beta} \), because \( \omega_2\in\min(\modelsOf{\negOf{\beta}},\leq_{\Psi\contraction\alpha}) \). 
	Therefore, it is the case that \( \omega_1 <_{\Psi\contraction\alpha} \omega_2 \) and thus the postulate \eqref{pstl:MCR} is satisfied.
	
    \item[]\hspace{-4.4ex}\emph{\eqref{pstl:MCR} implies \eqref{pstl:MC}.}
	Suppose that \( \negOf{\alpha}\models\gamma \), \( \beta\lor\alpha\not\equiv\top \)  and \( \Psi\contraction\alpha\models\gamma \).
	Towards a contraction assume that \( \Psi\contraction\alpha\contraction\beta \not\models \gamma \).
	This implies that there is some \( \omega_2\in\modelsOf{\Psi\contraction\alpha\contraction\beta} \) with \( \omega_2\not\models\gamma \).
	By Proposition \ref{prop:es_contraction}, we obtain that \( \omega_2\in\min(\modelsOf{\negOf{\beta}},\leq_{\Psi\contraction\alpha}) \). Consequently, we obtain that \( \omega_2\models\alpha \).
	Moreover, because \( \beta\lor\alpha\not\equiv\top \), there is some \( \omega_1\in\Omega \) with \( \omega_1\models \negOf{\beta}\land\negOf{\alpha} \).
	Therefore, we have \( \omega_1 \models\gamma \).
	In summary, we obtain that \( \omega_2\leq_{\Psi\contraction\alpha}\omega_1 \), which is a contradiction to \eqref{pstl:MCR}.\qedhere
\end{itemize}
\end{proof}
    The proof of Proposition \ref{prop:moderatecontractio} shows additionally that \eqref{pstl:MCR} is equivalent to
    \begin{description}
        \item[\normalfont\eqref{pstl:MC}] \( \ksIF \negOf{\alpha}\models\gamma \ksAND \alpha\lor\beta\not\equiv\top  \ksTHEN \Psi\contraction\alpha\models\gamma \ksIMPLIES \Psi\contraction\alpha \contraction\beta \models \gamma\) ,
    \end{description}
when considering AGM contraction operators. 
We obtain the following proposition on the relation between moderate contraction and independence.
\begin{proposition}\label{prop:moderate_ind}
    For every AGM contraction operator, satisfaction of \eqref{pstl:MC} implies satisfaction of \eqref{pstl:IndC}.
    Thus, every moderate AGM contraction operator satisfies \eqref{pstl:IndC}.
\end{proposition}
\begin{proof}
    Let \( \contraction \) is an AGM contraction operator and let  \( \Psi \mapsto {\leq_{\Psi}} \) be a faithful assignment \ref{eq:repr_es_contraction} with \( \contraction \).
    We obtain that satisfaction of \eqref{pstl:MCR} implies satisfaction of \eqref{pstl:IndCR}.
    Because of Proposition \ref{prop:ind_contraction}, satisfaction of \eqref{pstl:IndCR} implies satisfaction of \eqref{pstl:IndC}.
        Moreover, from the proof of Proposition \ref{prop:moderatecontractio} we obtain that satisfaction of \eqref{pstl:MCR} implies satisfaction of \eqref{pstl:MC}.
        We obtain that  satisfaction of \eqref{pstl:MC} implies  satisfaction of \eqref{pstl:IndC}.
        Consequently, every AGM contraction operator satisfies \eqref{pstl:IndC}.
\end{proof}
A direct consequence of Proposition \ref{prop:moderate_ind} and Proposition \ref{prop:ind_trivial} (cf. Section \ref{sec:ind_contraction}) is that moderate contractions are not trivial contractions operators. In particular, \eqref{pstl:MC} is incompatible with trivial AGM contraction operators.
\begin{corollary}
    Every trivial AGM contraction operator violates \eqref{pstl:MC}.
\end{corollary}

We close this section with a remark: one might wonder about the term moderate contraction, because the strategy of moderate contraction seems to be analogue to the idea of (simple) lexicographic revision \cite{KS_NayakPagnuccoPeppas2003} (see Section \ref{sec:conditionals_revision_principle}).
Note that there is a notion of lexicographic contraction, which is different from moderate contraction \cite{KS_Nayak1994}. Moreover, lexicographic revision \cite{KS_Nayak1994} only coincides to simplex lexicographic revision in certain cases \cite[p. 214]{KS_NayakPagnuccoPeppas2003}.

\section{Conclusion and Future Work}
\label{sec:conclusion}
We take a conditional perspective on iterated contraction.
In particular, we considered different contraction analogues to the postulates by Darwiche and Pearl of revision.
For the first set of postulates \eqref{pstl:Prop1} to \eqref{pstl:Prop4} we showed that they are either unsatisfiable or capture only certain cases.
The second set of postulates \eqref{pstl:C8cond} to \eqref{pstl:C11condnew} is shown to be equivalent to the known semantic principles \eqref{pstl:CR8} to \eqref{pstl:CR11} for iterated contraction.
Additionally, we provide a set of syntactic postulates \eqref{pstl:C8} to \eqref{pstl:C11new} for iterated contraction which are more succinct than \eqref{pstl:KPP8} to \eqref{pstl:KPP11} proposed in \cite{KS_KoniecznyPinoPerez2017}.
Moreover, without \eqref{pstl:C8} to \eqref{pstl:C11new} it would be difficult to obtain \eqref{pstl:C8cond} to \eqref{pstl:C11condnew}.
We proved an extended characterization theorem for iterated contraction, which shows that all these different sets of postulates describe the same set of operators.

The notion of $ \alpha $-equivalence was introduced as a form of equivalence between epistemic states with respect to a proposition $ \alpha $. 
We showed the usefulness of this relation, in particular for the formulation of postulates, and how it allows us to provide insights about invariants in belief change.

For the postulates \eqref{pstl:Prop1} to \eqref{pstl:Prop4} and \eqref{pstl:C8cond} to \eqref{pstl:C11condnew} we use specific conditionals, called contractionals, which are connected to contractions in the same manner as the Ramsey test draws a connection to revisions.
Contractionals have been studied in the context of inference by Bochman \cite{KS_Bochman2001}. 
Furthermore, we showed that $ \alpha $-equivalence is connected to the acceptance of contractionals.
To the best of our knowledge, contractionals have not been used in the context of iterated belief contraction so far.
Note also that the theory developed here was already successfully used for investigations on decrement operators \cite{KS_SauerwaldBeierle2019}.
    
For natural contraction, the counter-part to natural revision, we considered alternative postulates that show more similarity to the postulates for natural revision.
Additionally, we investigated the independence condition by Jin and Thielscher \cite{KS_JinThielscher2007} in the context of contraction.

For future work, we like to investigate the role of specific epistemic states and the interrelation between contraction and revision by the means of contractionals and Ramsey test conditionals.
Another open question, and closely related one, is the notion of expansion for iterated belief change \cite{KS_FermeWassermann2018}. We think the application of the approach presented in this article, by employing conditionals for expansion, could be fruitful to tackle this question.

\section*{Acknowledgements}

This work was supported by the Deutsche Forschungsgemeinschaft (DFG, German Research Foundation), grants BE 1700/9-1 and BE 1700/10-1 awarded to Christoph Beierle 
and grants KE~1413/10-1 and \hbox{KE~1413/12-1} awarded to Gabriele Kern-Isberner
as part of the priority program "Intentional Forgetting in Organizations" (SPP 1921). Kai Sauerwald is supported by the grants BE 1700/9-1 and BE 1700/10-1.

\pagebreak[4]
\bibliographystyle{plain}
\newcommand{\verzeichnisBibtex}{\string~/BibTeXReferencesSVNlink}
\bibliography{bibexport%
} 

\clearpage
\appendix
\pagenumbering{Roman}\gdef\thesection{\Alph{section}} 
\makeatletter
\renewcommand\@seccntformat[1]{Appendix \csname the#1\endcsname.\hspace{0.5em}}
\makeatother
\section{Full Proofs}\label{adx:proofs}

\newenvironment{apxproposition}[1]{\medskip\noindent\textbf{Proposition \ref{#1}.}}{}

\begin{apxproposition}{prop:dp12dp3noncontraction}
    There is no AGM contraction operator \( \contraction \) that satisfies one or more postulates from  \eqref{pstl:DP1}--\eqref{pstl:DP3}.
\end{apxproposition}
\begin{proof}
    Let \( \contraction \) be a AGM contraction operator and let \( \Psi\mapsto{\leq_{\Psi}} \) be a faithful-assignment which is \ref{eq:repr_es_contraction} compatible with \( \contraction \).
We consider each of the postulates \eqref{pstl:DP1}--\eqref{pstl:DP4} independently.
\begin{description}
    \item[\eqref{pstl:DP1}] Assume that \( \alpha,\beta\in\propLang \) are non-tautological consistent formulas such that \( \beta\models\alpha \).
    Clearly, we obtain that \( \modelsOf{\negOf{\alpha}} \subseteq \modelsOf{\negOf{\beta}} \) holds.
    From the compatibility and satisfaction of \eqref{pstl:DP1} we obtain 
    \begin{equation*}\tag{\( \star \)}\label{eq:proofdp1_nocontraction}
        \modelsOf{\Psi} \cup \min(\modelsOf{\negOf{\beta}},\leq_{\Psi}) = \modelsOf{\Psi} \cup \min(\modelsOf{\negOf{\alpha}},\leq_{\Psi})  \cup \min(\modelsOf{\negOf{\beta}},\leq_{\Psi\contraction\alpha}).
    \end{equation*}
    for each \( \Psi\in\setAllES \). 
    Because \( \contraction \) satisfies inclusion \eqref{pstl:C1} and because of \( \modelsOf{\negOf{\alpha}} \subseteq \modelsOf{\negOf{\beta}} \), we have \( (\modelsOf{\Psi}\cap\modelsOf{\negOf{\beta}}) \cup \min(\modelsOf{\negOf{\alpha}},\leq_{\Psi})  = \min(\modelsOf{\negOf{\beta}},\leq_{\Psi\contraction\alpha}) \).
    If \( \modelsOf{\Psi}\cap\modelsOf{\negOf{\beta}}=\emptyset \), then we obtain \( \min(\modelsOf{\negOf{\alpha}},\leq_{\Psi})  = \min(\modelsOf{\negOf{\beta}},\leq_{\Psi\contraction\alpha}) \).
    Thus, by \eqref{eq:proofdp1_nocontraction}, we obtain \( \min(\modelsOf{\negOf{\alpha}},\leq_{\Psi}) =  {\min(\modelsOf{\negOf{\beta}},\leq_{\Psi})} \).
    If \( \modelsOf{\Psi}\cap\modelsOf{\negOf{\beta}}\neq\emptyset \), we obtain from inclusion \eqref{pstl:C1} and faithfulness of \( \Psi\mapsto{\leq_{\Psi}} \) that \( \modelsOf{\Psi}\cap\modelsOf{\negOf{\beta}} = \min(\modelsOf{\negOf{\beta}},\leq_{\Psi}) \) holds. This together with \eqref{eq:proofdp1_nocontraction} implies \( \min(\modelsOf{\negOf{\beta}},\leq_{\Psi})=\min(\modelsOf{\negOf{\alpha}},\leq_{\Psi}) \).
    
    Consequently, we have  \( \min(\modelsOf{\negOf{\beta}},\leq_{\Psi})=\min(\modelsOf{\negOf{\alpha}},\leq_{\Psi}) \).
    Now let \( \omega_1,\omega_2 \in \Omega \) with \( \omega_1\neq\omega_2 \) such that \( \modelsOf{\negOf{\beta}}=\{\omega_1,\omega_2\} \) and \( \modelsOf{\negOf{\alpha}}=\{\omega_2\} \).
    Remember that \( \setAllES \) is \eqref{pstl:unbiased}, and thus there is some epistemic state \( \Psi_2 \) with \( \modelsOf{\Psi_2}=\{\omega_2\} \).
    From the faithfulness of \( \Psi\mapsto{\leq_{\Psi}} \) we obtain \( \min(\modelsOf{\negOf{\beta}},\leq_{\Psi_2})=\{ \omega_2 \} \). 
    This is a contradiction, because we have \( \min(\modelsOf{\negOf{\alpha}},\leq_{\Psi_2})=\{ \omega_1 \}\neq\{ \omega_2 \} \).
    
    \item[\eqref{pstl:DP2}] The proof for \eqref{pstl:DP2} is analogue to the proof for \eqref{pstl:DP1}, one has to exchange \( \alpha \) by \( \negOf{\alpha} \). For the sake of completeness, we present the proof in the following.
    
    Assume that \( \alpha,\beta\in\propLang \) are non-tautological consistent formulas such that \( \beta\models\negOf{\alpha} \).
    Clearly, we obtain that \( \modelsOf{\alpha} \subseteq \modelsOf{\negOf{\beta}} \) holds.
    From the compatibility of \( \Psi\mapsto{\leq_{\Psi}} \) with \( \contraction \) and satisfaction of \eqref{pstl:DP2} we obtain 
    \begin{equation*}\tag{\( \star \)}\label{eq:proofdp2_nocontraction}
        \modelsOf{\Psi} \cup \min(\modelsOf{\negOf{\beta}},\leq_{\Psi}) = \modelsOf{\Psi} \cup \min(\modelsOf{\alpha},\leq_{\Psi})  \cup \min(\modelsOf{\negOf{\beta}},\leq_{\Psi\contraction\negOf{\alpha}}).
    \end{equation*}
    for each \( \Psi\in\setAllES \). 
    Because \( \contraction \) satisfies inclusion \eqref{pstl:C1} and because of \( \modelsOf{\alpha} \subseteq \modelsOf{\negOf{\beta}} \), we have \( (\modelsOf{\Psi}\cap\modelsOf{\negOf{\beta}}) \cup \min(\modelsOf{\alpha},\leq_{\Psi})  = \min(\modelsOf{\negOf{\beta}},\leq_{\Psi\contraction\negOf{\alpha}}) \).
    If \( \modelsOf{\Psi}\cap\modelsOf{\negOf{\beta}}=\emptyset \), then we obtain \( \min(\modelsOf{\alpha},\leq_{\Psi})  = \min(\modelsOf{\negOf{\beta}},\leq_{\Psi\contraction\negOf{\alpha}}) \).
    Thus, by \eqref{eq:proofdp2_nocontraction}, we obtain \( \min(\modelsOf{\alpha},\leq_{\Psi}) =  {\min(\modelsOf{\negOf{\beta}},\leq_{\Psi})} \).
    If \( \modelsOf{\Psi}\cap\modelsOf{\negOf{\beta}}\neq\emptyset \), we obtain from inclusion \eqref{pstl:C1} and faithfulness of \( \Psi\mapsto{\leq_{\Psi}} \) that \( \modelsOf{\Psi}\cap\modelsOf{\negOf{\beta}} = \min(\modelsOf{\negOf{\beta}},\leq_{\Psi}) \) holds. This together with \eqref{eq:proofdp2_nocontraction} implies \( \min(\modelsOf{\negOf{\beta}},\leq_{\Psi})=\min(\modelsOf{\alpha},\leq_{\Psi}) \).
    
    Consequently, we have  \( \min(\modelsOf{\negOf{\beta}},\leq_{\Psi})=\min(\modelsOf{\alpha},\leq_{\Psi}) \).
    Now let \( \omega_1,\omega_2 \in \Omega \) with \( \omega_1\neq\omega_2 \) such that \( \modelsOf{\negOf{\beta}}=\{\omega_1,\omega_2\} \) and \( \modelsOf{\alpha}=\{\omega_2\} \).
    Remember that \( \setAllES \) is \eqref{pstl:unbiased}, and thus there is some epistemic state \( \Psi_2 \) with \( \modelsOf{\Psi_2}=\{\omega_2\} \).
    From the faithfulness of \( \Psi\mapsto{\leq_{\Psi}} \) we obtain \( \min(\modelsOf{\negOf{\beta}},\leq_{\Psi_2})=\{ \omega_2 \} \). 
    This is a contradiction, because we have \( \min(\modelsOf{\alpha},\leq_{\Psi_2})=\{ \omega_1 \}\neq\{ \omega_2 \} \).
    
    \item[\eqref{pstl:DP3}] Let \( \alpha \) be a non-tautological belief. Because \( \contraction \) satisfies \eqref{pstl:C3} we obtain \( \alpha\notin\beliefsOf{\Psi\contraction\alpha} \).
    By inclusion \eqref{pstl:C1}, we directly obtain \(  \alpha\notin\beliefsOf{\Psi\contraction\alpha\contraction\beta} \).
    The latter is contradiction to the satisfaction of \eqref{pstl:DP3}.
\end{description}  
In summary, we have shown that there is no AGM contraction operator \( \contraction \) that satisfies one or more postulates from  \eqref{pstl:DP1}--\eqref{pstl:DP3}. \end{proof}

\pagebreak[3]
\begin{apxproposition}{prop:eqrelated_c8_c9}
Let $ \contraction $  be an AGM contraction operator and let $ \Psi\mapsto{\leq_{\Psi}} $ be a faithful assignment \ref{eq:repr_es_contraction} with \( \contraction \).
    Then the following statements are equivalent:
    \begin{enumerate}[(a)]
        \item The operator $ \contraction $ satisfies the postulates \eqref{pstl:C8} and \eqref{pstl:C9}.
        \item The operator $ \contraction $ satisfies the postulates \eqref{pstl:CR8} and \eqref{pstl:CR9}.
    \end{enumerate}
\end{apxproposition}
\begin{proof}
\begingroup%
We will show that (a) implies (b) and that (b) implies (a).

\textbf{Part I:}
We start with the (b) to (a) direction. Let $ \change $ be an AGM contraction operator for epistemic states, thus fulfilling \eqref{pstl:C1} to \eqref{pstl:C7}, and let $ \Psi\mapsto\leq_{\Psi} $ be a faithful assignment related to $ \change $ by \eqref{eq:repr_es_contraction} such that \eqref{pstl:CR8} to \eqref{pstl:CR9} are fulfilled.
We show that \eqref{pstl:C8} and \eqref{pstl:C9} are satisfied:
\begin{description}
	\item[\eqref{pstl:C8}] 
	Let $ \negOf{\beta}\models\alpha $, which is equivalent to $ \negOf{\alpha}\models\beta $. 
	As in the proof of \eqref{pstl:CR9}, Equation \eqref{eq:repr_es_contraction} implies
	\begin{align}
	\modelsOfES{\Psi\change\alpha\change\beta}  & = \modelsOfES{\Psi} \cup \min(\modelsOf{\negOf{\alpha}},\leq_{\Psi}) \cup \min(\modelsOf{\negOf{\beta}},\leq_{\Psi\change\alpha}), \label{eq:proof:C8:4}
	\end{align}
	for every $ \alpha,\beta $.
	Furthermore, by \eqref{pstl:CR8} and $ \negOf{\beta}\models\alpha $ it holds that:
	\begin{equation}
      \min(\modelsOf{\negOf{\beta}},\leq_{\Psi}) = \min(\modelsOf{\negOf{\beta}},\leq_{\Psi\change\alpha}) \label{eq:proof:C8:5}
	\end{equation}
	Combining \eqref{eq:proof:C8:4} with \eqref{eq:proof:C8:5} yields:
	\begin{equation}
	\modelsOfES{\Psi\change\alpha\change\beta}  = \modelsOfES{\Psi} \cup \min(\modelsOf{\negOf{\alpha}},\leq_{\Psi}) \cup \min(\modelsOf{\negOf{\beta}},\leq_{\Psi}) \label{eq:proof:C8:6}
	\end{equation}
	By using $ \modelsOfES{\Psi\change\beta} = \modelsOfES{\Psi} \cup \min(\modelsOf{\negOf{\beta}},\leq_{\Psi}) $ obtained from Equation \eqref{eq:repr_es_contraction} and using $ \min(\modelsOf{\negOf{\alpha}},\leq_{\Psi})  \cap \modelsOf{\alpha} = \emptyset $,  from \eqref{eq:proof:C8:6} we conclude that
	\begin{equation}
	\modelsOfES{\Psi\change\alpha\change\beta} \cap \modelsOf{\alpha} = \modelsOfES{\Psi\change\beta} \cap \modelsOf{\alpha}
	\end{equation}
	holds,
	which is equivalent to the required result $ \beliefsOf{\Psi\change\alpha\change\beta} =_\alpha \beliefsOf{\Psi\change\beta} $.
	\item[\eqref{pstl:C9}] %
	Let $ \negOf{\beta}\models\negOf{\alpha} $, which is equivalent to $ \alpha\models\beta $. 
		By \eqref{pstl:CR9} and $ \negOf{\beta}\models\negOf{\alpha} $ it holds that:
		\begin{equation}
		\min(\modelsOf{\negOf{\beta}},\leq_{\Psi}) = \min(\modelsOf{\negOf{\beta}},\leq_{\Psi\change\alpha}) \label{eq:proof:C9:5}
		\end{equation}
		Combining Equation 
		\eqref{eq:proof:C8:4} that holds for every $ \alpha,\beta $
		with \eqref{eq:proof:C9:5} yields:
		\begin{equation}
		\modelsOfES{\Psi\change\alpha\change\beta}  = \modelsOfES{\Psi} \cup \min(\modelsOf{\negOf{\alpha}},\leq_{\Psi}) \cup \min(\modelsOf{\negOf{\beta}},\leq_{\Psi}) \label{eq:proof:C9:6}
		\end{equation}
		By using $ \negOf{\beta}\models\negOf{\alpha} $ we conclude that there are only two possible cases:
		\begin{align}
		\min(\modelsOf{\negOf{\alpha}},\leq_{\Psi})  \cap \modelsOf{\negOf{\beta}} = \emptyset ,\ksOR \label{eq:proof:C9:7} \\
		\min(\modelsOf{\negOf{\alpha}},\leq_{\Psi})  \cap \modelsOf{\negOf{\beta}} = \min(\modelsOf{\negOf{\beta}},\leq_{\Psi})  \cap \modelsOf{\negOf{\beta}}  . \label{eq:proof:C9:8} 
		\end{align}
		In both cases, \eqref{eq:proof:C9:7} and \eqref{eq:proof:C9:8}, from \eqref{eq:proof:C9:6} we directly infer:
		\begin{equation}
		\modelsOfES{\Psi\change\alpha\change\beta}  \cap \modelsOf{\negOf{\beta}}  = \modelsOfES{\Psi} \cup \min(\modelsOf{\negOf{\beta}},\leq_{\Psi})  \cap \modelsOf{\negOf{\beta}}  \label{eq:proof:C9:9}
		\end{equation}
		From \eqref{eq:proof:C9:9} and 
		$ \modelsOfES{\Psi\change\beta} = \modelsOfES{\Psi} \cup \min(\modelsOf{\negOf{\beta}},\leq_{\Psi}) $, obtained from Equation \eqref{eq:repr_es_contraction},
		we conclude that
		\begin{equation}
		\modelsOfES{\Psi\change\alpha\change\beta} \cap \modelsOf{\negOf{\beta}} = \modelsOfES{\Psi\change\beta} \cap \modelsOf{\negOf{\beta}}
		\end{equation}
		holds,
		which is equivalent to $ \beliefsOf{\Psi\change\alpha\change\beta} =_\negOf{\beta} \beliefsOf{\Psi\change\beta} $.
\end{description}

\textbf{Part II:}
For the (a) to (b) direction suppose that $ \change $ is an AGM contraction operator for epistemic states. Further assume that $ \div $ satisfies \eqref{pstl:C8} and \eqref{pstl:C9}.
By Proposition \ref{prop:es_contraction} there exists a faithful assignment $ \Psi\mapsto \leq_\Psi $ such that for every proposition $ \alpha $ Equation \eqref{eq:repr_es_contraction} holds.
We will show that $ \Psi\mapsto \leq_\Psi $ satisfies \eqref{pstl:CR8} and \eqref{pstl:CR9}.
\begin{description}
	\item[\eqref{pstl:CR8}]  Suppose $ \omega_1,\omega_2\in\modelsOf{\alpha} $. We choose $ \beta=\negOf{(\omega_1\lor\omega_2)} $ and therefore, we have $ \negOf{\alpha}\models\beta $ and $ \negOf{\beta}\models\alpha $.
	By \eqref{pstl:C8} we have $ \beliefsOf{\Psi\change\alpha\change\beta}=_\alpha\beliefsOf{\Psi\change\beta} $, which implies:
	\begin{equation}
	\modelsOfES{\Psi\change\beta} =_\alpha \modelsOfES{\Psi\change\alpha\change\beta} \label{eq:ual:CR8},
	\end{equation}
	which is equivalent to $ \modelsOfES{\Psi\change\beta} \cap \modelsOf{\alpha} = \modelsOfES{\Psi\change\alpha\change\beta} \cap \modelsOf{\alpha} $.
	From Equation \eqref{eq:repr_es_contraction} we obtain that
	\begin{align}
	\modelsOfES{\Psi\change\alpha\change\beta} & =\modelsOfES{\Psi\change\alpha}\cup\min(\modelsOf{\negOf{\beta}},\leq_{\Psi\change\alpha})                                  \notag \\
	& =\modelsOfES{\Psi}\cup\min(\modelsOf{\negOf{\alpha}},\leq_{\Psi})\cup\min(\modelsOf{\negOf{\beta}},\leq_{\Psi\change\alpha}) \label{eq:CR8:2}
	\end{align} 
	and
	\begin{equation}
	\modelsOfES{\Psi\change\beta}=\modelsOfES{\Psi}\cup\min(\modelsOf{\negOf{\beta}},\leq_{\Psi}) . \label{eq:CR8:3}
	\end{equation}
	Substituting \eqref{eq:CR8:2} and \eqref{eq:CR8:3}  into Equation \eqref{eq:ual:CR8} leads to 
	\begin{multline}
	\modelsOfES{\Psi}\cup\min(\modelsOf{\negOf{\alpha}},\leq_{\Psi})\cup\min(\modelsOf{\negOf{\beta}},\leq_{\Psi\change\alpha}) \\ =_\alpha 
	\modelsOfES{\Psi}\cup\min(\modelsOf{\negOf{\beta}},\leq_{\Psi}) . \label{eq:CRX:1}
	\end{multline}
	Equation \eqref{eq:CRX:1} is equivalent to:
	\begin{multline}
	\left( \modelsOfES{\Psi}\cup\min(\modelsOf{\negOf{\alpha}},\leq_{\Psi})\cup\min(\modelsOf{\negOf{\beta}},\leq_{\Psi\change\alpha}) \right) \cap \modelsOf{\alpha} \\
	 = 	\left( \modelsOfES{\Psi}\cup\min(\modelsOf{\negOf{\beta}},\leq_{\Psi}) \right) \cap \modelsOf{\alpha} \label{eq:CRX:2}
	\end{multline}
	Because $ \min(\modelsOf{\negOf{\alpha}},\leq_{\Psi}) \!\cap\! \modelsOf{\alpha}\!=\!\emptyset $, Equation \eqref{eq:CRX:1} is equivalent to:
	\begin{multline}
	\left( \modelsOfES{\Psi}\cup\min(\modelsOf{\negOf{\beta}},\leq_{\Psi\change\alpha}) \right) \cap \modelsOf{\alpha} \\ =
	\left( \modelsOfES{\Psi}\cup\min(\modelsOf{\negOf{\beta}},\leq_{\Psi}) \right) \cap \modelsOf{\alpha} \label{eq:CRX:3}
	\end{multline}
	Remember that $ \negOf{\beta}\models\alpha $ and therefore $ \modelsOf{\negOf{\beta}}\subseteq\modelsOf{\alpha} $.
	Equation \eqref{eq:CRX:3} implies
	\begin{multline}
	\left( \modelsOfES{\Psi}\cup\min(\modelsOf{\negOf{\beta}},\leq_{\Psi\change\alpha}) \right) \cap \modelsOf{\negOf{\beta}} \\
	 =
	\left( \modelsOfES{\Psi}\cup\min(\modelsOf{\negOf{\beta}},\leq_{\Psi}) \right) \cap \modelsOf{\negOf{\beta}} \label{eq:CRX:4},
	\end{multline}%
	which is equivalent to $ \modelsOfES{\Psi}\cup\min(\modelsOf{\negOf{\beta}},\leq_{\Psi\change\alpha}) =_\negOf{\beta} \modelsOfES{\Psi}\cup\min(\modelsOf{\negOf{\beta}},\leq_{\Psi}) $.

	We will now show that the following holds:
	\begin{equation}
	\min(\modelsOf{\negOf{\beta}},\leq_{\Psi\change\alpha}) = \min(\modelsOf{\negOf{\beta}},\leq_{\Psi}) \label{eq:CR8:4}
	\end{equation}
	\textbf{Case 1:} Consider the case of $ \modelsOfES{\Psi}\cap\modelsOf{\negOf{\beta}}\neq\emptyset $. 
	Because of the faithfulness of the assignment, it must be the case that $  \min(\modelsOf{\negOf{\beta}},\leq_{\Psi}) \subseteq \modelsOfES{\Psi} $. 
	Moreover, $ \min(\modelsOf{\negOf{\beta}},\leq_{\Psi}) $ is exactly the set of models of $ \negOf{\beta} $ contained in $ \modelsOfES{\Psi} $, i.e.:
	\begin{equation}
	\modelsOfES{\Psi} \cap \modelsOf{\negOf{\beta}} = \min(\modelsOf{\negOf{\beta}},\leq_{\Psi}) \label{eq:CRX:5}
	\end{equation}
	From Equation \eqref{eq:repr_es_contraction} in Proposition \eqref{prop:es_contraction}
	we easily get
	\begin{equation}
	\modelsOfES{\Psi\change\alpha} \cap \modelsOf{\alpha} = \left( \modelsOfES{\Psi}\cup\min(\modelsOf{\negOf{\alpha}},\leq_{\Psi}) \right) \cap \modelsOf{\alpha},
	\end{equation}
	which is, because of $ \min(\models(\negOf{\alpha}),\leq_{\Psi}) \cap \modelsOf{\alpha}=\emptyset $, equivalent to:
	\begin{equation}
	\modelsOfES{\Psi\change\alpha} \cap \modelsOf{\alpha} = \modelsOfES{\Psi} \cap \modelsOf{\alpha}
	\end{equation}
	Since $ \negOf{\beta}\models\alpha $, 		
	the set $ \modelsOfES{\Psi\change\alpha} $ contains the same models of $ \negOf{\beta} $ as $ \modelsOfES{\Psi} $, i.e.:
	\begin{equation}
	\modelsOfES{\Psi\change\alpha} \cap \modelsOf{\negOf{\beta}} = \modelsOfES{\Psi} \cap \modelsOf{\negOf{\beta}} \label{eq:CRX:6}
	\end{equation}
	Due to our assumption $ \modelsOfES{\Psi}\cap\modelsOf{\negOf{\beta}} \neq \emptyset $, this implies $ \modelsOfES{\Psi\change\alpha}\cap\modelsOf{\negOf{\beta}}\neq\emptyset $. 
	Then, because of the faithfulness of the assignment, it must be the case that $  \min(\modelsOf{\negOf{\beta}},\leq_{\Psi\change\alpha}) \subseteq \modelsOfES{\Psi\change\alpha} $. 
	Moreover, $ \min(\modelsOf{\negOf{\beta}},\leq_{\Psi\change\alpha}) $ is exactly the set of models of $ \negOf{\beta} $ contained in $ \modelsOfES{\Psi\change\alpha} $, i.e.:
	\begin{equation}
	\modelsOfES{\Psi\change\alpha} \cap \modelsOf{\negOf{\beta}} = \min(\modelsOf{\negOf{\beta}},\leq_{\Psi\change\alpha}) \label{eq:CRX:7}
	\end{equation}
	Together, Equations \eqref{eq:CRX:5}, \eqref{eq:CRX:6} and \eqref{eq:CRX:7}, imply
	Equation \eqref{eq:CR8:4} in
	this case.

	\textbf{Case 2:} We now consider the other case $ \modelsOfES{\Psi}\cap\modelsOf{\negOf{\beta}}=\emptyset $.
	Since $ \negOf{\beta}\models\alpha $ and $ \min(\modelsOf{\negOf{\alpha}},\leq_{\Psi}) $ contains no models of $ \alpha $, it must be the case that
	\begin{equation}
	\modelsOfES{\Psi}\cup\min(\modelsOf{\negOf{\beta}},\leq_{\Psi\change\alpha}) =_\alpha 
	\modelsOfES{\Psi}\cup\min(\modelsOf{\negOf{\beta}},\leq_{\Psi}) .  \label{eq:CR8:3a}
	\end{equation}
	We directly conclude from Equation \eqref{eq:CR8:3a} that Equation \eqref{eq:CR8:4} holds, which finishes the proof of Equation \eqref{eq:CR8:4}.

	Note that $ \modelsOf{\negOf{\beta}} $ has only two elements, $ \modelsOf{\negOf{\beta}}=\{\omega_1,\omega_2\} \subseteq \modelsOf{\alpha} $, and thus information about the minima provides us the relative order of the two elements $ \omega_1$ and $\omega_2 $. So, from Equation \eqref{eq:CR8:4}, we can conclude that $ \omega_1 \leq_\Psi \omega_2 $ if and only if $ \omega_1 \leq_{\Psi\change\alpha} \omega_2 $.
	\medskip
	\item[\eqref{pstl:CR9}] Suppose $ \omega_1,\omega_2\in\modelsOf{\negOf{\alpha}} $. We choose $ \beta=\negOf{(\omega_1\lor\omega_2)} $ and therefore, we have $ \alpha\models\beta $.
	By \eqref{pstl:C9} we have $ \beliefsOf{\Psi\change\alpha\change\beta}=_\negOf{\beta}\beliefsOf{\Psi\change\beta} $, which implies:
	\begin{equation}
	\modelsOfES{\Psi\change\beta} \cap \modelsOf{\negOf{\beta}} = \modelsOfES{\Psi\change\alpha\change\beta} \cap \modelsOf{\negOf{\beta}} \label{eq:ual:CR9:1}
	\end{equation}
	From Equation \eqref{eq:repr_es_contraction} we obtain that 
	\begin{align}
	\modelsOfES{\Psi\change\alpha} & =\modelsOfES{\Psi}\cup\min(\modelsOf{\negOf{\alpha}},\leq_{\Psi})  \label{eq:CR9:X1} , \\
	\modelsOfES{\Psi\change\beta} & =\modelsOfES{\Psi}\cup\min(\modelsOf{\negOf{\beta}},\leq_{\Psi})  \label{eq:CR9:3}
	\end{align}
	and employing Equation \eqref{eq:repr_es_contraction} twice yields:		
	\begin{align}
	\modelsOfES{\Psi\change\alpha\change\beta} 
	=\modelsOfES{\Psi}\cup\min(\modelsOf{\negOf{\alpha}},\leq_{\Psi})\cup\min(\modelsOf{\negOf{\beta}},\leq_{\Psi\change\alpha}) \label{eq:CR9:2}.
	\end{align} 
	Substituting \eqref{eq:CR9:2} and \eqref{eq:CR9:3}  into Equation \eqref{eq:ual:CR9:1} leads to 
\begin{multline}
	\left( \modelsOfES{\Psi}\!\cup\!\min(\modelsOf{\negOf{\beta}},\leq_{\Psi}) \right) \!\cap\! \modelsOf{\negOf{\beta}} \\
	= \left(\modelsOfES{\Psi}\!\cup\!\min(\modelsOf{\negOf{\alpha}},\leq_{\Psi})\!\cup\!\min(\modelsOf{\negOf{\beta}},\leq_{\Psi\change\alpha}) \right) \! \cap \! \modelsOf{\negOf{\beta}}. \label{eq:ual:CR9:4}
	\end{multline}
	Note that every model of $ \negOf{\beta} $ is a model of $ \negOf{\alpha} $, therefore, either one of the following holds:
	\begin{align}		\min(\modelsOf{\negOf{\alpha}},\leq_{\Psi}) \cap \modelsOf{\negOf{\beta}} & = \emptyset ,\ksOR \label{eq:ual:CR9:5} \\
	\min(\modelsOf{\negOf{\alpha}},\leq_{\Psi}) \cap \modelsOf{\negOf{\beta}} & = \min(\modelsOf{\negOf{\beta}},\leq_{\Psi})  . \label{eq:ual:CR9:6} 
	\end{align}
	In the case of \eqref{eq:ual:CR9:5}, Equation \eqref{eq:ual:CR9:4} reduces to $  
	\left({\modelsOfES{\Psi}\cup\min(\modelsOf{\negOf{\beta}},\leq_{\Psi})}\right) \cap \modelsOf{\negOf{\beta}} 
	=
	\left({\modelsOfES{\Psi}\cup\min(\modelsOf{\negOf{\beta}},\leq_{\Psi\change\alpha})}\right) \cap \modelsOf{\negOf{\beta}} $. 
	Furthermore, from \eqref{eq:ual:CR9:5} and $ \negOf{\beta}\models\negOf{\alpha} $ we conclude $ \modelsOfES{\Psi}\cap\modelsOf{\negOf{\beta}}=\emptyset $. This allows to conclude $ \min(\modelsOf{\negOf{\beta}},\leq_{\Psi\change\alpha}) = {\min(\modelsOf{\negOf{\beta}},\leq_{\Psi})} $.
	
	For the other case, the case of \eqref{eq:ual:CR9:6}, note that $ \min(\modelsOf{\negOf{\alpha}},\leq_{\Psi}) \subseteq \modelsOfES{\Psi \change \alpha} $. 
	By Equation \eqref{eq:ual:CR9:6} it must hold that $ {\min(\modelsOf{\negOf{\beta}},\leq_{\Psi})} \subseteq {\modelsOfES{\Psi \change \alpha}}  $. By the faithfulness of $ \Psi\mapsto\leq_{\Psi} $ (in particular condition \eqref{pstl:FA1} in Definition \ref{def:faithful_assignment}) we have $ \min(\modelsOf{\negOf{\beta}},\leq_{\Psi})=\min(\modelsOf{\negOf{\beta}},\leq_{\Psi\change\alpha}) $, because the minimal models of $ \negOf{\beta} $ with respect to $ \leq_{\Psi\change\alpha} $ are contained in $ \modelsOfES{\Psi \change \alpha} $ by \eqref{eq:CR9:X1} and \eqref{eq:ual:CR9:6}.
	
	In summary, in both cases, \eqref{eq:ual:CR9:5} and \eqref{eq:ual:CR9:6}, we can conclude:
	\begin{equation}
	\min(\modelsOf{\negOf{\beta}},\leq_{\Psi\change\alpha}) = \min(\modelsOf{\negOf{\beta}},\leq_{\Psi}) \label{eq:ual:CR9:7}
	\end{equation}
	Note that $ \modelsOf{\negOf{\beta}} $ has only two elements, $ \modelsOf{\negOf{\beta}}=\{\omega_1,\omega_2\} \subseteq \modelsOf{\negOf{\alpha}} $, and thus information about the minima provides us the relative order of the two elements $ \omega_1$ and $\omega_2 $. From Equation \eqref{eq:ual:CR9:7} we can conclude that $ \omega_1 \leq_\Psi \omega_2 $ if and only if $ \omega_1 \leq_{\Psi\change\alpha} \omega_2 $.\qedhere
\end{description}
\endgroup \end{proof}

\pagebreak[3]
\begin{apxproposition}{prop:eqrelated_c10_C11new}
Let $ \contraction $  be an AGM contraction operator and let $ \Psi\mapsto{\leq_{\Psi}} $ be a faithful assignment \ref{eq:repr_es_contraction} with \( \contraction \).
    Then the following statements are equivalent:
    \begin{enumerate}[(a)]
        \item The operator $ \contraction $ satisfies the postulates \eqref{pstl:C10new} (respectively \eqref{pstl:C11new})
        \item The operator $ \contraction $ satisfies the postulates \eqref{pstl:CR10} (respectively \eqref{pstl:CR11})
    \end{enumerate}
\end{apxproposition}
\begin{proof}
	In the following, we will show that (b) implies (a) and that (a) implies (b).

\textbf{Part I:} We show the (b) to (a) direction.
Let $ \contraction $ be a belief change operator fulfilling \eqref{pstl:C1} to \eqref{pstl:C7}. By Proposition \ref{prop:eqrelated_c8_c9}, there is a faithful assignment $ \Psi\mapsto\leq_{\Psi} $ related to $ \contraction $ by \eqref{eq:repr_es_contraction}.
We will show that \eqref{pstl:C10new} and \eqref{pstl:C11new} are satisfied.
\begin{description}
	\item[\eqref{pstl:C10new}] 
		Let $ \gamma\models\beta $ and $ \Psi\contraction\beta\models\alpha\to\gamma $. 
		We show $ \Psi\contraction\alpha\contraction\beta \models \alpha\to\gamma $ by contradiction, i.e., conclude a contradiction from $ \Psi\contraction\alpha\contraction\beta \not\models \alpha\to\gamma $.
		In particular, we consider $ \omega \in \modelsOf{\Psi\contraction\alpha\contraction\beta} $ such that $ \omega\models \neg\gamma \land \alpha  $.
		
		By Equation \eqref{eq:repr_es_contraction} we obtain
		\begin{equation}
			\modelsOfES{\Psi\contraction\beta} = \modelsOfES{\Psi} \cup \min( \modelsOf{\negOf{\beta}} , \leq_{\Psi} ) \subseteq \modelsOf{\alpha\to\gamma}
			\label{eq:proof:C11:1_new}
		\end{equation}
		and with Equation \eqref{eq:proof:C8:4} which holds for every $ \alpha,\beta $ we get:
		\begin{align}
			\modelsOfES{\Psi\!\contraction\!\alpha\!\contraction\!\beta} 
			& \!=\!  \modelsOfES{\Psi} \!\cup\! \min( \modelsOf{\negOf{\alpha}} , \leq_{\Psi} ) \!\cup\! \min( \modelsOf{\negOf{\beta}} , \leq_{\Psi\contraction\alpha} ) 
			.
			\label{eq:proof:C11:2_new}
		\end{align}
		By Equation \eqref{eq:proof:C11:2_new} either $ \omega\in\modelsOfES{\Psi} $, $ \omega\in{\min( \modelsOf{\negOf{\alpha}} , \leq_{\Psi} )} $ or $ \omega\in{\min( \modelsOf{\negOf{\beta}} , \leq_{\Psi\contraction\alpha} )} $. 
		For these three cases we have:
		\begin{itemize}
			\item If $ \omega\in\modelsOfES{\Psi} $, then by Equation \eqref{eq:proof:C11:1_new} we have $ \omega\models\alpha\to\gamma $.
			\item Because of $ \omega\models \alpha $ the case of $ \omega \in \min( \modelsOf{\negOf{\alpha}} , \leq_{\Psi\contraction\alpha} ) $ is impossible.
			\item It remains the case of $ \omega \in {\min( \modelsOf{\negOf{\beta}} , \leq_{\Psi\contraction\alpha} )} $ with $ \omega\not\in \modelsOf{\Psi} $.
			By Equation \eqref{eq:proof:C11:1_new} there is an interpretation $ \omega' $ with $ \omega' \in {\min(\modelsOf{\neg\beta},\leq_{\Psi})} $, and consequently $ \omega' <_\Psi \omega $.
			In particular, remember that $ \gamma\models\beta $, and thus, $ \omega'\models\neg\alpha $. 			
			By applying \eqref{pstl:CR10} we obtain $ \omega' <_{\Psi\contraction\alpha} \omega  $, which contradicts $ \omega \in \min( \modelsOf{\negOf{\alpha}} , \leq_{\Psi\contraction\alpha} ) $.
		\end{itemize}
		Im summary, in all possible cases we obtain a contradiction which finishes this part of the proof.
	\medskip
	\item[\eqref{pstl:C11new}]
		
		Let $ \gamma\models\beta $ and $ \Psi\contraction\alpha\contraction\beta \models \neg\alpha \to \gamma $. We show that $ \Psi\contraction\beta \models \neg\alpha \to \gamma  $.

		Towards a contradiction, assume an $ \omega \in \modelsOf{\Psi\contraction\beta} $ with $ \omega \not\models \neg\alpha \to \gamma $, i.e., $ \omega \models \neg\alpha\land\neg\gamma $.  
		By Equation \eqref{eq:repr_es_contraction} we have either $ \omega\in\modelsOf{\Psi} $ or $ \omega\in {\min(\modelsOf{\neg\beta},\leq_{\Psi})} $. In the first case, we obtain the contradiction of $ \omega \in \modelsOf{\Psi\contraction\alpha\contraction\beta}  $.
		
		We consider the remaining case of $  \omega\in {\min(\modelsOf{\neg\beta},\leq_{\Psi})} $ and $ \omega\notin\modelsOf{\Psi} $. 
		This implies that $ \Psi\models\beta $ and consequently by Equation \eqref{eq:repr_es_contraction}, every interpretation of $ {\min(\modelsOf{\neg\beta},\leq_{\Psi\contraction\alpha})}$ is contained in $ \modelsOf{\alpha} $.
		Thus, there is an interpretation $ \omega' \in \modelsOf{\Psi\contraction\alpha\contraction\beta} $ with $ \omega' \in {\min(\modelsOf{\neg\beta},\leq_{\Psi\contraction\alpha})} $ and $ \omega' \in \modelsOf{\alpha} $. 
		Because of $ \omega\models \neg\beta \land \neg\gamma $ we obtain $ \omega' <_{\Psi\contraction\alpha} \omega $.
		Applying $ \omega\models \neg\alpha $ and \eqref{pstl:CR11} yields $ \omega' <_{\Psi} \omega $.
		This completes the proof by contradicting $  \omega\in {\min(\modelsOf{\neg\beta},\leq_{\Psi})} $.
\end{description}

\textbf{Part II:} We show the (a) to (b) direction.
Suppose that $ \contraction $ is a belief change operator that satisfies \eqref{pstl:C1} to \eqref{pstl:C7} and \eqref{pstl:C10new} (respectively \eqref{pstl:C11new}).
By Proposition \ref{prop:es_contraction} there exists a faithful assignment $ \Psi\mapsto \leq_\Psi $ such that for every proposition $ \alpha $ Equation \eqref{eq:repr_es_contraction} holds.
We will show that $ \Psi\mapsto \leq_\Psi $ satisfies \eqref{pstl:CR10} and \eqref{pstl:CR11}.
\begin{description}
	\item[\eqref{pstl:CR10}] Suppose $ \omega_1\in\modelsOf{\negOf{\alpha}} $, $ \omega_2\in\modelsOf{{\alpha}} $ and $ \omega_1 <_{\Psi}\omega_2 $. 
		We want to show $ \omega_1 <_{\Psi\contraction\alpha} \omega_2 $.
		Since $ \Psi\mapsto\leq_{\Psi} $ is a faithful assignment (especially by \eqref{pstl:FA2}) it must be the case that $ \omega_2\notin \modelsOfES{\Psi} $.
		
		In the case of $ \omega_1\in\modelsOf{\Psi} $ we obtain  $ \omega_2\not\in \modelsOf{\Psi\contraction\alpha}  $ from Equation \eqref{eq:repr_es_contraction}. Thus, the desired result $ \omega_1 <_{\Psi\contraction\alpha} \omega_2 $ holds.
		
		For the remaining case of $ \omega_1\notin\modelsOf{\Psi} $ let $ \beta=\negOf{(\omega_1\lor\omega_2)} $ and $ \gamma $ such that $ \modelsOf{\gamma}=\modelsOf{\Psi} $.  
		Clearly, we have $ \gamma\models\beta $ and $ \Psi\contraction\beta \models \alpha\to\gamma $. By applying \eqref{pstl:C10new} we obtain $ \Psi\contraction\alpha\contraction\beta \models \alpha\to\gamma $.
		This implies that $ \omega_2\notin \modelsOfES{\Psi\contraction\alpha\contraction\beta} $. Note that $ \modelsOf{\negOf{\beta}}=\{\omega_1,\omega_2\} $ and thus, by Equation \eqref{eq:repr_es_contraction} it must be the case that $ \omega_1\in \modelsOfES{\Psi\contraction\alpha\contraction\beta} $ or $ \omega_2\in \modelsOfES{\Psi\contraction\alpha\contraction\beta} $.
		Since the latter leads to a contradiction, we conclude $ \omega_1\in \modelsOfES{\Psi\contraction\alpha\contraction\beta} $, and in summary $ \omega_1 <_{\Psi\contraction\alpha} \omega_2 $.
	\item[\eqref{pstl:CR11}]  
		We show \eqref{pstl:CR11} by contraposition. 
		Suppose $ \omega_1\in\modelsOf{\negOf{\alpha}} $, $ \omega_2\in\modelsOf{{\alpha}} $ and $ \omega_2 <_{\Psi\contraction\alpha}\omega_1 $. 
		We will show $ \omega_2 <_{\Psi}\omega_1 $.
		
		Since $ \Psi\mapsto\leq_{\Psi} $ is a faithful assignment, and thus fulfils especially \eqref{pstl:FA2},
		it must be the case that $ \omega_1\notin \modelsOfES{\Psi\contraction\alpha} $.		
		For $ \beta=\negOf{(\omega_1\lor\omega_2)} $ we can conclude
		by Equation \eqref{eq:repr_es_contraction} that $ \modelsOfES{\Psi\contraction\alpha\contraction\beta}=\modelsOfES{\Psi\contraction\alpha} \cup {\min( \modelsOf{\negOf{\beta}} , \leq_{\Psi\contraction\alpha} )} $.
		Because of $ \omega_2 <_{\Psi\contraction\alpha} \omega_1 $ we have $ \modelsOfES{\Psi\contraction\alpha\contraction\beta} = \modelsOfES{\Psi\contraction\alpha}\cup\{ \omega_2 \} $.
		
		If $ \omega_2 \in \modelsOf{\Psi\contraction\alpha} $, then because of $ \modelsOfES{\Psi\contraction\alpha}=\modelsOfES{\Psi} \cup {\min( \modelsOf{\negOf{\alpha}} , \leq_{\Psi} )} $ we have $ \omega_2\in \modelsOf{\Psi} $.
		From $ \omega_1\notin \modelsOfES{\Psi\contraction\alpha\contraction\beta} $ we obtain that $ \omega_1\notin \modelsOf{\Psi} $. Consequently, $ \omega_2 <_\Psi \omega_1 $.
		
		If $ \omega_2 \notin \modelsOf{\Psi\contraction\alpha} $, then $ \omega_2 \in {\min( \modelsOf{\negOf{\beta}} , \leq_{\Psi\contraction\alpha} )} $. 
		For $ \gamma $ with $ \gamma\equiv \beliefsOf{\Psi\contraction\alpha} $ observe that $ \gamma\models\beta $ and consequently $ \Psi\contraction\alpha\contraction\beta \models \neg\alpha \to \gamma $.
		Application of \eqref{pstl:C11new} yields $ \Psi\contraction\beta \models \negOf{\alpha} \to \gamma $.
		Remember that $ \modelsOf{\negOf{\beta}}=\{\omega_1,\omega_2\} $ and thus either $ \omega_1\in \modelsOfES{\Psi\contraction\beta} $ or $ \omega_2\in \modelsOfES{\Psi\contraction\beta} $ by Equation \eqref{eq:repr_es_contraction}.
		Note that the first case is impossible, hence $ \omega_1\not\models \negOf{\alpha} \to \gamma $.
		Consequently, we obtain $ \omega_2 <_{\Psi}\omega_1 $ which completes the proof. \qedhere
\end{description} \end{proof}

\begin{apxproposition}{prop:eqrelated_c10_c11_old}
    For a AGM contraction operator $ \contraction $ the following two statements hold:
    \begin{enumerate}[(a)]
        \item If \( \contraction \)  satisfies \eqref{pstl:C8}, then $ \contraction $ satisfies \eqref{pstl:C10}  if and only if \( \contraction \) satisfies \eqref{pstl:C10new}.
        \item If \( \contraction \)  satisfies \eqref{pstl:C9}, then $ \contraction $ satisfies \eqref{pstl:C11}  if and only if \( \contraction \) satisfies \eqref{pstl:C11new}.
    \end{enumerate}
\end{apxproposition}
\begin{proof}
	Note that due to Proposition \ref{prop:eqrelated_c8_c9}, satisfaction of \eqref{pstl:C8} is equivalent to satisfaction of \eqref{pstl:CR8}, and satisfaction of \eqref{pstl:C9} is equivalent to satisfaction of \eqref{pstl:CR9}, respectively.
The general proof strategy is to show that \eqref{pstl:C10} is equivalent to \eqref{pstl:CR10} in the light of \eqref{pstl:CR8}, and to show that \eqref{pstl:C11} is equivalent to \eqref{pstl:CR11} in the light of \eqref{pstl:CR9}, respectively.

We start by showing that $ \Psi\mapsto \leq_\Psi $ satisfies \eqref{pstl:CR10} when \eqref{pstl:C10} is satisfied, and we show satisfaction of \eqref{pstl:CR11}  when \eqref{pstl:C11} is satisfied, respectively. 
Suppose that $ \change $ is a belief change operator that satisfies \eqref{pstl:C1} to \eqref{pstl:C7}.
By Proposition \ref{prop:es_contraction} there exists a faithful assignment $ \Psi\mapsto \leq_\Psi $ such that for every proposition $ \alpha $ Equation \eqref{eq:repr_es_contraction} holds.
\begin{description}
    \item[\eqref{pstl:CR10}] 
    Suppose $ \omega_1\in\modelsOf{\negOf{\alpha}} $, $ \omega_2\in\modelsOf{{\alpha}} $ and $ \omega_1 <_{\Psi}\omega_2 $. We want to show $ \omega_1 <_{\Psi\change\alpha} \omega_2 $.
    For this purpose let $ \beta=\negOf{(\omega_1\lor\omega_2)} $. 
    Since $ \Psi\mapsto\leq_{\Psi} $ is a faithful assignment (especially by \eqref{pstl:FA2}) it must be the case that $ \omega_2\notin \modelsOfES{\Psi} $.
    By use of Equation \eqref{eq:repr_es_contraction} we can conclude that $ \omega_2\notin\modelsOfES{\Psi\change\beta} $ and $ \omega_1\in\modelsOfES{\Psi\change\beta} $.
    Now let $ \gamma=\gamma'\lor\negOf{\alpha} $, where $ \gamma' $ is a formula such that $ \modelsOfES{\Psi\change\beta}\cup\{ \omega_1 \}= \modelsOf{\gamma'} $.
    Thus, $ \negOf{\alpha} \models \gamma $, whereby $ \omega_1\models\gamma $ and $ \omega_2\not\models\gamma $. From \eqref{pstl:C10} we conclude $ \Psi\change\alpha\change\beta\models \gamma $.
    This implies that $ \omega_2\notin \modelsOfES{\Psi\change\alpha\change\beta} $. Note that $ \modelsOf{\negOf{\beta}}=\{\omega_1,\omega_2\} $ and thus, by Equation \eqref{eq:repr_es_contraction} it must be the case that $ \omega_1\in \modelsOfES{\Psi\change\alpha\change\beta} $ or $ \omega_2\in \modelsOfES{\Psi\change\alpha\change\beta} $.
    Since the latter leads to a contradiction, we conclude $ \omega_1\in \modelsOfES{\Psi\change\alpha\change\beta} $, and in summary $ \omega_1 <_{\Psi\change\alpha} \omega_2 $.
    \item[\eqref{pstl:CR11}] We show \eqref{pstl:CR11} by contraposition. 
    Suppose $ \omega_1\in\modelsOf{\negOf{\alpha}} $, $ \omega_2\in\modelsOf{{\alpha}} $ and $ \omega_2 <_{\Psi\change\alpha}\omega_1 $. 
    We will show $ \omega_2 <_{\Psi}\omega_1 $.
    Since $ \Psi\mapsto\leq_{\Psi} $ is a faithful assignment, and thus fulfils especially \eqref{pstl:FA2},
    it must be the case that $ \omega_1\notin \modelsOfES{\Psi\change\alpha} $.
    For $ \beta=\negOf{(\omega_1\lor\omega_2)} $ we can conclude
    by Equation \eqref{eq:repr_es_contraction} that $ \modelsOfES{\Psi\change\alpha\change\beta}=\modelsOfES{\Psi\change\alpha} \cup \min( \modelsOf{\negOf{\beta}} , \leq_{\Psi\change\alpha} ) $.
    Because of $ \omega_2 <_{\Psi\change\alpha} \omega_1 $ we have $ \modelsOfES{\Psi\change\alpha\change\beta} = \modelsOfES{\Psi\change\alpha}\cup\{ \omega_2 \} $.
    Now let $ \gamma=\gamma'\lor\alpha $, where $ \gamma' $ is a formula such that $ \modelsOf{\gamma'}=\modelsOfES{\Psi\change\alpha} \cup \{ \omega_2 \} $.
    By definition of $ \gamma $ it is the case that $ \Psi\change\alpha\change\beta\models \gamma $. 
    Furthermore, by definition, $ \alpha\models\gamma $ and $ \omega_1\not\models\gamma$ and $ \omega_2\models\gamma $.
    By using \eqref{pstl:C11} we can conclude $ \Psi\change\beta \models \gamma $.
    Note that by Equation \eqref{eq:repr_es_contraction} it must be the case that $ \omega_1\in \modelsOfES{\Psi\change\beta} $ or $ \omega_2\in \modelsOfES{\Psi\change\beta} $.
    Since $ \Psi\change\beta \models \gamma $ the former is not possible, so by \eqref{pstl:FA2} we have $ \omega_2 <_\Psi \omega_1 $.
\end{description}
Because of Proposition \ref{prop:eqrelated_c10_C11new}, this shows that \eqref{pstl:C10} implies \eqref{pstl:C10new}, and that \eqref{pstl:C10} implies \eqref{pstl:C11new}, respectively.

For the other direction, we will show that \eqref{pstl:C10} is satisfied when \eqref{pstl:CR8} and \eqref{pstl:CR10} are given, and we show that \eqref{pstl:C11} is satisfied when \eqref{pstl:CR9} and \eqref{pstl:CR11} are given, respectively.
\begin{description}
    \item[\eqref{pstl:C10}] 
    Let $ \negOf{\alpha}\models \gamma $ and $ \Psi\change\beta\models\gamma $. 
    We want to show that $ \Psi\change\alpha\change\beta\models\gamma $ holds.
    By Equation \eqref{eq:repr_es_contraction} we obtain
    \begin{equation}
        \modelsOfES{\Psi\change\beta} = \modelsOfES{\Psi} \cup \min( \modelsOf{\negOf{\beta}} , \leq_{\Psi} ) \subseteq \modelsOf{\gamma}
        \label{eq:proof:C11:1}
    \end{equation}
    and with Equation \eqref{eq:proof:C8:4} which holds for every $ \alpha,\beta $ we get:
    \begin{align}
        \modelsOfES{\Psi\!\change\!\alpha\!\change\!\beta} 
        & \!=\!  \modelsOfES{\Psi} \!\cup\! \min( \modelsOf{\negOf{\alpha}} , \leq_{\Psi} ) \!\cup\! \min( \modelsOf{\negOf{\beta}} , \leq_{\Psi\change\alpha} ) 
        .
        \label{eq:proof:C11:2}
    \end{align}
    We show that every $ \omega \in \modelsOfES{\Psi\change\alpha\change\beta} $ is a model of $ \gamma $. 
    By Equation \eqref{eq:proof:C11:2} either $ \omega\in\modelsOfES{\Psi} $, $ \omega\in\min( \modelsOf{\negOf{\alpha}} , \leq_{\Psi} ) $ or $ \omega\in\min( \modelsOf{\negOf{\beta}} , \leq_{\Psi\change\alpha} ) $. 
    For these three cases we have:
    \begin{itemize}
        \item If $ \omega\in\modelsOfES{\Psi} $, then by Equation \eqref{eq:proof:C11:1} we have $ \omega\models\gamma $.
        \item If $ \omega \in \min( \modelsOf{\negOf{\alpha}} , \leq_{\Psi\change\alpha} ) $, then by the assumption $ \negOf{\alpha}\models\gamma $ we have $ \omega\models\gamma $.
        \item For $ \omega \in \min( \modelsOf{\negOf{\beta}} , \leq_{\Psi\change\alpha} ) $ assume that $ \omega\models\negOf{\gamma} $. If $ \omega\models\negOf{\alpha} $, then $ \omega\models\gamma $ by the assumption $ \negOf{\alpha}\models \gamma $.
        Therefore, we can safely assume $ \omega\models\alpha $.
        Since $ \min( \modelsOf{\negOf{\beta}} , \leq_{\Psi} ) \subseteq \modelsOf{\gamma} $, there must be $ \omega_1\in \min( \modelsOf{\negOf{\beta}} , \leq_{\Psi} ) $ such that $ \omega_1 <_\Psi \omega $.
        If $ \omega_1,\omega \in \modelsOf{\alpha} $, then $ \omega_1 <_{\Psi\change\alpha} \omega $ by \eqref{pstl:CR8}.
        For $ \omega_1 \in \modelsOf{\negOf{\alpha}} $ and $ \omega \in \modelsOf{{\alpha}} $ we conclude $ \omega_1 <_{\Psi\change\alpha} \omega $ by \eqref{pstl:CR10}.
        Thus, it must be the case that $ \omega_1 <_{\Psi\change\alpha} \omega $, which is a contradiction to the minimality of $ \omega $ with respect to $ \leq_{\Psi\change\alpha} $.
    \end{itemize}
    Equation \eqref{eq:proof:C11:2} implies that $ \omega \models \gamma $, and therefore $ \Psi\change\alpha\change\beta\models\gamma $.
    \medskip
    \item[\eqref{pstl:C11}] Let $ \alpha,\beta,\gamma $ be such that 
    $ \alpha\models\gamma $
    and $ \Psi\change\alpha\change\beta\models\gamma $. 
    We want to show $ \Psi\change\beta\models\gamma $.
    By Equation \eqref{eq:repr_es_contraction} we have $ \modelsOfES{\Psi}\subseteq\modelsOf{\gamma} $ and $ {\min(\modelsOf{\negOf{\beta}},\leq_{\Psi\change\alpha})\subseteq\modelsOf{\gamma}} $.
    Now let $ \omega_1\in\modelsOf{\negOf{\beta}} $ such that $ \omega_1 \notin \min(\modelsOf{\negOf{\beta}},\leq_{\Psi\change\alpha}) $. We show that $ \omega_1\notin \min(\modelsOf{\negOf{\beta}},\leq_{\Psi}) $ or $ \omega_1\models\gamma $. Let $ \omega_2\in \min(\modelsOf{\negOf{\beta}},\leq_{\Psi\change\alpha}) $ and thus, $ \omega_2 <_{\Psi\change\alpha} \omega_1 $. We differentiate by cases:
    \begin{enumerate}
        \item For $ \omega_1\in\modelsOf{\alpha} $  the assumption $ \alpha\models\gamma $ yields $ \omega\models\gamma $.
        \item In the case of $ \omega_1\in\modelsOf{\negOf{\alpha}} $ and $ \omega_2\in\modelsOf{\alpha} $ we conclude $ \omega_2 <_{\Psi} \omega_1 $ by contraposition of \eqref{pstl:CR11}.
        \item In the case of $ \omega_1,\omega_2\in\modelsOf{\negOf{\alpha}} $ we conclude by \eqref{pstl:CR9} that $ \omega_2 <_{\Psi} \omega_1 $ holds.
    \end{enumerate}
    This shows that either $ \omega_2 <_\Psi \omega_1 $ or $ \omega_1\models\gamma $.
    The first case implies $ \omega_1\notin \min( \modelsOf{\negOf{\beta}} ,\leq_{\Psi} ) $, and thus yields $ {\min(\modelsOf{\negOf{\beta}},\leq_{\Psi})} \subseteq \modelsOf{\gamma} $.
    In summary, we have $ \modelsOfES{\Psi\change\beta} = \modelsOfES{\Psi} \cup {\min(\modelsOf{\negOf{\beta}},\leq_{\Psi})} \subseteq \modelsOf{\gamma} $.
\end{description}
In summary, we have shown the desired result.\qedhere \end{proof} 
\end{document}